\documentclass[final,3p]{elsarticle}

\usepackage{amssymb}
\usepackage{amsmath}
\usepackage{amsmath,amsfonts}
\usepackage{algorithmicx}
\usepackage{algorithm}
\usepackage{algpseudocode}
\renewcommand{\algorithmicrequire}{\textbf{Input:}}
\renewcommand{\algorithmicensure}{\textbf{Output:}} 
\usepackage{array}
\usepackage{textcomp}
\usepackage{stfloats}
\usepackage{url}
\usepackage{verbatim}
\usepackage{graphicx}
\usepackage{float} 
\usepackage{subfigure}
\usepackage{bm}
\usepackage{todonotes}
\usepackage{booktabs}
\usepackage{siunitx}
\usepackage{multirow}
\usepackage{balance}
\usepackage[table]{xcolor}
\usepackage{enumitem}
\usepackage{amssymb}
\usepackage{amsmath}
\usepackage{multicol}
\usepackage{paracol} 
\usepackage[switch]{lineno}
\usepackage{array}
\usepackage{xcolor}
\usepackage{hyphenat}
\usepackage{pifont}
\usepackage{appendix}

\journal{Neural Networks}

\begin{document}

\begin{frontmatter}

\title{Interpretable and Fair Generalized Additive Neural Networks via Multi-objective Learning}

\author[a]{Ziming Wang}
\author[b]{Changwu Huang\corref{cor}}
\author[a]{Ke Tang\corref{cor}}
\author[c,d]{\\ Yew-Soon Ong}
\author[e]{Xin Yao}

\cortext[cor]{Corresponding authors.}

\affiliation[a]{organization={Guangdong Provincial Key Laboratory of Brain-inspired Intelligent Computation, Department of Computer Science and Engineering, Southern University of Science and Technology},
            city={Shenzhen},
            postcode={518055}, 
            country={China}}
\affiliation[b]{organization={School of AI and Liberal Arts, Beijing Normal-Hong Kong Baptist University},
            city={Zhuhai},
            postcode={519087}, 
            country={China}}
            
\affiliation[c]{organization={Agency for Science, Technology and Research (A*STAR)},
            postcode={138632}, 
            country={Singapore}}
\affiliation[d]{organization={College of Computing and Data Science, Nanyang Technological University},
            postcode={639798}, 
            country={Singapore}}
\affiliation[e]{organization={School of Data Science, Lingnan University},
            city={Hong Kong}, 
            country={China}}
        
\begin{abstract}
Interpretability and fairness are two of the most emphasized dimensions in trustworthy artificial intelligence (AI). Various explainable AI methods have been introduced to improve interpretability. This paper focuses on neural network (NN)-based generalized additive models (GAMs), a class of self-interpretable models. While most existing research has prioritized improving the accuracy of NN-based GAMs, their interpretability remains largely underexplored. To address this gap, this paper introduces explicit quantitative metrics for evaluating the interpretability of NN-based GAMs, empirically examines their effectiveness, and explores strategies for improving interpretability within these models. In addition, the simultaneous and explicit optimization of both interpretability and fairness, along with their trade-offs and the underlying reasons, remains underexplored. To address this, we propose a multi-objective neural basis model (MONBM) framework based on multi-objective evolutionary learning to consider accuracy, interpretability, and fairness simultaneously. A partial retraining strategy is further developed to facilitate the practical application of evolutionary multi-objective optimization to deep model architectures. Based on \textit{MONBM}, this paper reveals the complex relationships between these dimensions and the reasons behind these intricate relationships. This analysis demonstrates how multi-objective optimization can be combined with self-interpretable models to reveal relationships among trustworthiness objectives. In addition, \textit{MONBM} obtains a set of models with different trade-offs between dimensions, and the competitiveness of the approach is validated by comparing it with state-of-the-art methods.
\end{abstract}

\begin{keyword}
Explainable AI \sep Fairness in machine learning \sep Generalized additive model \sep Multi-objective learning \sep Neural networks
\end{keyword}

\end{frontmatter}

\section{Introduction}

With the widespread adoption of artificial intelligence (AI), there is growing concern and emphasis on developing trustworthy AI systems. Trustworthy AI is a multi-dimensional concept that includes interpretability, fairness, safety, and security, among other aspects~\cite{huang2022overview,li2021trust}. Among them, interpretability and fairness are the most frequently emphasized and discussed~\cite{huang2022overview}.

To improve interpretability, various explainable AI (XAI) methods have been proposed~\cite{wang2024roadmap}, among which self-interpretable models are widely favored due to their high fidelity. In this paper, we focus on a class of self-interpretable models called generalized additive models (GAMs), which are popular for their high accuracy and ability to provide both global and local explanations. GAMs can be further classified based on the methods used to learn shape functions, including spline-based~\cite{hastie2017generalized}, tree-based~\cite{nori2019interpretml}, and neural network (NN)-based methods~\cite{agarwal2021neural,radenovic2022neural}, among others. This paper focuses on NN-based GAMs due to their superior performance and scalability~\cite{agarwal2021neural}.

However, existing research on GAMs often assumes that they inherently provide satisfactory explanations, with much of the effort devoted to architectural design for improving predictive accuracy~\cite{agarwal2021neural,radenovic2022neural}. Although GAMs do provide global and local explanations, the quality of their explanations, i.e., whether they are easy to understand, is not actually guaranteed. To the best of our knowledge, only limited research has been dedicated to improving the interpretability of GAMs, especially NN-based GAMs. Traditional regularization techniques, including weight decay and Bayesian approaches, can indirectly encourage smoother and simpler models by controlling model complexity. However, these approaches usually influence interpretability implicitly and do not provide explicit quantitative criteria for evaluating or directly optimizing interpretability as an independent objective. To address this limitation, we introduce explicitly defined quantitative metrics—smoothness and monotonicity—to evaluate and directly enhance the interpretability of NN-based GAMs from a functional perspective.

Concurrently, fairness in machine learning (ML) aims to ensure that model decisions are not biased towards any individual or group~\cite{mehrabi2021survey,pessach2022review}. It is generally categorized into individual fairness and group fairness. This paper focuses on group fairness, which requires that different groups defined by sensitive attributes should be treated equally~\cite{mehrabi2021survey}. To evaluate and improve fairness in ML models, various metrics~\cite{anahideh2021choice} and methods~\cite{zhang2022mitigating,zhang2024fairness} have been proposed. However, existing methods primarily focus on black-box models and rarely consider fairness in self-interpretable models, especially GAMs.

While interpretability and fairness have each received considerable attention in the development of trustworthy AI, they are often studied in isolation. Prior work has highlighted the conflict between accuracy and fairness~\cite{zhang2022mitigating,zhang2024fairness}, and interpretability has been framed as a multi-objective optimization problem~\cite{wang2024multi,zhou2024evolutionary,monteiro2022review}. However, research examining the relationships among accuracy, interpretability, and fairness remains relatively limited, particularly in terms of systematic empirical analysis~\cite{kemmerzell2024quantifying}. More importantly, the underlying mechanisms driving these dimensional relationships are still unclear.

Additionally, different stakeholders often have divergent needs and expectations regarding fairness and interpretability, depending on their roles and contexts. For example, critical domains like justice and healthcare both require explainability and fairness, but with differing emphases. Justice typically emphasizes fairness, whereas healthcare places greater importance on explainability.

These motivate us to formulate the problem as a multi-objective learning problem while simultaneously considering the model's accuracy, interpretability, and fairness—three often conflicting objectives—and propose an evolutionary multi-objective learning-based method to solve this problem. Leveraging the ability of multi-objective learning to obtain a set of diverse solutions, our method not only has the potential to provide tailored models for different stakeholders, but also enables us to analyze the complex relationship between optimization objectives and the underlying causes by combining with the self-explanatory capabilities of GAMs. Additionally, we validate the effectiveness of the proposed metrics and methods by comparing them with state-of-the-art methods.

Our research can be broken down into three more specific questions:

\begin{enumerate}[leftmargin=36pt]
    \item[\textit{\textbf{RQ1.}}] Are the global interpretability metrics we have devised reasonable, and can our method obtain models with the desired global interpretability?
    \item[\textit{\textbf{RQ2.}}] What are the relationships among the accuracy, interpretability, and fairness of ML models, and why do these relationships exist?
    \item[\textit{\textbf{RQ3.}}] How well can our proposed method address the identified problem, and what are its advantages compared to existing self-interpretable and black-box models?
\end{enumerate}

To answer these research questions, this paper proposes a multi-objective neural basis model (MONBM) framework\footnote{The code used in this paper is available at https://github.com/oddwang/MONBM/.} that simultaneously considers accuracy, interpretability, and fairness. The main contributions of this paper are as follows: 

\begin{enumerate}
    \item[(1)] We introduce explicit quantitative metrics for smoothness and monotonicity that are designed to directly measure and improve the global interpretability of NN-based GAMs. Unlike traditional implicit regularization methods, these metrics allow interpretability properties to be explicitly quantified, monitored, and incorporated as independent optimization objectives within a multi-objective framework. We experimentally verify both the effectiveness of the metrics and the ability of our method to obtain NBMs with desirable global interpretability.
    \item[(2)] We formulate the problem of constructing trustworthy models as a multi-objective learning problem and propose the \textit{MONBM} framework to address it. As the first to apply multi-objective learning to GAMs, the framework considers accuracy, interpretability, and fairness simultaneously, and we implement multiple instantiations of it by optimizing different objectives. Notably, our instantiations adopt an efficient strategy by retraining only the ``feature specific'' output layer of the pre-trained NBM, enabling evolutionary multi-objective optimization (EMO) to be efficiently applied to large-scale networks.
    \item[(3)] We use \textit{MONBM} to obtain a set of NBMs with different trade-offs among accuracy, interpretability, and fairness, and leverage the explanations provided by NBM to reveal the relationships between multiple dimensions and the reasons behind them. This analysis combines EMO with self-interpretable models, providing a new perspective for revealing the interrelationships among various trustworthiness objectives by leveraging the diversity of the non-dominated solution set to delve into the ``why'' of objective interplay.
    \item[(4)] We verify that \textit{MONBM} can solve the proposed problem well, obtaining a set of NBMs with desirable performance and trade-offs in multiple dimensions, and are competitive with state-of-the-art self-interpretable and black-box models.
\end{enumerate}

The rest of this paper is organized as follows. Section~\ref{sec:related_work} introduces GAM and reviews related work. Section~\ref{sec:MONBM} proposes global interpretability metrics and our \textit{MONBM} framework and its instantiations. Section~\ref{sec:experimental} answers the three research questions posed through experimental studies. Sections~\ref{sec:discussion} and~\ref{sec:conclusion} provide further discussion and summarize this paper, respectively.

\section{Background}
\label{sec:related_work}

In this section, we introduce GAM, focusing on NN-based GAM, and discuss its interpretability. Then, we discuss metrics and methods for evaluating and improving group fairness in ML. Finally, we summarize representative studies on GAM to clarify the positioning of this research.

\subsection{Generalized Additive Model (GAM)}

GAM is an extension of the linear model that establishes dependencies between input features and the target variable by summing multiple independent nonlinear mappings, known as shape functions~\cite{hastie2017generalized}. Formally, GAM can be expressed as:
\begin{equation}
    g(\mathbb{E}(y))=\beta_{0}+f_1(x_1)+f_2(x_2)+...+f_{d}(x_{d}),
    \label{eq:GAM}
\end{equation}
where $\textnormal{\textbf{x}}=\{x_1, x_2,...,x_d\}\in\mathbb{R}^d$ is the input with $d$ features, $y$ is the target variable, $g(\cdot)$ is the link function (e.g., the logistic function), and each $f_i$ is a univariate shape function.

\begin{figure}[htbp]
    \centering
    \subfigure[Global Explanation]{
        \includegraphics[width=0.4\textwidth]{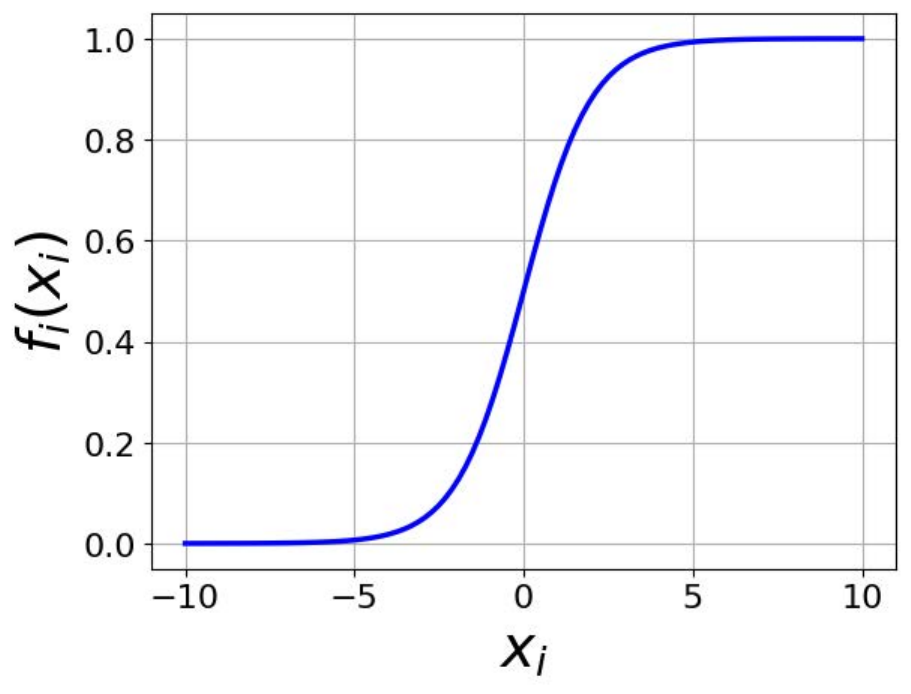}
        \label{fig:global_explaination_example}
    }
    \subfigure[Local Explanation]{
        \includegraphics[width=0.36\textwidth]{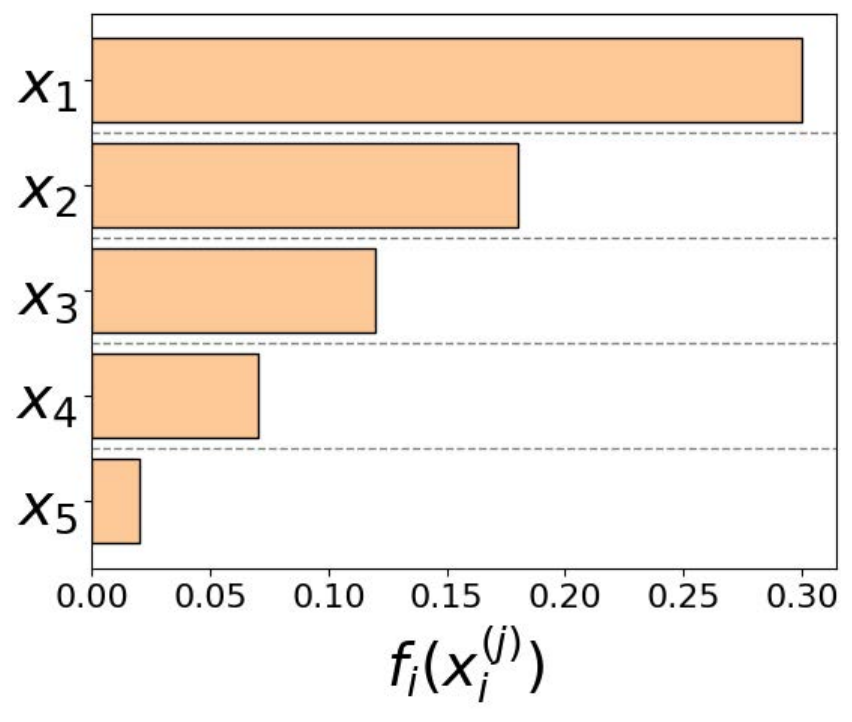}
        \label{fig:local_explaination_example}
    }
    \caption{Examples of explanations provided by GAM. For the global explanation, as shown in Fig.~\ref{fig:global_explaination_example}, the x-axis represents the input feature values $x_i$, while the y-axis corresponds to the values of the shape function $f_i(x_i)$. For the local explanation, as shown in Fig.~\ref{fig:local_explaination_example}, the y-axis represents the features $x_i$, and the x-axis represents the shape function value $f_i(x_i^{(j)})$ for the corresponding feature $x_i^{(j)}$ of data point $\textnormal{\textbf{x}}^{(j)}$. }
    \label{fig:explaination_example}
\end{figure}

Since the effect of each feature in a GAM is modeled independently without considering feature interactions, it clearly decomposes the influence of each feature on the ML model's decisions. This decomposition allows GAMs to provide two main types of explanations~\cite{kraus2024interpretable}:

\begin{enumerate}
    \item[(1)] \textbf{Global Explanation}~\cite{kraus2024interpretable}: According to Eq. (\ref{eq:GAM}), each shape function $f_i$ depends only on the $i$-th feature $x_i$ of the input vector $\textnormal{\textbf{x}}$. A global explanation is generated by visualizing each shape function $f_i$, as illustrated in Fig.~\ref{fig:global_explaination_example}. The figure illustrates how changes in the input feature affect the predicted output. By visualizing the shape functions of all features, we can capture the overall decision logic of the ML model.
    \item[(2)] \textbf{Local Explanation}~\cite{kraus2024interpretable}: As shown in Eq. (\ref{eq:GAM}), the final prediction of a GAM is the sum of all shape functions and the intercept term. For a given data point $\textnormal{\textbf{x}}^{(j)}$, the shape function value $f_i(x_i^{(j)})$ for each feature $x_i^{(j)}$ can be visualized, as illustrated in Fig.~\ref{fig:local_explaination_example}. This visualization provides insight into the contribution of each feature to the ML model's decision for that particular data point, offering a local explanation.
\end{enumerate}

Although the explanation provided by GAM appears to be heuristic, it actually provides a precise description of how GAM makes decisions. The key to GAM is how to learn its shape functions~\cite{radenovic2022neural}; its methods can be classified into spline-based~\cite{hastie2017generalized}, tree-based~\cite{nori2019interpretml}, and NN-based~\cite{agarwal2021neural,radenovic2022neural}, among others. This paper focuses on NN-based GAMs because of their superior performance and scalability~\cite{agarwal2021neural}. 

\begin{figure}[htbp]
    \centering
    \includegraphics[width=0.8\textwidth]{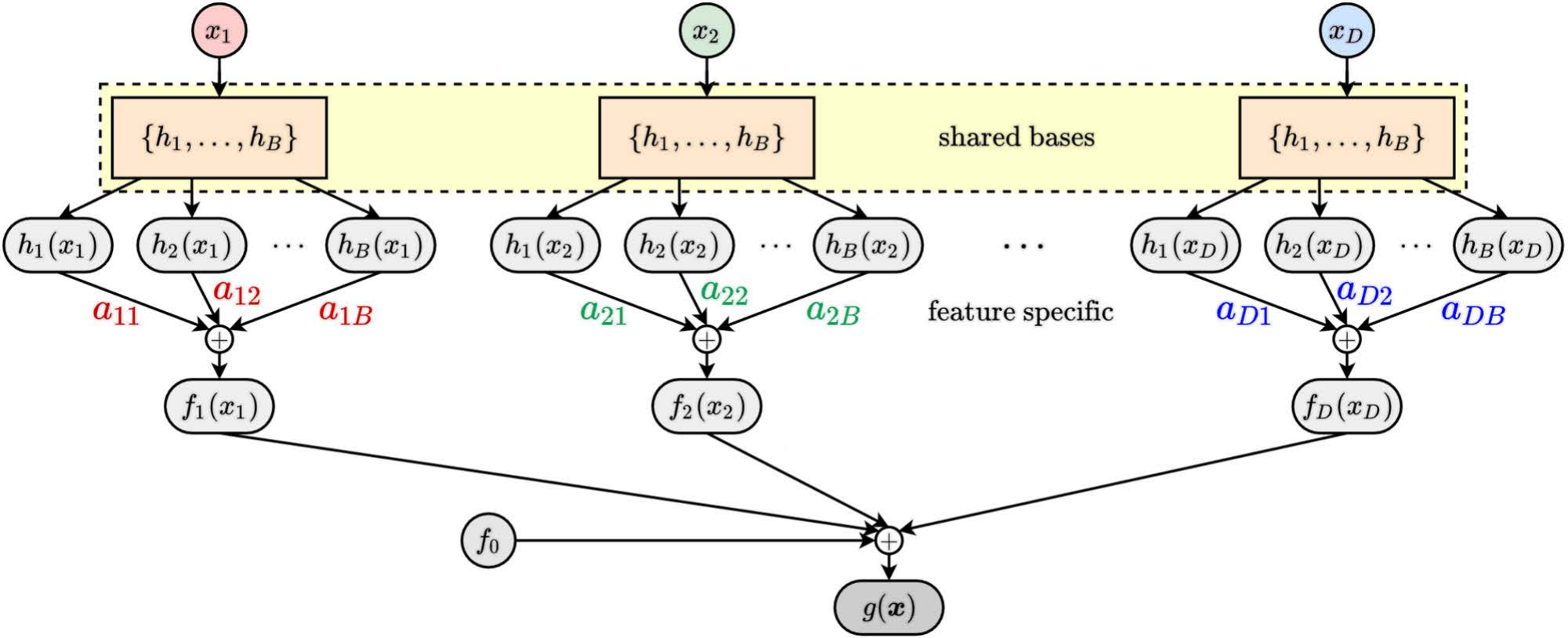}
    \caption{NBM architecture for the binary classification task, the figure taken from~\cite{radenovic2022neural} with slight modifications.}
    \label{fig:NBM}
\end{figure}

\textbf{\textit{NN-based GAMs:}} Agarwal \textit{et al.}~\cite{agarwal2021neural} proposed the first NN-based GAM, known as the neural additive model (NAM), in which each shape function is represented by a deep neural network (DNN). However, NAM's design, which requires a separate DNN for each feature, limits its scalability when applied to datasets with many features. To address this, Radenovic \textit{et al.}~\cite{radenovic2022neural} proposed the neural basis model (NBM), as illustrated in Fig.~\ref{fig:NBM}. NBM employs a single DNN, specifically a multi-layer perceptron (MLP) with \textit{one}-input and \textit{B}-outputs, as a shared basis function, replacing the need for a separate DNN for each feature. The shape function of each feature is then obtained through a linear combination of the ``feature specific'' layer, which has parameters of size $D\times (B+1)$, significantly reducing the ML model's parameters, particularly when handling datasets with high dimensions.

It is worth mentioning that, to improve predictive performance, standard GAMs have been extended to model pairwise feature interactions, resulting in the GAM with pairwise interactions ($\text{GA}^2\text{M}$)~\cite{lou2013accurate}. However, this extension often reduces interpretability and introduces substantial computational complexity. In particular, modeling bivariate interactions leads to a combinatorial increase in candidate interactions ($O(d^2)$), making effective structure discovery and feature selection essential. To address this issue, recent studies have introduced dedicated regularization and structure-learning mechanisms. For example, Chang \textit{et al.}~\cite{chang2021node} employ differentiable trees to improve the scalability of interaction discovery, while Yang \textit{et al.}~\cite{yang2021gami} introduce structural constraints to identify important main effects and interactions. Similarly, Xu \textit{et al.}~\cite{xu2023sparse} incorporate group sparsity to enable automatic feature selection, and Zhu \textit{et al.}~\cite{zhu2024neural} propose a learnable gating mechanism to distinguish different feature types. Since the primary focus of this paper is not to pursue extreme predictive performance at the expense of model transparency, both our method and the baseline methods center on standard GAM formulations. Nevertheless, our proposed method can be naturally extended to $\text{GA}^2\text{M}$ and integrated with existing structure discovery techniques. Specifically, each individual in the population can be configured as a $\text{GA}^2\text{M}$ (e.g., $\text{NB}^2\text{M}$) to natively accommodate pairwise interactions. Concurrently, existing sparsity constraints or gating techniques can be applied directly to the pre-trained NBM to effectively filter out redundant interactions, thereby reducing the number of candidate features while retaining key structural components.

Parallel to these structural advancements within the tabular domain, recent research has also expanded the scope of GAMs to specialized structured modalities, such as time series and images. For example, GATSM~\cite{kim2026transparent} adapts additive modeling to multivariate time-series forecasting by combining an NBM-based feature representation with an interpretable temporal modeling module. In computer vision settings, the original NBM framework~\cite{radenovic2022neural} further demonstrates that additive reasoning can operate on human-interpretable concepts extracted from images rather than directly on raw pixels. Since \textit{MONBM} is designed as a model-agnostic optimization framework that separates optimization objectives from the underlying self-interpretable architecture, and notably because these specialized frameworks~\cite{kim2026transparent,radenovic2022neural} share the exact same NBM backbone architecture as our instantiation, these developments suggest a feasible extension path. Specifically, temporal or concept-level additive models could serve as individuals within \textit{MONBM}, while existing optimization objectives could be adapted to evaluate task-specific notions of accuracy and interpretability. Such extensions may enable systematic analysis of trade-offs among multiple trustworthiness dimensions in broader application scenarios, which we leave for future work.

\subsection{Interpretability of NN-based GAM}
\label{sec:related_work_Understandability}

The interpretability of NN-based GAMs can be further categorized into local interpretability and global interpretability based on the type of explanation they provide, as described below.

\subsubsection{Local Interpretability (LI)} \textit{LI} emphasizes the ability to interpret a given data point at a local level. Bhatt \textit{et al.}~\cite{ijcai2020p417} and Chalasani \textit{et al.}~\cite{chalasani2020concise} proposed the $\mu_C$ and $G$ metrics, respectively, to assess the complexity of local explanations, which refers to the ability to explain model decisions with a small number of features~\cite{wang2024multi}. The $\mu_C$ and $G$ metrics are defined as follows:
\begin{equation}
    \mu_C(g;\textnormal{\textbf{x}}^{(j)})=-\sum_{i=1}^{d}\mathbb{P}(i)\textnormal{ln}(\mathbb{P}(i)), 
    \label{eq:com_uc}
\end{equation}
\begin{equation}
    \mathbb{P}(i)=\frac{|f_i(x_i^{(j)})|}{\sum_{k=1}^{d}|f_k(x_k^{(j)})|}.
    \label{eq:com_p_g}
\end{equation}
\begin{equation}
    G(g;\textnormal{\textbf{x}}^{(j)})=1-2\sum_{i=1}^d\frac{v_i^{(j)}}{||\bm{v}^{(j)}||_1}\left(\frac{d-i+0.5}{d}\right),
    \label{eq:com_G}
\end{equation}
\begin{equation}
    \bm{v}^{(j)}=\textnormal{sort}_{\textnormal{non-decreasing}}([|f_1(x_1^{(j)})|,|f_2(x_2^{(j)})|,...,|f_d(x_d^{(j)})|]).
\end{equation}

The complexity metric is based on the idea that the local explanations are more complex when feature importance scores (derived from the shape function) are evenly distributed across features, and simpler when concentrated on a few features~\cite{wang2024multi}. In other words, explanations involving fewer dominant features are generally easier to understand. Although prior work has assessed~\cite{ijcai2020p417,chalasani2020concise} or reduced~\cite{wang2024multi} the complexity of local explanations from post-hoc methods, no studies have applied these approaches to self-interpretable models, particularly GAMs. Some NN-based GAMs, such as those incorporating feature selection~\cite{xu2023sparse,yang2021gami,zhu2024neural}, indirectly simplify local explanations by retaining only the most important features. However, this approach is suboptimal, and methods specifically aimed at reducing the complexity of local explanations in GAMs remain lacking.

\subsubsection{Global Interpretability (GI)} \textit{GI} concerns the ML model’s ability to explain the relationships between input features and the target variable at a global level.
Existing work indicates that the global interpretability provided by GAM should exhibit monotonicity~\cite{zhong2024exploring} and smoothness~\cite{maindonald2010smoothing}. These properties help ensure that the learned feature–target relationships are stable, intuitive, and consistent with domain knowledge.

For smoothness, it is worth noting that most existing techniques are not designed to explicitly enhance the smoothness of global explanations. Some approaches rely on implicit regularization to control overall model complexity. Classical techniques such as weight decay~\cite{krogh1991simple} and Bayesian priors~\cite{mackay1992bayesian} constrain parameter magnitudes and reduce overfitting, which may incidentally lead to smoother shape functions. However, their primary objective is to improve predictive generalization rather than to enforce interpretable smoothness, and they do not provide explicit quantitative measures of the smoothness of the learned functions. Similarly, while early spline-based GAMs allowed smoothness adjustment through dedicated smoothing parameters~\cite{maindonald2010smoothing}, such mechanisms are intrinsic to spline formulations and are not applicable to modern tree-based~\cite{nori2019interpretml} and NN-based GAMs~\cite{agarwal2021neural,radenovic2022neural}. Recent tree-based and NN-based GAM variants have largely focused on predictive performance, typically neglecting the smoothness of global explanations and lacking explicit metrics for their evaluation and improvement.

For monotonicity, Zhong \textit{et al.}~\cite{zhong2024exploring} emphasized that certain features should satisfy monotonicity to align with user intuition. Recent research~\cite{chen2022monotonic} encourages monotonic behaviour through penalty terms, particularly in regulated domains like credit scoring. However, existing NN-based GAM approaches still lack explicit quantitative metrics that both measure monotonicity and allow it to be optimized directly as an independent objective within multi-objective learning frameworks.

Overall, there remains a significant gap in metrics and methods to explicitly evaluate and improve the interpretability of NN-based GAMs. This gap motivates this paper to propose explicit smoothness and monotonicity metrics and to obtain NN-based GAMs with high interpretability by applying these metrics.

\subsection{Group Fairness in Machine Learning}

Group fairness refers to ML models that treat different groups equally~\cite{mehrabi2021survey}. To assess group fairness, various metrics have been proposed~\cite{anahideh2021choice}, with demographic parity (DP) being one of the most widely used~\cite{dwork2012fairness,calders2010three}, which is defined as:

\begin{equation}
\label{eq:DP}
DP = \left | P\left(\hat{y}=1 | s=s_{1}\right) - P\left(\hat{y}=1 | s=s_{2}\right) \right |,
\end{equation}
where $\hat{y}$ is the predicted label, $s$ represents the sensitive attribute (e.g., sex), and $s_1$ and $s_2$ denote two different groups associated with the sensitive attribute. In this paper, we use the \textit{DP} metric to assess the fairness of GAMs.

Based on the defined fairness metrics, researchers use three types of methods to improve fairness: (1) pre-processing methods, which adjust data before training to reduce bias~\cite{zemel2013learning,kamiran2012data}; (2) in-processing methods, which incorporate fairness considerations during training~\cite{zhang2022mitigating}; and (3) post-processing methods, which adjust the model's output after training to improve fairness~\cite{kamiran2012decision,xian2023fair,xian2024unified}. 

However, the majority of existing in-processing research is tailored specifically for standard black-box architectures, such as deep neural networks. These techniques are often closely tied to specific model structures (e.g., customized loss regularization or adversarial training strategies for unconstrained multilayer perceptrons), making them difficult to directly adapt to the additive structure of inherently interpretable models such as GAMs. To the best of our knowledge, Kraus \textit{et al.}~\cite{kraus2024interpretable} are among the few who have identified fairness issues in NN-based GAMs. They proposed a strategy to mitigate unfairness by minimizing the influence of the sensitive attribute on decision-making. However, our experiments find limited improvement in fairness with their method. Therefore, this paper addresses fairness in GAMs by incorporating fairness metrics explicitly as one of the optimization objectives within the \textit{MONBM} framework.

\subsection{Summary and Research Positioning}

To clearly present the current state of research on GAMs and clarify the positioning of this study, we summarize representative GAM approaches in Table~\ref{tab:GAM_compare} to provide a structured overview of the literature. The table highlights their model categories (e.g., NN-based or traditional GAMs) and indicates which key trustworthiness dimensions they explicitly consider, including fairness, local interpretability (model complexity), and global interpretability (smoothness and monotonicity).

\begin{table*}[h]
    \caption{Comparison of the proposed \textit{MONBM} framework with existing GAM-based approaches.}
    \label{tab:GAM_compare}
    \centering
    \resizebox{\textwidth}{!}{
    \begin{tabular}{cccccc}
        \toprule
        \multirow{2}{*}{Method} & \multirow{2}{*}{Category} & \multirow{2}{*}{Fairness} & LI & GI & GI \\
        & & & (Complexity) & (Smoothness) & (Monotonicity) \\
        \midrule
        Spline-GAM~\cite{hastie2017generalized} & Spline & \ding{56} & \ding{56} & \ding{52} & \ding{56} \\
        EBM~\cite{nori2019interpretml} & Tree & \ding{56} & \ding{56} & \ding{56} & \ding{56} \\
        NAM~\cite{agarwal2021neural} & NN & \ding{56} & \ding{56} & \ding{56} & \ding{56} \\
        NBM~\cite{radenovic2022neural} & NN & \ding{56} & \ding{56} & \ding{56} & \ding{56} \\
        SNAM~\cite{xu2023sparse} & NN & \ding{56} & \ding{52} & \ding{56} & \ding{56} \\
        GAMI-Net~\cite{yang2021gami} & NN & \ding{56} & \ding{52} & \ding{56} & \ding{56} \\
        NPLAM~\cite{zhu2024neural} & NN & \ding{56} & \ding{52} & \ding{56} & \ding{56} \\
        MNAM~\cite{chen2022monotonic} & NN & \ding{56} & \ding{56} & \ding{56} & \ding{52} \\
        Fair-NBM~\cite{kraus2024interpretable} & NN & \ding{52} & \ding{56} & \ding{56} & \ding{56} \\
        MONBM (Ours) & NN & \ding{52} & \ding{52} & \ding{52} & \ding{52} \\
        \bottomrule
    \end{tabular}
    }
\end{table*}

As shown in Table~\ref{tab:GAM_compare}, although substantial progress has been made across individual trustworthiness dimensions, existing GAM studies typically address these aspects in isolation rather than in an integrated manner. Most prior work focuses on at most one trustworthiness dimension at a time—for example, improving model complexity for local interpretability, enforcing monotonicity for global interpretability, or mitigating bias for fairness—without jointly considering their interactions. First, many NN-based GAMs (such as NAM~\cite{agarwal2021neural} and NBM~\cite{radenovic2022neural}) primarily emphasize predictive performance while neglecting interpretability and fairness. Second, although some recent studies have incorporated specific requirements such as complexity~\cite{xu2023sparse,yang2021gami,zhu2024neural}, monotonicity~\cite{chen2022monotonic}, or fairness~\cite{kraus2024interpretable}, these efforts generally introduce a single additional constraint through regularization or penalty terms within a single-objective optimization framework. As a result, they neither explicitly model multiple trustworthiness objectives simultaneously nor systematically explore the trade-offs among them.

Within this context, \textit{MONBM} formulates accuracy, fairness, and both local and global interpretability as explicit objectives in a unified multi-objective evolutionary learning framework. By jointly optimizing these four dimensions rather than introducing them as auxiliary regularizers, the proposed approach enables systematic analysis of their interactions. Instead of producing a single optimized solution, \textit{MONBM} generates a set of Pareto-optimal models, allowing practitioners to directly examine and select among different trade-offs. This perspective shifts GAM research from implicit or single-objective constraints toward an explicit multi-objective framework for trustworthy model design.

\section{Multi-objective Evolutionary Learning to Construct Accurate, Interpretable, and Fair Models}
\label{sec:MONBM}

In this section, we introduce explicit quantitative metrics designed to measure and optimize the global interpretability of NN-based GAMs. Then, the \textit{MONBM} framework is proposed to solve the considered multi-objective learning problem. The framework uses multi-objective learning methods~\cite{wang2024multi,zhang2022mitigating,chandra2006ensemble} to obtain a set of accurate, interpretable, and fair models with different trade-offs. Finally, multiple instantiations of our framework are provided by specifying the self-interpretable models used, the optimizer, and considering different optimization objectives.

\subsection{Smoothness and Monotonicity Metrics}

In global explanation, smoothness means that shape functions should change gradually over the range of feature values without abrupt fluctuations, while monotonicity requires that a feature’s shape function consistently increases or decreases, aligning with user intuition or domain knowledge. However, monotonicity should not be considered a general requirement for interpretability. In many practical scenarios, non-monotonic relationships (e.g., U-shaped patterns or periodic variations) are natural and necessary for accurately reflecting real data patterns. Therefore, monotonicity is treated in this work as an optional constraint that can be applied when supported by domain knowledge or stakeholder expectations, rather than as a default assumption. Furthermore, smoothness and monotonicity are generally only required for continuous features, not discrete ones.

Traditional regularization approaches, including weight decay and Bay\-esian methods, can indirectly promote smoother models by penalizing model complexity. However, they generally do not provide explicit quantitative measures of interpretability properties from the perspective of shape functions. To complement these approaches, we introduce two explicit quantitative metrics designed to measure and optimize global interpretability properties: $\mathcal{M}_s$ for smoothness and $\mathcal{M}_m$ for monotonicity, with the latter divided into $\mathcal{M}_{m\uparrow}$ (monotonic increasing) and $\mathcal{M}_{m\downarrow}$ (monotonic decreasing). Given a feature $x_i$, its shape function $f_i$, and $n$ points taken uniformly between the minimum and maximum values of $x_i$ (denoted as $x_{i,\textnormal{min}}$ and $x_{i,\textnormal{max}}$, respectively), represented as $\{x_{i,1}, x_{i,2},...,x_{i, n}\}$, the metrics $\mathcal{M}_{s}$, $\mathcal{M}_{m\uparrow}$, and $\mathcal{M}_{m\downarrow}$ are defined as follows:

\begin{equation}
    \mathcal{M}_{s}(f_i)=\sum_{j=1}^{n-1}w_{i,j}\;\textnormal{max}\left\{\frac{|f_i(x_{i,j+1})-f_i(x_{i,j})|}{f_{i,\textnormal{max}}-f_{i,\textnormal{min}}}-k,0\right\},
    \label{eq:smoothness}
\end{equation}

\begin{equation}
\mathcal{M}_{m\uparrow}(f_i)=\sum_{j=1}^{n-1} \textnormal{max}\left\{\frac{f_i(x_{i,j})-f_i(x_{i,j+1})}{f_{i,\textnormal{max}}-f_{i,\textnormal{min}}},0\right\},
\end{equation}

\begin{equation}
\mathcal{M}_{m\downarrow}(f_i)=\sum_{j=1}^{n-1} \textnormal{max}\left\{\frac{f_i(x_{i,j+1})-f_i(x_{i,j})}{f_{i,\textnormal{max}}-f_{i,\textnormal{min}}},0\right\},
\end{equation}
where $f_{i,\textnormal{max}}$ and $f_{i,\textnormal{min}}$ denote the maximum and minimum values of the shape function $f_i$ over the feature's range, and the hyper-parameter $k$ controls the smoothing degree. The term $w_{i,j}$ is an optional interval-specific weight that allows different levels of smoothness to be imposed across different regions of the input space. This design enables users to incorporate domain knowledge when certain regions require stronger or weaker smoothness constraints. In this work, we use the default setting $w_{i,j}=1$ for all intervals, which provides a uniform penalty when no prior knowledge about local smoothness is available. Both $\mathcal{M}_{s}$ and $\mathcal{M}_{m}$ metrics are the smaller, the better.

The proposed metrics have several desirable properties. First, they are intuitive and easy to understand: smoothness is defined by measuring the degree of abrupt change in function values, while monotonicity directly captures the continuous upward or downward trend of a function. These definitions are highly compatible with human intuition and domain knowledge. Second, the smoothness metric balances robustness and sensitivity by using a threshold-based approach that allows users to ignore small changes on demand while clearly identifying significant fluctuations. In addition, both metrics can be easily converted into differentiable forms for integration into existing ML or multi-objective optimization frameworks to guide the search for generating GAMs with higher global interpretability.

\subsection{MONBM Framework}

We first formally define the considered multi-objective learning problem to obtain a set of accurate, interpretable, and fair models. Given a dataset $\bm{\mathcal{D}}$, a self-interpretable model $g$, and its parameter matrix to be optimized $\textnormal{\textbf{w}}$, the multi-objective learning problem can be defined as:
\begin{equation}
    \mathop {\rm Minimize}_{\textnormal{\textbf{w}}}\:\bm{\mathcal{M}}(g(\textnormal{\textbf{w}});\bm{\mathcal{D}})=[\mathcal{M}_1(g(\textnormal{\textbf{w}});\bm{\mathcal{D}}),...,\mathcal{M}_k(g(\textnormal{\textbf{w}});\bm{\mathcal{D}})],
    \label{eq:MOFAE}
\end{equation}
where $\textnormal{\textbf{w}}$ is optimized to minimize a selected set of metrics $\bm{\mathcal{M}}(\cdot)$. Each $\mathcal{M}_i$ for $i=1,...,k$ represents a distinct objective, such as accuracy, interpretability, or fairness.

To solve this problem, in the \textit{MONBM} framework, we use EMO~\cite{chandra2006ensemble} to train self-interpretable models, aiming at obtaining a set of models with different trade-offs in accuracy, interpretability, and fairness. The overall process of the \textit{MONBM} framework is shown in Algorithm~\ref{alg:MONBM}.

\renewcommand{\algorithmicrequire}{\textbf{Input:}}
\renewcommand{\algorithmicensure}{\textbf{Output:}}

\begin{algorithm}[h]
	\caption{Multi-objective neural basis model (MONBM) framework} 
    \label{alg:MONBM}
	\begin{algorithmic}[1]
	\Require A training dataset $\bm{\mathcal{D}}_{\textnormal{\textit{train}}}$, an initialized model set $G=\{g_1, ... ,g_\tau\}$, a set of evaluation metrics $\bm{\mathcal{M}} = \{\mathcal{M}_1, \cdots, \mathcal{M}_k \} $, an environment selection strategy $\pi$, and a reproduction strategy ${\xi}$.
    \Ensure  A set of optimized models $G^*$.
        \State {$\bm{m}$ $\leftarrow$ Evaluate the value of metrics $\bm{\mathcal{M}}$ on $G$ using $\bm{\mathcal{D}}_{\textnormal{\textit{train}}}$.}
		\While {iteration termination conditions not met}
            \State \parbox[t]{\dimexpr\linewidth-\algorithmicindent}{%
            $G'$ ${\leftarrow}$ Select $n$ models from $G$ according to $\bm{m}$ and the environment selection strategy $\pi$.%
            }
            \vspace{0.1mm}
            \State \parbox[t]{\dimexpr\linewidth-\algorithmicindent}{%
            $G''$ ${\leftarrow}$ Generate $\mu$ models from $G'$ according to the reproduction strategy $\xi$.%
            }
            \vspace{0.1mm}
            \State \parbox[t]{\dimexpr\linewidth-\algorithmicindent}{%
            $G''$ ${\leftarrow}$ Partially train~\cite{yao1997new,yao1999evolving} $G''$ using $\bm{\mathcal{D}}_{\textnormal{\textit{train}}}$.%
            }
            \State \parbox[t]{\dimexpr\linewidth-\algorithmicindent}{\raggedright%
            $\bm{m}''$ $\leftarrow$ Evaluate the value of the metrics $\bm{\mathcal{M}}$ on $G''$ using $\bm{\mathcal{D}}_{\textnormal{\textit{train}}}$.%
            }
            \State \parbox[t]{\dimexpr\linewidth-\algorithmicindent}{%
    		$<g_1, \bm{m}_1>,\dots,<g_\tau, \bm{m}_\tau>$$\leftarrow$ Select $\tau$ models from $G \cup G''$ based on $\bm{m}$ and $\bm{m}''$ through the environment selection strategy $\pi$ and then update $G=\{g_1,\dots,g_\tau\}$ and $\bm{m}$, accordingly.%
            }
            \vspace{0.1mm}
		\EndWhile
		\State \Return the non-dominated solutions in the set of optimized models $G$ as $G^*$.
	\end{algorithmic} 
\end{algorithm}

Given a training dataset $\bm{\mathcal{D}}_{\textnormal{\textit{train}}}$, an initialized model set $G$, a set of evaluation metrics $\bm{\mathcal{M}}$, and the corresponding environment selection and reproduction strategies $\pi$ and $\xi$, the general process of the \textit{MONBM} framework is as follows: First, the initialized model set is evaluated (line 1). Then, the evolutionary loop begins (lines 2-8). In each generation, $n$ models are first selected as $G'$ according to the environment selection strategy $\pi$ (line 3). An offspring $G''$ is then generated according to the reproduction strategy $\xi$ and partial training~\cite{yao1997new,yao1999evolving} (lines 4-5). Next, the offspring $G''$ is evaluated (line 6), and a new population $G$ is selected according to the environment selection strategy $\pi$ (line 7). The above steps are repeated until the iteration termination condition is reached. Finally, an optimized model set $G^*$ is returned (line 9).

The key of \textit{MONBM} is to generate ideal offspring self-interpretable models and environment selection based on the given metrics. Multi-objective evolutionary algorithms (MOEAs)~\cite{10.1145/2792984} offer ideal reproduction and environment selection strategies that help our framework generate a set of models with good convergence and diversity.

\subsection{Instantiation of the MONBM Framework}

\begin{figure}[htbp]
    \centering
    \includegraphics[width=0.8\textwidth]{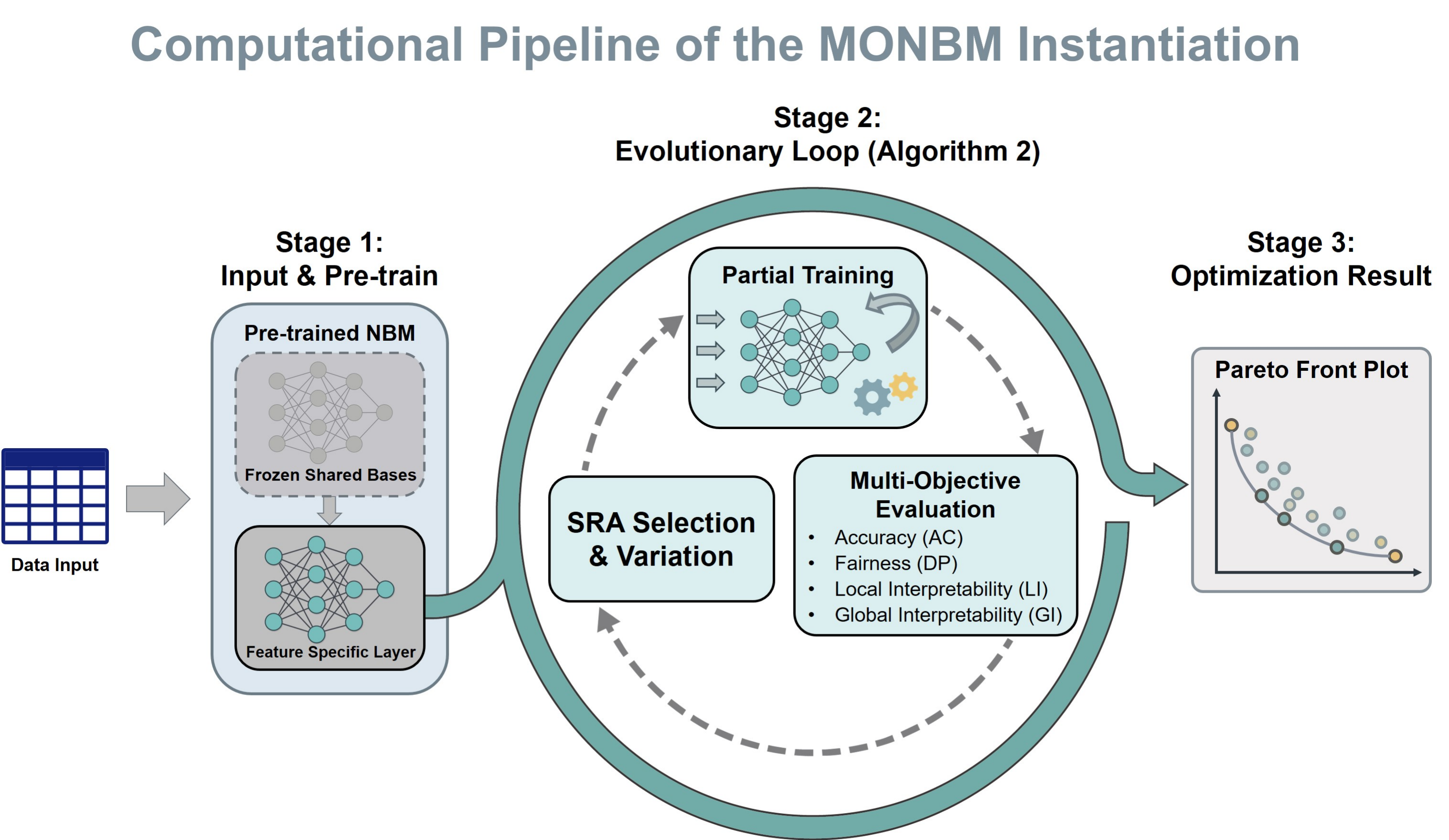}
    \caption{The computational pipeline of the \textit{MONBM} instantiation.}
    \label{fig:MONBM_framework}
\end{figure}

In the proposed \textit{MONBM} framework, the choice of self-interpretable model, initialization and optimization schemes, evaluation metrics, and optimizer can be adapted based on the specific tasks and requirements. Fig.~\ref{fig:MONBM_framework} illustrates the overall computational pipeline of our instantiation.

The workflow  is organized into three primary stages:

\begin{enumerate}
    \item[(1)] Stage 1 performs data input and pre-training of an NBM to obtain a high-accuracy base model. The shared basis functions are then frozen, which stabilizes the learned representation and reduces the search space for subsequent optimization.

    \item[(2)] Stage 2 executes the evolutionary optimization loop described in Algorithm~\ref{alg:MONBM_instantiation}. During this stage, only the unfrozen feature-specific layer is evolved and partially retrained, while the shared basis functions remain fixed. This stage combines evolutionary search with partial gradient-based training and SRA-based environmental selection to jointly optimize multiple objectives.

    \item[(3)] Stage 3 outputs the final set of non-dominated models, which approximate the Pareto front and represent different trade-offs among the considered objectives.
\end{enumerate}

Below, we provide several instantiations of the framework, which will be experimentally evaluated in Section~\ref{sec:experimental}.

\subsubsection{Model} The set of models in the \textit{MONBM} framework can be various self-inter\-pretable models. In our instantiations, a set of NBMs~\cite{radenovic2022neural} is used as individuals. Each NBM's weights and biases are encoded as a real-valued vector, representing an individual. Due to the large architecture of NBMs, training a set of NBMs via multi-objective learning is complex and time-consuming. To address this, we pre-trained an NBM, denoted as $f_{\textnormal{\textit{NBM}}}$, to achieve high accuracy. Subsequently, we re-trained only its last layer, i.e., the ``feature specific'' layer in Fig.~\ref{fig:NBM}, using the \textit{MONBM} framework. In our instantiations, each individual $g_i$ in the initialized population $G=\{g_1, ... ,g_\tau\}$ is the pre-trained NBM $f_{\textnormal{\textit{NBM}}}$, and the corresponding last layer matrix of $G$ to be optimized is denoted as $\textnormal{\textbf{W}}=\{\textnormal{\textbf{w}}_1, ... ,\textnormal{\textbf{w}}_\tau\}$. This partial-retraining strategy facilitates the practical application of EMO to large-scale neural models by reducing optimization complexity, making EMO more applicable in modern deep learning settings.

\subsubsection{Evaluation Metrics as Optimization Objectives}

In our instantiations, four evaluation metrics are explicitly treated as optimization objectives (corresponding to the $\mathcal{M}_i$ components in Eq. (\ref{eq:MOFAE})), addressing accuracy, fairness, local interpretability, and global interpretability:

\begin{enumerate}
    \item[(1)] \textbf{Accuracy}: Cross-entropy (CE) was used during optimization. The accuracy metric (ACC) was used for experimental analysis and comparisons.
    \item[(2)] \textbf{Fairness}: The widely used group fairness metric \textit{DP}~\cite{dwork2012fairness,calders2010three} was used to evaluate fairness. Obviously, other fairness metrics~\cite{anahideh2021choice} can also be considered.
    \item[(3)] \textbf{Local Interpretability (LI)}: Local interpretability was measured using the G-mean~\cite{derringer1980simultaneous} of the complexity metrics $\mu_C$~\cite{ijcai2020p417} and $G$~\cite{chalasani2020concise}, which is defined as:
    \begin{equation}
        LI=\sqrt{\mu_C \times G}.
        \label{eq:LU}
    \end{equation}
    \item[(4)] \textbf{Global Interpretability (GI)}: Global interpretability is evaluated based on the proposed smoothness and monotonicity metrics. For continuous features, smoothness is always required, and monotonicity is imposed based on prior knowledge. No requirements are imposed on discrete features. The \textit{GI} metric is defined as:
        \begin{equation}
            \footnotesize
            GI=\sum_{i=1}^d
            \begin{cases}
                \mathcal{M}_s(f_i), & \text{if } f_i \text{ requires smooth}, \\
                \mathcal{M}_s(f_i)+\mathcal{M}_{m\uparrow}(f_i), & \text{if } f_i \text{ requires smooth and} \uparrow, \\
                \mathcal{M}_s(f_i)+\mathcal{M}_{m\downarrow}(f_i), & \text{if } f_i \text{ requires smooth and} \downarrow, \\
                0, & \text{if } f_i \text{ is a discrete feature}.
            \end{cases}
            \label{eq:GU}
        \end{equation}
        
        Here, $\uparrow$ and $\downarrow$ indicate monotonically increasing and decreasing, respectively. It is worth noting that in this paper, to simplify optimization and align with \textit{LI}, we combine smoothness and monotonicity into a single \textit{GI} objective, despite their potential conflict. They remain separable within the \textit{MONBM} framework if required.
    \end{enumerate}

The optimal values for \textit{CE}, \textit{DP}, \textit{LI}, and \textit{GI} metrics are 0.0, while the optimal \textit{ACC} is 100$\%$.

\subsubsection{Optimizer} In our instantiations, we used the stochastic ranking algorithm (SRA)~\cite{li2016stochastic} as the environment selection strategy $\pi_{\textnormal{\textit{SRA}}}$. \textit{SRA} uses stochastic ranking to balance different metrics~\cite{runarsson2000stochastic} and can select individuals with good convergence and diversity~\cite{li2016stochastic}. For the reproduction strategy $\xi$, we used the variant of weight crossover~\cite{gong2018multiobjective} and isotropic Gaussian perturbation~\cite{minku2013software}, denoted as $\xi_c$ and $\xi_p$, respectively. Since all four objectives considered in our instantiations are differentiable, partial training~\cite{yao1997new,yao1999evolving} was applied to assist in exploration. 

It is worth noting that although all objective functions are differentiable and could in principle be combined into a scalarized loss, we adopt the EMO framework because the goal of this work is not only to obtain a single optimized model but also to systematically explore the trade-offs among accuracy, fairness, local, and global interpretability. Scalarization typically requires careful manual adjustment of multiple regularization weights, and the relationship between these weights and the resulting model behavior can be complex when several competing objectives are involved. By contrast, EMO naturally produces a diverse set of Pareto-optimal solutions in a single optimization process. This allows stakeholders to select models according to different practical priorities and provides a structured basis for analyzing relationships among multiple trustworthiness dimensions. In addition, using an EMO framework maintains flexibility for future extensions where certain fairness or interpretability metrics may not be differentiable, thereby improving the general applicability of the proposed framework.

\begin{algorithm}[h!]
	\caption{Instantiation of the \textit{MONBM} framework} 
    \label{alg:MONBM_instantiation}
	\begin{algorithmic}[1]
	\Require A training dataset $\bm{\mathcal{D}}_{\textnormal{\textit{train}}}$, an initialized NBM set $G=\{g_1, ... ,g_\tau\}$ and its last layer $\textnormal{\textbf{W}}=\{\textnormal{\textbf{w}}_1, ... ,\textnormal{\textbf{w}}_\tau\}$, a set of evaluation metrics $\bm{\mathcal{M}} = \{\mathcal{M}_1, \cdots, \mathcal{M}_k \} $, an environment selection strategy $\pi_{\textnormal{\textit{SRA}}}$, reproduction strategies ${\xi_c}$ and $\xi_p$, and parameter $N$ for partial training.
    \Ensure  The set of optimized NBMs $G^*$.
        \State {$\bm{m}$ $\leftarrow$ Evaluate the value of metrics $\bm{\mathcal{M}}$ on $G$ using $\bm{\mathcal{D}}_{\textnormal{\textit{train}}}$.}
		\While {iteration termination conditions not met}
            \State {$G_{best}''$ ${\leftarrow}$ $\emptyset$.}
            \For{$i\in\{1,...,k\}$}
                \State \parbox[t]{\dimexpr\linewidth-\algorithmicindent-15pt}{%
                    $G_{best\_i}'$ ${\leftarrow}$ Select the top $N$ best models on $\mathcal{M}_i$ from $G$.%
                }
                \State \parbox[t]{\dimexpr\linewidth-\algorithmicindent-15pt}{%
                    $G_{best\_i}''$ ${\leftarrow}$ Generate $N$ models from $G_{best\_i}'$ according to the reproduction strategy $\xi_p$.%
                }
                \vspace{0.1mm}
                \State \parbox[t]{\dimexpr\linewidth-\algorithmicindent-15pt}{%
                    $G_{best\_i}''$ ${\leftarrow}$ Partially train~\cite{yao1997new,yao1999evolving} $G_{best\_i}''$ using $\bm{\mathcal{D}}_{\textnormal{\textit{train}}}$ on the metric $\mathcal{M}_i$.%
                }
                \vspace{0.1mm}
                \State {$G_{best}''$ ${\leftarrow}$ $G_{best}'' \cup G_{best\_i}''$.}
            \EndFor
            \State {$\lambda=\tau-k\times N$.}
            \State \parbox[t]{\dimexpr\linewidth-\algorithmicindent}{%
            $G'$ ${\leftarrow}$ Select $\lambda$ models from $G$ according to $\bm{m}$ and environment selection strategy $\pi_{SRA}$.%
            }
            \vspace{0.1mm}
            \State \parbox[t]{\dimexpr\linewidth-\algorithmicindent}{%
            $G''$ ${\leftarrow}$ Generate $\lambda$ models from $G'$ according to the reproduction strategies $\xi_c$ and $\xi_p$.%
            }
            \vspace{0.1mm}
            \State {$G''=G''\cup G_{best}''$.}
            \State \parbox[t]{\dimexpr\linewidth-\algorithmicindent}{%
                $\bm{m}''$ $\leftarrow$ Evaluate the value of the metrics $\bm{\mathcal{M}}$ on $G''$ using $\bm{\mathcal{D}}_{\textnormal{\textit{train}}}$.%
            }
            \State \parbox[t]{\dimexpr\linewidth-\algorithmicindent}{%
    		$<g_1, \bm{m}_1>,\dots,<g_\tau, \bm{m}_\tau>$$\leftarrow$ Select $\tau$ models from $G \cup G''$ based on $\bm{m}$ and $\bm{m}''$ through the environment selection strategy $\pi_{\textnormal{\textit{SRA}}}$ and then update $G=\{g_1,\dots,g_\tau\}$ and $\bm{m}$, accordingly.%
            }
            \vspace{0.1mm}
		\EndWhile
		\State \Return the non-dominated solutions in the set of optimized NBMs $G$ as $G^*$.
	\end{algorithmic} 
\end{algorithm}

The detailed process of the instantiation based on the \textit{MONBM} framework is presented in Algorithm~\ref{alg:MONBM_instantiation}. In the evolutionary loop (lines 2-16), for each evaluation metric $\mathcal{M}_i$, the $N$ best-performing individuals are selected and trained according to the reproduction strategy $\xi_p$ and $\mathcal{M}_i$-based partial training, resulting in $k\times N$ offspring (lines 4-9). Subsequently, $\lambda$ offspring are generated based on the environmental selection strategy $\pi_{\textnormal{\textit{SRA}}}$ and reproduction strategies $\xi_c$ and $\xi_p$ (lines 11-12). Then, all offspring are evaluated (line 14), and a new population $G$ is produced according to the environmental selection strategy $\pi_{\textnormal{\textit{SRA}}}$ (line 15). This iterative cycle continues until the convergence criterion of 200 generations is met.

The overall time complexity per generation is $O(k\tau^2 + \tau \cdot n \cdot D \log D)$, where $k$ denotes the number of optimization objectives, $\tau$ represents the population size, $D$ is the feature dimension, and $n$ is the data size. A more detailed derivation of both time and space complexity is provided in Section S-II.2 of the \textit{Supplementary Material}. In addition, further analysis of the framework’s generalization behavior and robustness is presented in Section S-II.3 of the \textit{Supplementary Material}.

To analyze the relationships between dimensions and compare them with state-of-the-art methods, we considered four instantiations, each with different optimization objectives: (1) optimizing the \textit{CE} and \textit{DP} metrics; (2) optimizing the \textit{CE} and \textit{LI} metrics; (3) optimizing the \textit{CE} and \textit{GI} metrics; and (4) optimizing the \textit{CE}, \textit{DP}, \textit{LI}, and \textit{GI} metrics. These four instantiations are referred to as \textit{MONBM-CD}, \textit{MONBM-CL}, \textit{MONBM-CG}, and \textit{MONBM-CDLG}, respectively.

\section{Experimental Studies}
\label{sec:experimental}

In this section, we first give the experimental setup and then answer the three research questions posed through experiments. 

\subsection{Experimental Setup}

This subsection describes the experimental setup, including the datasets used, baseline methods for comparison, parameter settings, and evaluation criteria.

\subsubsection{Datasets} We consider six datasets that are widely used in the field of fair ML~\cite{le2022survey}, namely \textit{Adult}~\cite{Dua:2019}, \textit{Bank}~\cite{Dua:2019}, \textit{COMPAS}~\cite{angwin2016machine}, \textit{Default}~\cite{Dua:2019}, \textit{Dutch}~\cite{paul2000the}, and \textit{LSAT}~\cite{wightman1998lsac}, whose specific information can be found in Tables S-I of the \textit{Supplementary Material}. The same pre-processing was applied to each dataset: label encoding of discrete features, followed by normalization of all features with Z-score. Then, they were randomly divided into training set $\bm{\mathcal{D}}_{\textnormal{\textit{train}}}$ and test set $\bm{\mathcal{D}}_{\textnormal{\textit{test}}}$ in a ratio of 4:1. 

\subsubsection{Baseline Methods} To validate the effectiveness of the \textit{MONBM} framework to answer \textit{RQ3}, a comparison was made with the following baseline methods:

\begin{enumerate}
    \item[(1)] For \textit{MONBM-CD} instantiation, we considered the following baselines: (a) applying the method proposed in~\cite{kraus2024interpretable} to specifically improve the fairness of GAM to NBM, and referring to it as \textit{Fair-NBM}; (b) combining the existing approaches to improve the fairness of ML models at the pre-processing stage, i.e., learning fair representation (LFR)~\cite{zemel2013learning} and reweighting~\cite{kamiran2012data}, as well as at the post-processing stage, i.e., reject option classification (ROC)~\cite{kamiran2012decision}, linear post-processing (LPP)~\cite{xian2023fair}, Oracle~\cite{xian2024unified}, and FRAPP{\'E}~\cite{ctifrea2024frappe} with NBM, called \textit{LFR-NBM}, \textit{Re-NBM}, \textit{ROC-NBM}, \textit{LPP-NBM}, \textit{Oracle-NBM}, and \textit{FRAPP{\'E}-NBM}, respectively; and (c) a method based on multi-objective optimization to improve the fairness of MLP in the in-processing stage, called \textit{FairEMOL}~\cite{zhang2022mitigating}.
    \item[(2)] For \textit{MONBM-CL} instantiation, we compared it with three approaches to improve \textit{LI} based on feature selection, i.e., \textit{SNAM}~\cite{xu2023sparse}, \textit{GAMI-Net}~\cite{yang2021gami}, and \textit{NPLAM}~\cite{zhu2024neural}.
\end{enumerate}

The instantiations \textit{MONBM-CG} and \textit{MONBM-CDLG} were not compared to baseline methods due to the absence of existing work that considers global interpretability.

\subsubsection{Parameter Settings} For the NBM, the parameter settings and hyper-parameter search approach were consistent with the original configuration~\cite{radenovic2022neural}, as detailed in Section S-III.1 of the \textit{Supplementary Material}. The optimal hyper-parameter configurations obtained from the search were applied to the pre-trained NBM $f_{\textnormal{\textit{NBM}}}$ as well as to the baseline methods \textit{Fair-NBM}, \textit{LFR-NBM}, \textit{Re-NBM}, \textit{ROC-NBM}, \textit{LPP-NBM}, \textit{Oracle-NBM}, and \textit{FRAPP{\'E}-NBM}, except for differences in the number of training epochs. The parameter settings for all baseline methods are provided in Section S-III.2 of the \textit{Supplementary Material}. Additionally, the computational budget comparisons and wall-clock time comparisons are presented in Sections S-II.1 and S-II.4 of the \textit{Supplementary Material}, respectively.

In our instantiations of the \textit{MONBM} framework, the population size $\tau$ was set to 100, the mutation strength of the reproduction strategy $\xi_p$ was set to 0.01, and the iteration termination condition of the evolutionary loop was set to a maximum of 200 generations~\cite{zhang2022mitigating}. This iteration limit was established through preliminary experiments to ensure sufficient convergence across all benchmark datasets. Exploring adaptive early stopping mechanisms based on convergence or stagnation is identified as a potential direction for future work. The hyper-parameter $N$ for partial training was set to 3. We divided the dataset using 15 different random seeds and conducted independent trials.

The parameter $k$ in the smoothness metric $\mathcal{M}_s$ serves as a tolerance threshold that controls the permissible degree of localized fluctuations in shape functions. We define $k$ relative to the average smoothness of the pre-trained NBM $f_{NBM}$, i.e., $k=\eta \cdot \frac{1}{(n-1)}\sum_{j=1}^{n-1}\frac{|f_{\textnormal{\textit{NBM}},i}(x_{i,j+1})-f_{\textnormal{\textit{NBM}},i}(x_{i,j})|}{f_{\textnormal{\textit{NBM}},i,\textnormal{max}}-f_{\textnormal{\textit{NBM}},i,\textnormal{min}}}$. Here, $\eta$ is a scaling factor that modulates the stringency of the smoothness constraint. In our primary experiments, we set $\eta = 1.0$, implying that the optimized model is expected to maintain a local smoothness at every interval that is at least as consistent as the global average smoothness of the pre-trained NBM. Furthermore, a detailed discussion regarding the impact of different values for the scaling factor $\eta$ is provided in Section S-IV of the \textit{Supplementary Material}. Consistent with intuition, smaller $\eta$ values correspond to stricter smoothness requirements (effectively reducing the tolerance threshold $k$), resulting in more regularized and smoother shape functions. This design provides practical flexibility, allowing stakeholders to adjust $\eta$ according to domain knowledge or interpretability requirements.

For the monotonicity metric $\mathcal{M}_m$, we specified a subset of features required to satisfy monotonicity, as detailed in Section S-V of the \textit{Supplementary Material}. This setting is intended to evaluate whether the proposed metrics and training framework can guide shape functions toward user-specified trends, rather than to imply that monotonicity is universally required. In practical applications, the choice of whether a feature should satisfy monotonicity depends on task requirements, domain knowledge, and stakeholder preferences.

\subsubsection{Evaluation Criteria}

To answer \textit{RQ3}, we compared all non-dominated solutions obtained by \textit{MONBM} with baseline methods on the corresponding objectives (e.g., \textit{ACC} and \textit{DP} metrics on the \textit{MONBM-CD} instantiation). In comparison with population-based methods (i.e., \textit{FairEMOL}~\cite{zhang2022mitigating}), we used three commonly used metrics: hypervolume (HV) \cite{zitzler1998multiobjective}, maximum spread (MS) \cite{zitzler2000comparison}, and pure diversity (PD) \cite{solow1993measurement} to assess the convergence and diversity of the solution set. Among them, \textit{HV} is the only Pareto-compliant metric that is used to comprehensively assess the performance of the solution set~\cite{li2019quality}. 

To calculate the \textit{HV} value, we first identify the non-dominated solutions from all solutions obtained by all methods in all generations under the current data split. These solutions are used as the pseudo-Pareto front to normalize all objective values. Then, we used $(1.1, ...,1.1)$ as the reference point~\cite{zhang2022mitigating} to compute the final hypervolume, measuring the dominated space relative to the normalized front. The \textit{MS} and \textit{PD} metrics, on the other hand, focus on evaluating the diversity of the solution set. \textit{MS} evaluates the maximum expansion of each objective, while \textit{PD} measures the degree to which each solution differs from the others~\cite{li2019quality}. All three metrics are the larger, the better. 

Additionally, we compared \textit{MONBM} with population-based and single solution-based approaches to assess mutual dominance relationships, following the comparison methodologies outlined in the literature~\cite{wang2024multi} and~\cite{zhang2022mitigating}, respectively. The comparison methods in~\cite{wang2024multi} and~\cite{zhang2022mitigating} are used to evaluate the mutual dominance relationship between two solution sets and between a solution set and a single solution, respectively. A more detailed description of the comparative methods is given in Section S-VI of the \textit{Supplementary Material}.

\subsection{Experimental Results}
\label{sec:experimental results}

In this subsection, we address the three research questions introduced earlier through empirical evaluation.

\subsubsection{Validation of the Proposed Metrics (\textbf{RQ1})} 

To validate the proposed metrics, we present the global explanations of continuous features generated by models corresponding to two extreme points and one ``middle point'' obtained from \textit{MONBM-CG}. In this paper, the ``middle point'' refers to a model that performs moderately on the trustworthy dimension, selected from all the knee points. The knee point represents the most concave region of the Pareto front and is considered the best trade-off for its neighborhood in the absence of specific preferences~\cite{zhang2014knee}. 

In this paper, we referred to the method of Zhang \textit{et al.}~\cite{zhang2014knee} for selecting knee points, where the parameter $r_g$ was set to 0.02. Taking the \textit{MONBM-CG} on the \textit{Adult} dataset as an example, the extreme points of \textit{CE} and \textit{GI} are at values of 0.1713 and 0.0303, respectively, on the \textit{GI} metric of the training set. Therefore, the ``middle point'' is the NBM where the \textit{GI} value is closest to the average of the two extremes on the \textit{GI} metric among all knee points, i.e., $(0.1713+0.0303)/2=0.1008$. 

\begin{figure*}[h]
    \centering
    \subfigure[Global explanation results for continuous features of \textit{MONBM-CG} instantiation on the \textit{Adult} dataset]{
        
        \includegraphics[width=\textwidth]{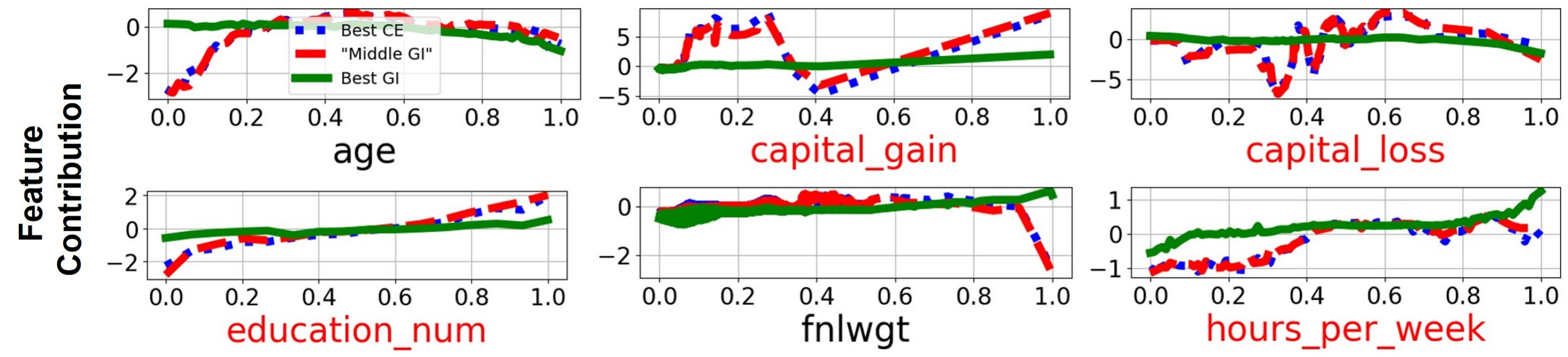}
        
        \label{fig:adult_GU_merge_global_con}
    }
    \subfigure[Global explanation results for continuous features of \textit{MONBM-CG} instantiation on the \textit{Bank} dataset]{
        \includegraphics[width=\textwidth]{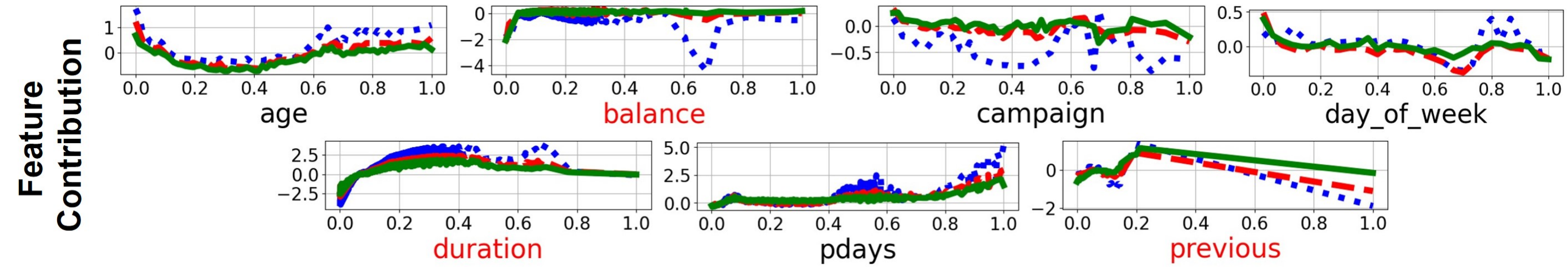}
        \label{fig:bank_GU_merge_global_con}
    }
    \caption{Comparison of global explanations of continuous features to validate the proposed smoothness and monotonicity metrics. For each dataset, the blue, red, and green lines represent the global explanation results of the continuous features by the model corresponding to the best \textit{CE}, ``middle \textit{GI},'' and best \textit{GI} points obtained by the \textit{MONBM-CG} instantiation in one arbitrary trial, respectively. Each subfigure corresponds to the global explanation of a feature, with the x-axis representing feature values and the y-axis showing the outputs of the shape functions. Each continuous feature has a smoothness requirement, and features highlighted in red also have a monotonicity requirement.}
    \label{fig:GU_merge_global_con}
\end{figure*}

Due to space limitations, we provided results for the \textit{Adult} and \textit{Bank} datasets, as shown in Fig.~\ref{fig:GU_merge_global_con}. The corresponding results for the remaining datasets are presented in Fig. S-3 of the \textit{Supplementary Material}. To further illustrate the continuity of the trade-off surface, Fig. S-4 in the \textit{Supplementary Material} presents the global explanations for the complete set of non-dominated knee-point solutions across all datasets.

Across all datasets, a consistent pattern is observed: as \textit{GI} improves, the learned shape functions become progressively smoother while preserving their primary functional trends. Unnecessary local oscillations and abrupt variations are gradually reduced, without collapsing the overall feature–target relationships. This demonstrates that the proposed framework provides a spectrum of balanced solutions that mitigate excessive fluctuations while maintaining meaningful structure in the explanations. In addition, improvements in \textit{GI} are accompanied by clearer monotonic tendencies in relevant features. Although strict smoothness or perfect monotonicity cannot be guaranteed for continuous functions learned by NN-based GAMs, the empirical results show that promoting approximate smoothness and monotonic trends is both feasible and effective in practice.

Overall, the explanation results demonstrate that our method can yield NBMs with high global interpretability. Moreover, these improvements in visual interpretability are consistently aligned with the observed changes in the \textit{GI} metric values across all datasets, as will be demonstrated in Table~\ref{tab:MONBM_CG_point}, further validating the effectiveness of the proposed metrics.

    \subsubsection{Analyzing the Relationship Between Dimensions (\textbf{RQ2})} To answer \textit{RQ2}, we analyzed each of the four instantiations in two steps: (1) selecting representative points to examine the relationship between dimensions based on the performance of the final generation on the training set; (2) exploring the experimental results using the global and local explanations provided by the obtained NBMs to further reveal the reasons behind these relationships.

For the two-objective instantiations, i.e., \textit{MONBM-CD}, \textit{MONBM-CL}, and \textit{MONBM-CG}, we chose two extreme points and one ``middle point'' for analysis in each case. For the four-objective instantiation \textit{MONBM-CDLG}, the analysis first focuses on the four extreme points to characterize the primary multidimensional trade-offs, and is then further extended to the full set of non-dominated solutions to provide a more comprehensive validation of these relationships.

\paragraph{\textbf{Relationship Between Accuracy and Fairness (MONBM-CD)}} On the \textit{MO\-NBM-CD} instantiation, Table~\ref{tab:MONBM-CD_point} shows the performance of the models corresponding to the extreme and ``middle \textit{DP}'' points on the test set (the performance on the training set is shown in Table S-V of the \textit{Supplementary Material}). From Table~\ref{tab:MONBM-CD_point}, we can draw the following two conclusions: (1) Moderate fairness improvement (defined as a 50\% reduction in the \textit{DP} metric relative to its extremes) has little effect on accuracy. Across six datasets, the impact of reducing the \textit{DP} metrics by half on the \textit{ACC} is only $0.48\%$ on average; (2) Pursuing extreme fairness significantly affects accuracy, with the degree of impact varying across datasets. For the \textit{Adult}, \textit{COMPAS}, and \textit{Dutch} datasets, reducing the \textit{DP} metric to zero leads to an average \textit{ACC} decrease of $17.06\%$, while the remaining three datasets show a much smaller average reduction of $2.15\%$.

\begin{table*}[h]
    \caption{Comparison of the mean values of the \textit{ACC} and \textit{DP} on the test set of the models corresponding to the three points chosen from the final solution set obtained by the \textit{MONBM-CD} instantiation over 15 trials. \textit{ACC} values are expressed as percentages ($\%$). The best average is highlighted with a gray background. The difference between it and the best value is shown in parentheses.}
    \label{tab:MONBM-CD_point}
    \centering
    \resizebox{\textwidth}{!}{
    \begin{tabular}{l|cc|cc|cc}
        \toprule
        & \multicolumn{2}{c|}{Adult} & \multicolumn{2}{c|}{Bank} & \multicolumn{2}{c}{COMPAS} \\
        & ACC ($\%$) $\uparrow$ & DP $\downarrow$ & ACC ($\%$) $\uparrow$ & DP $\downarrow$ & ACC ($\%$) $\uparrow$ & DP $\downarrow$ \\
        \midrule
        Best CE & \cellcolor{gray!40}86.36 & 0.1686 (+0.1685) & \cellcolor{gray!40}90.26 & 0.0357 (+0.0357) & \cellcolor{gray!40}68.13 & 0.2025 (+0.2023) \\
        ``Middle DP'' & 85.61 (-0.75) & 0.0851 (+0.0850) & 90.21 (-0.05) & 0.0176 (+0.0176) & 67.77 (-0.35) & 0.1041 (+0.1039) \\
        Best DP & 76.60 (-9.76) & \cellcolor{gray!40}0.0001 & 88.76 (-1.49) & \cellcolor{gray!40}0.0000 & 53.86 (-14.27) & \cellcolor{gray!40}0.0002 \\
        \bottomrule
        \toprule
        & \multicolumn{2}{c|}{Default} & \multicolumn{2}{c|}{Dutch} & \multicolumn{2}{c}{LSAT} \\
        & ACC ($\%$) $\uparrow$ & DP $\downarrow$ & ACC ($\%$) $\uparrow$ & DP $\downarrow$ & ACC ($\%$) $\uparrow$ & DP $\downarrow$ \\
        \midrule
        Best CE & \cellcolor{gray!40}81.82 & 0.0181 (+0.0181) & \cellcolor{gray!40}83.13 & 0.3318 (+0.3307) & \cellcolor{gray!40}90.00 & 0.1666 (+0.1653) \\
        ``Middle DP'' & 81.75 (-0.07) & 0.0093 (+0.0093) & 81.87 (-1.26) & 0.1672 (+0.1661) & 89.69 (-0.31) & 0.0838 (+0.0825) \\
        Best DP & 77.88 (-3.94) & \cellcolor{gray!40}0.0000 & 56.09 (-27.05) & \cellcolor{gray!40}0.0011 & 88.97 (-1.02) & \cellcolor{gray!40}0.0013 \\
        \bottomrule
    \end{tabular}}
\end{table*}

\begin{figure*}[h]
    \centering
    \subfigure[Global explanation results of \textit{MONBM-CD} instantiation on the \textit{COMPAS} dataset]{
        \includegraphics[width=\textwidth]{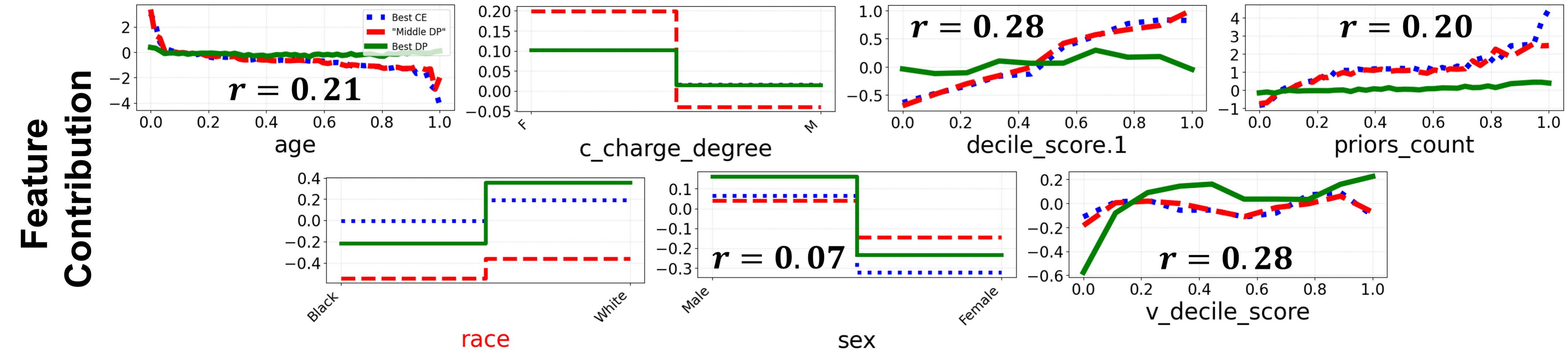}
        \label{fig:COMPAS_DP_global}
    }
    \subfigure[Global explanation results of \textit{MONBM-CD} instantiation on the \textit{LSAT} dataset]{
        \includegraphics[width=\textwidth]{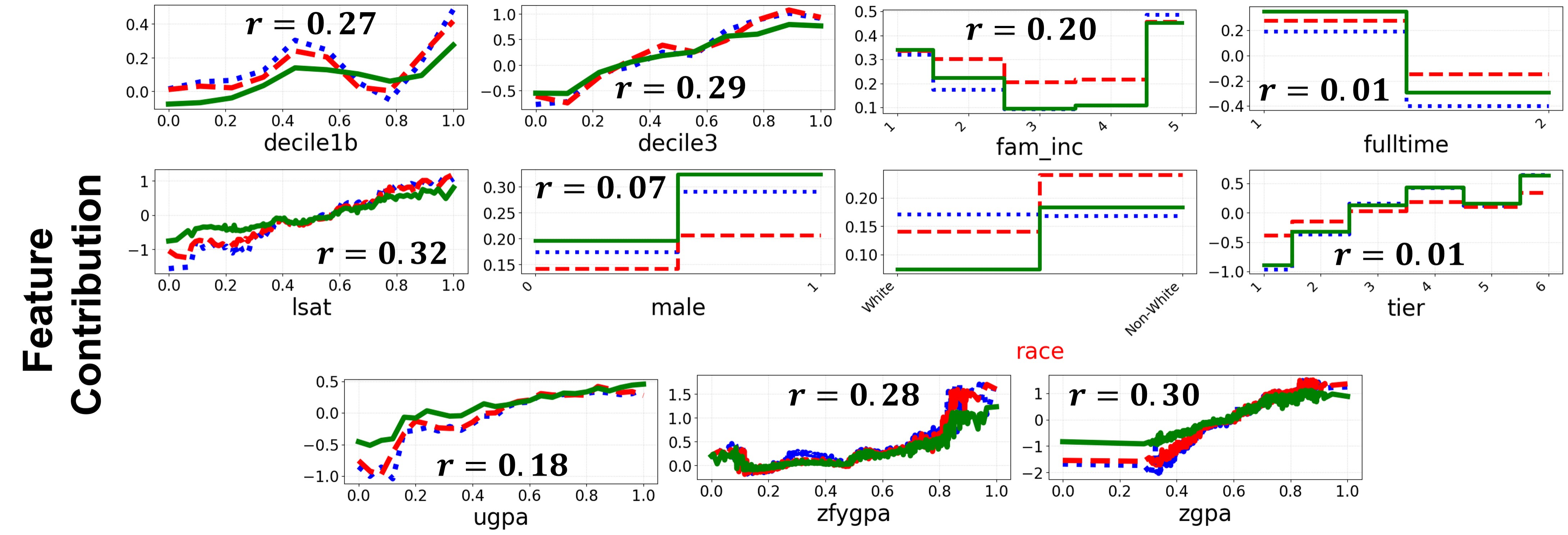}
        \label{fig:LSAT_DP_global}
    }
    \caption{Comparison of global explanations to assess the impact of improved fairness on each feature. For each dataset, the blue, red, and green lines represent the global explanation results of the model corresponding to the best \textit{CE}, ``middle \textit{DP}'', and best \textit{DP} points obtained by the \textit{MONBM-CD} instantiation in one arbitrary trial, respectively. Each subfigure corresponds to the global explanation of a feature, with the x-axis representing feature values and the y-axis showing the outputs of the shape functions. The sensitive attribute is highlighted in red. The Pearson correlation coefficient between each attribute and the sensitive attribute is annotated in the respective subplots.}
    \label{fig:DP_global_explanation}
\end{figure*}

To further analyze this phenomenon, we analyzed the global explanation results of the three models. Due to space limitations, we chose the \textit{COMPAS} and \textit{LSAT} datasets as representative of the significant and moderate effects of pursuing extreme fairness on accuracy, respectively, and their global explanation results are shown in Fig.~\ref{fig:DP_global_explanation}. The corresponding results for the remaining datasets are shown in Fig. S-5 of the \textit{Supplementary Material}. As can be seen in these figures, for each dataset, compared to the extreme point of accuracy (the blue line), moderate fairness enhancement (the red line) primarily influences the sensitive attribute, thus having a low impact on accuracy. However, the impact of further pursuing extreme fairness (the green line) on the features varies across datasets. For example, \textit{COMPAS} (Fig.~\ref{fig:COMPAS_DP_global}) has the vast majority of features affected in the pursuit of extreme fairness, thus significantly affecting accuracy. In contrast, \textit{LSAT} (Fig.~\ref{fig:LSAT_DP_global}) changes only a few features and, therefore, has little impact on accuracy. Furthermore, the figures indicate that to enhance fairness, many features are adjusted to contribute minimally to the decision outcome, effectively removing their influence and improving fairness. 

To examine the influence of training sample variability, we additionally report the mean global explanations together with their standard deviations across 15 independent trials, provided in Fig. S-6 of the \textit{Supplementary Material}. Overall, the functional shapes observed in the single-run results (Figs.~\ref{fig:DP_global_explanation} and S-5) remain highly consistent with the averaged statistics (Fig. S-6), indicating that the main explanatory patterns are stable across different training samples. However, while the overall trends are preserved, the magnitude of the shape functions exhibits variability at moderate and extreme fairness operating points. A more detailed analysis of this variability is provided in Section S-IX of the \textit{Supplementary Material}.

To further investigate the robustness of these findings, we extended our analysis to a label-aware fairness metric, equalized odds (EOD), in Section S-VIII of the \textit{Supplementary Material}. As shown in Table S-IX and Fig. S-2, the experimental results under \textit{EOD} exhibit highly consistent trade-off trends and global explanation patterns with those of \textit{DP}. This demonstrates that the decision logic adjustments identified by \textit{MONBM} are intrinsic to the accuracy-fairness conflict in these datasets, rather than being an artifact of a specific fairness definition.

Overall, consistent with findings in the existing literature, a trade-off exists between accuracy and fairness~\cite{zhang2022mitigating,zhang2024fairness}. We further revealed the underlying reasons. Specifically, moderate improvement primarily necessitates adjusting the impact of the sensitive attribute on decision results, with minimal effects on accuracy. Moreover, to achieve extreme fairness, there are differences across datasets; some require substantial adjustments to multiple features, leading to significant accuracy losses, while others necessitate fewer modifications, resulting in relatively minor accuracy declines.

\paragraph{\textbf{Relationship Between Accuracy and Local Interpretability (MO\-NBM-CL)}} On the \textit{MONBM-CL} instantiation, Table~\ref{tab:MONBM_CL_point} shows the performance of the models corresponding to the extreme and ``middle \textit{LI}'' points on the test set (the performance on the training set is shown in Table S-VI of the \textit{Supplementary Material}). From Table~\ref{tab:MONBM_CL_point}, we can draw the following conclusions: (1) similar to the fairness metric \textit{DP}, a moderate improvement in \textit{LI} (defined as a 50\% reduction in the \textit{LI} metric relative to its extremes) has a relatively low impact on \textit{ACC} ($2.03\%$ decrease on average), but pursuing extreme \textit{LI} significantly affects \textit{ACC} ($12.59\%$ decrease on average); (2) the impact of improving \textit{LI} on \textit{ACC} is significantly larger than that of improving the fairness metric \textit{DP} ($0.48\%$ and $9.61\%$ decrease on average, respectively). 

\begin{table}[htbp]
    \caption{Comparison of the mean values of the \textit{ACC} and \textit{LI} on the test set of the models corresponding to the three points chosen from the final solution set obtained by the \textit{MONBM-CL} instantiation over 15 trials. \textit{ACC} values are expressed as percentages ($\%$). The best average is highlighted with a gray background. The difference between it and the best value is shown in parentheses.}
    \label{tab:MONBM_CL_point}
    \centering
    \resizebox{\textwidth}{!}{
    \begin{tabular}{l|cc|cc|cc}
        \toprule
            & \multicolumn{2}{c|}{Adult} & \multicolumn{2}{c|}{Bank} & \multicolumn{2}{c}{COMPAS} \\
            & ACC ($\%$) $\uparrow$ & LI $\downarrow$ & ACC ($\%$) $\uparrow$ & LI $\downarrow$ & ACC ($\%$) $\uparrow$ & LI $\downarrow$ \\
            \midrule
            Best CE & \cellcolor{gray!40}86.36 & 1.0271 (+0.9171) & \cellcolor{gray!40}90.25 & 1.1649 (+1.0811) & \cellcolor{gray!40}68.29 & 0.7289 (+0.4403) \\
            ``Middle LI'' & 84.81 (-1.55) & 0.5720 (+0.4619) & 89.75 (-0.50) & 0.6292 (+0.5454) & 61.51 (-6.78) & 0.5102 (+0.2216) \\
            Best LI & 65.55 (-20.81) & \cellcolor{gray!40}0.1100 & 83.23 (-7.03) & \cellcolor{gray!40}0.0838 & 53.53 (-14.76) & \cellcolor{gray!40}0.2886 \\
        \bottomrule
        \toprule
            & \multicolumn{2}{c|}{Default} & \multicolumn{2}{c|}{Dutch} & \multicolumn{2}{c}{LSAT} \\
            & ACC ($\%$) $\uparrow$ & LI $\downarrow$ & ACC ($\%$) $\uparrow$ & LI $\downarrow$ & ACC ($\%$) $\uparrow$ & LI $\downarrow$ \\
            \midrule
            Best CE & \cellcolor{gray!40}81.83 & 1.1209 (+0.9697) & \cellcolor{gray!40}83.13 & 0.7287 (+0.6032) & \cellcolor{gray!40}89.95 & 0.9791 (+0.8415) \\
            ``Middle LI'' & 81.17 (-0.66) & 0.6339 (+0.4827) & 80.67 (-2.46) & 0.4224 (+0.2969) & 89.74 (-0.21) & 0.5596 (+0.4220) \\
            Best LI & 64.68 (-17.14) & \cellcolor{gray!40}0.1512 & 68.30 (-14.84) & \cellcolor{gray!40}0.1255 & 88.97 (-0.98) & \cellcolor{gray!40}0.1376 \\
        \bottomrule
    \end{tabular}}
\end{table}

\begin{figure*}[h]
    \centering
    \subfigure[Average local explanation of \textit{MONBM-CL} instantiation on the \textit{Bank} dataset]{
        \includegraphics[width=\textwidth]{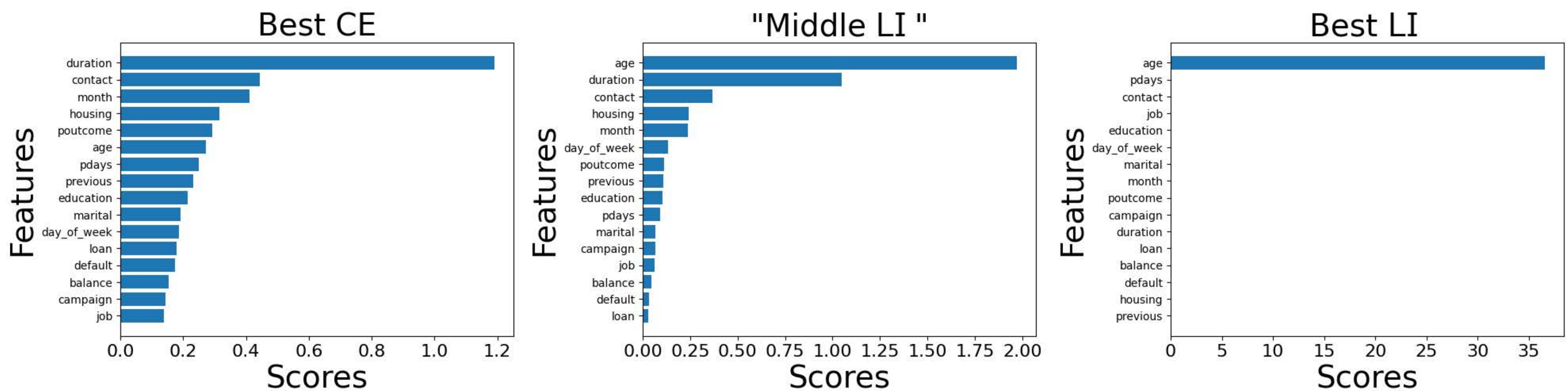}
        \label{fig:bank_LU_merge_local}
    }
    \subfigure[Average local explanation of \textit{MONBM-CD} instantiation on the \textit{Bank} dataset]{
        \includegraphics[width=\textwidth]{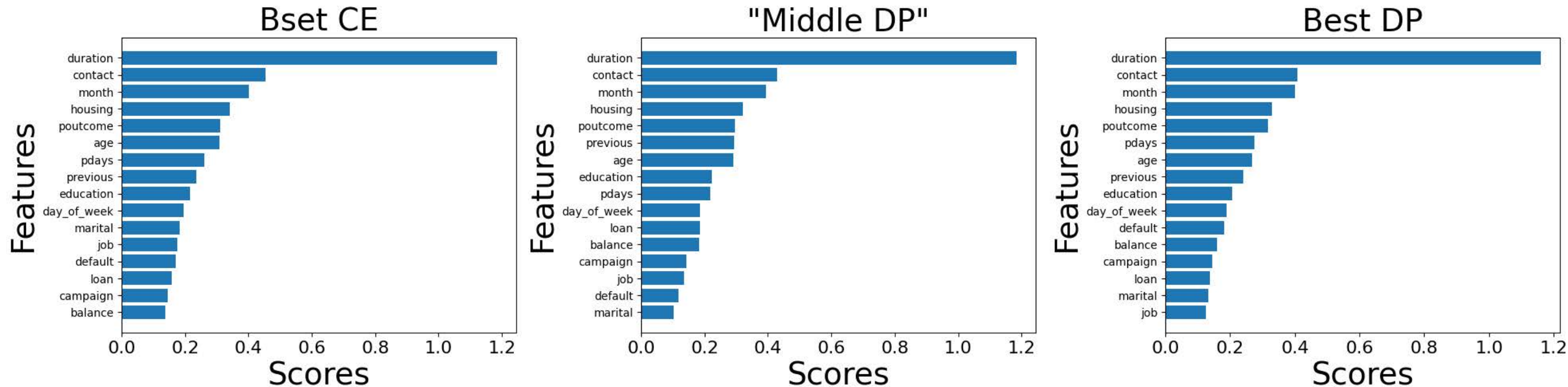}
        \label{fig:bank_DP_merge_local}
    }
    \caption{Comparison of the average local explanations of the test set to assess the impact of improved \textit{LI} on each feature. The three columns respectively display the average local explanation results of the model corresponding to: (1) the best \textit{CE} point, (2) the ``middle \textit{LI}'' or ``middle \textit{DP}'' point, and (3) the best \textit{LI} or best \textit{DP} point, obtained by the \textit{MONBM-CL} or \textit{MONBM-CD} instantiation on the \textit{Bank} dataset in one arbitrary trial. For each local explanation, the y-axis represents the features and the x-axis represents the average of the absolute values of the shape function outputs corresponding to the features for all data points in the test set.}
    \label{fig:LU_local_explanation}
\end{figure*}

To further analyze this phenomenon, we analyzed the local explanation results of the three models and compared them with the corresponding results from the \textit{MONBM-CD} instantiation. To ensure that the explanation results are representative, we present the average absolute values of local explanation results for all data points in the test set $\bm{\mathcal{D}}_{\textnormal{\textit{test}}}$. Due to space limitations, we selected the \textit{Bank} dataset as a representative, and the results are shown in Fig.~\ref{fig:LU_local_explanation}, while the results for other datasets are shown in Fig. S-7 in the \textit{Supplementary Material}. From Fig.~\ref{fig:bank_LU_merge_local} and the left column of Fig. S-7, it can be seen that compared with pursuing extreme \textit{ACC}, moderate \textit{LI} enhancement still allows most features to play a role in model decision-making, thus maintaining the model's predictive ability. In contrast, extreme \textit{LI} drastically weakens or removes the influence of most features, sharply reducing performance. 

Moreover, comparing with Fig.~\ref{fig:bank_DP_merge_local} and the right column of Fig. S-7, the \textit{LI} objective induces more pronounced shifts in relative feature importance than \textit{DP}. This pronounced disparity stems from the fundamental distinction between their underlying optimization mechanisms: whereas fairness constraints primarily refine the contributions of the sensitive attribute and its proxies, the \textit{LI} objective enforces structural sparsity across the entire feature set. As the model approaches the extreme \textit{LI} frontier, the framework increasingly favors a small number of highly informative features that can maintain predictive capability with minimal complexity. A representative example of this transition is observed in the \textit{Adult} dataset (Fig. S-7(a)), where the feature ``fnlwgt'' transitions from a relatively minor contributor in a multi-feature setting to a dominant predictor under strong sparsity constraints. Rather than indicating a limitation, this behavior helps reveal the hierarchy of feature importance and clarifies the performance limits of highly simplified models.

Furthermore, to evaluate the stability of these local explanations across different data partitions, we provide the mean local importance scores and their standard deviations over 15 independent trials in Fig. S-8 of the \textit{Supplementary Material}. While the overall feature rankings remain consistent with our single-run observations, we notice a non-negligible variance at the extreme points of fairness and local interpretability, which is further analyzed in Section S-IX of the \textit{Supplementary Material}.

Overall, there is a significant trade-off between accuracy and local interpretability. Moderate improvement preserves the model's reliance on most features, thereby preserving predictive power. However, pursuing extreme \textit{LI} requires removing most of the features' contribution to the model's decisions, significantly reducing the model's predictive power. Moreover, Fig.~\ref{fig:bank_LU_merge_local} and the left column of Fig. S-7 also visually illustrate that our method successfully produces NBMs with varying degrees of \textit{LI}.

 \paragraph{\textbf{Relationship Between Accuracy and Global Interpretability (MO\-NBM-CG)}} On the \textit{MONBM-CG} instantiation, Table~\ref{tab:MONBM_CG_point} shows the performance of the models corresponding to the extreme and ``middle \textit{GI}'' points on the test set (the performance on the training set is shown in Table S-VII of the \textit{Supplementary Material}). From Table~\ref{tab:MONBM_CG_point}, we observe that, unlike fairness and local interpretability, improving global interpretability has very little impact on accuracy. Moderate \textit{GI} improvement (defined as a 50\% reduction in the \textit{GI} metric relative to its extreme values) results in an average \textit{ACC} change of only 0.12\% across datasets, while pursuing extreme \textit{GI} leads to an average \textit{ACC} change of 0.85\%. Notably, on the \textit{Bank} and \textit{COMPAS} datasets, moderately or pursuing extreme \textit{GI} even slightly improves the \textit{ACC}. However, this does not imply that there is no trade-off between accuracy and global interpretability. We found that increasing \textit{GI} on both the \textit{Bank} and \textit{COMPAS} datasets leads to worse \textit{CE} metrics, even though it may bring about a slight improvement in \textit{ACC}. The reason behind this is more likely to be the non-strict linear relationships between \textit{CE} and \textit{ACC}, but it also suggests that improving \textit{GI} has a smaller impact on accuracy.

\begin{table*}[htbp]
    \caption{Comparison of the mean values of the \textit{ACC} and \textit{GI} on the test set of the models corresponding to the three points chosen from the final solution set obtained by the \textit{MONBM-CG} instantiation over 15 trials. \textit{ACC} values are expressed as percentages ($\%$). The best average is highlighted with a gray background. The difference between it and the best value is shown in parentheses.}
    \label{tab:MONBM_CG_point}
    \centering
    \resizebox{\textwidth}{!}{
    \begin{tabular}{l|cc|cc|cc}
        \toprule
            & \multicolumn{2}{c|}{Adult} & \multicolumn{2}{c|}{Bank} & \multicolumn{2}{c}{COMPAS} \\
            & ACC ($\%$) $\uparrow$ & GI $\downarrow$ & ACC ($\%$) $\uparrow$ & GI $\downarrow$ & ACC ($\%$) $\uparrow$ & GI $\downarrow$ \\
            \midrule
            Best CE & \cellcolor{gray!40}85.70 & 0.1651 (+0.1338) & 89.47 (-0.44) & 0.2212 (+0.1779) & 68.51 (-0.08) & 0.0129 (+0.0019) \\
            ``Middle GI'' & 85.57 (-0.13) & 0.1105 (+0.0792) & \cellcolor{gray!40}89.91 & 0.0791 (+0.0358) & \cellcolor{gray!40}68.59 & 0.0120 (+0.0009) \\
            Best GI & 81.01 (-4.69) & \cellcolor{gray!40}0.0313 & 89.65 (-0.26) & \cellcolor{gray!40}0.0433 & 68.19 (-0.40) & \cellcolor{gray!40}0.0111 \\
        \bottomrule
        \toprule
            & \multicolumn{2}{c|}{Default} & \multicolumn{2}{c|}{Dutch} & \multicolumn{2}{c}{LSAT} \\
            & ACC ($\%$) $\uparrow$ & GI $\downarrow$ & ACC ($\%$) $\uparrow$ & GI $\downarrow$ & ACC ($\%$) $\uparrow$ & GI $\downarrow$ \\
            \midrule
            Best CE & \cellcolor{gray!40}78.32 & 0.7045 (+0.6480) & \cellcolor{gray!40}82.54 & 0.0678 (+0.0627) & \cellcolor{gray!40}89.97 & 0.0314 (+0.0042) \\
            ``Middle GI'' & 78.08 (-0.24) & 0.3825 (+0.3261) & 81.70 (-0.84) & 0.0399 (+0.0348) & 89.95 (-0.02) & 0.0292 (+0.0020) \\
            Best GI & 78.31 (-0.01) & \cellcolor{gray!40}0.0564 & 82.32 (-0.22) & \cellcolor{gray!40}0.0051 & 89.95 (-0.02) & \cellcolor{gray!40}0.0272 \\
        \bottomrule
    \end{tabular}}
\end{table*}

\begin{figure*}[h]
    \centering
    \includegraphics[width=\textwidth]{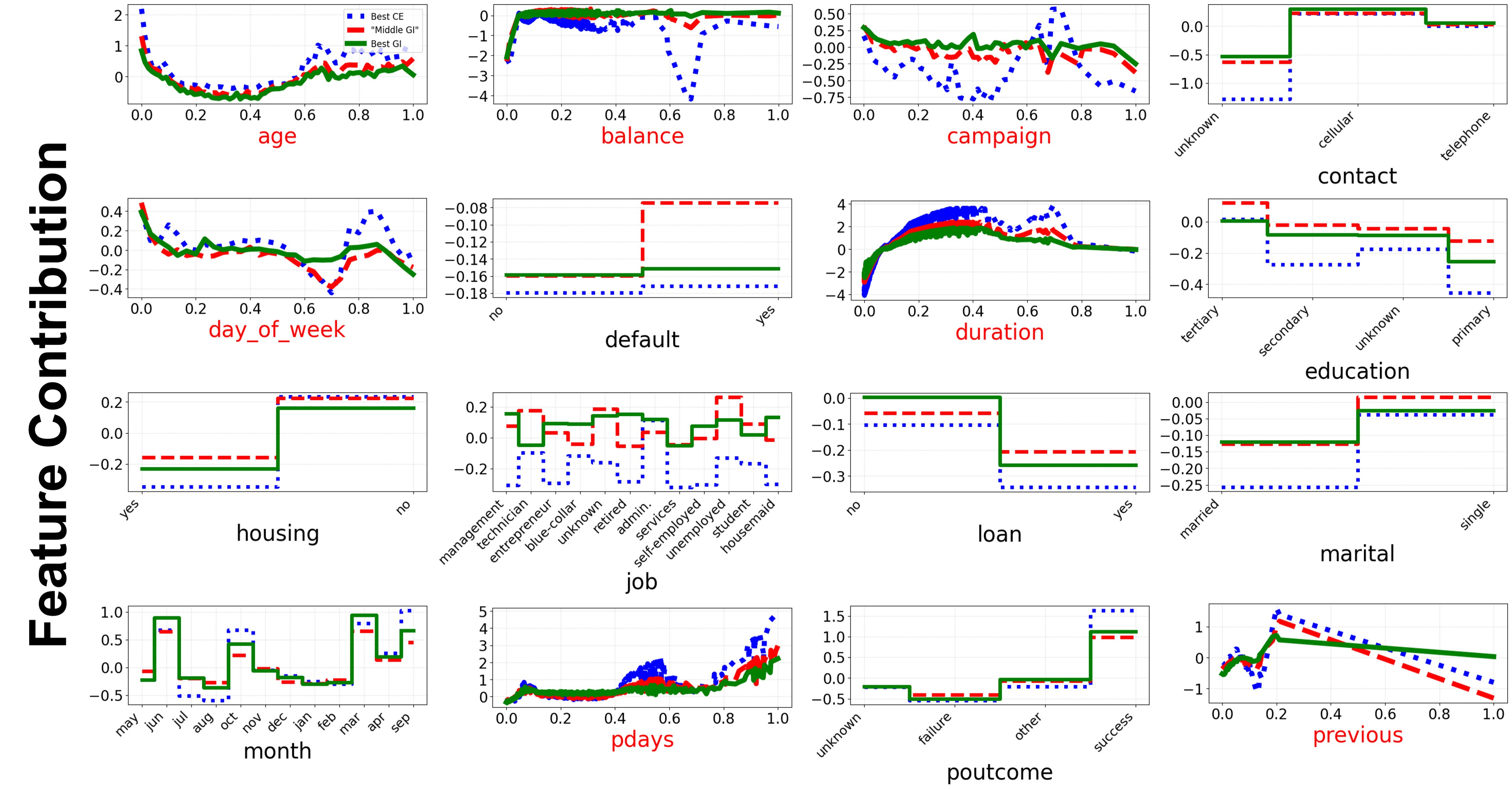}
    \caption{Comparison of global explanations to assess the impact of improved \textit{GI} on each feature. The blue, red, and green lines represent the global explanation results of the model corresponding to the best \textit{CE}, ``middle \textit{GI},'' and best \textit{GI} points obtained by the \textit{MONBM-CG} instantiation on the \textit{Bank} dataset in one arbitrary trial, respectively. Each subfigure corresponds to the global explanation of a feature, with the x-axis representing feature values and the y-axis showing the outputs of the shape functions. The continuous features are highlighted in red.}
    \label{fig:GU_global_explanation}
\end{figure*}

\begin{figure}[h]
    \centering
    \includegraphics[width=\textwidth]{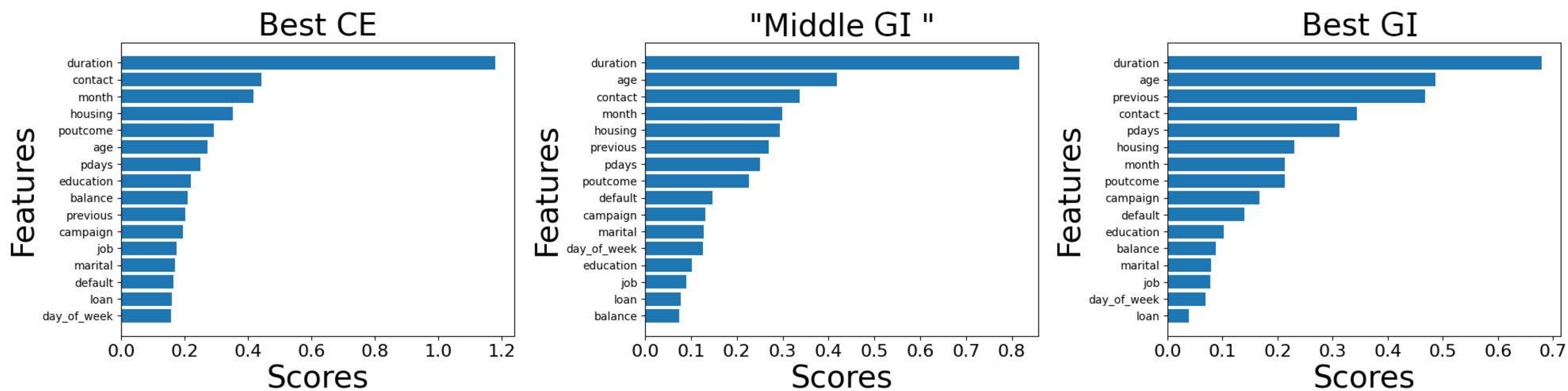}
    \caption{Comparison of the average local explanations of the test set to assess the impact of improved \textit{GI} on each feature. The three columns represent the average local explanation results of the model corresponding to the best \textit{CE}, ``middle \textit{GI},'' and best \textit{GI} points obtained by the \textit{MONBM-CG} instantiation on the \textit{Bank} dataset in one arbitrary trial, respectively. For each local explanation, the y-axis represents the features and the x-axis represents the average of the absolute values of the shape function outputs corresponding to the features for all data points in the test set.}
    \label{fig:GU_local_explanation}
\end{figure}

To further analyze this phenomenon, we presented the global and average local explanation results of the three corresponding models. To facilitate comparison with the \textit{MONBM-CL} and \textit{MONBM-CD} instantiations, we also selected the \textit{Bank} dataset as a representative, with its global and local explanation results shown in Figs.~\ref{fig:GU_global_explanation} and~\ref{fig:GU_local_explanation}, respectively, while the corresponding results for the other datasets are shown in Fig. S-9 and Fig. S-11 in the \textit{Supplementary Material}. 

From Fig.~\ref{fig:GU_local_explanation} and Fig. S-11, it can be seen that, compared with the extreme points of \textit{ACC} (first column), moderate \textit{GI} improvements (second column) and pursuit of extreme \textit{GI} (third column) have a significantly smaller impact on features than the corresponding results of \textit{MONBM-CL} and \textit{MONBM-CD} instantiations in Fig.~\ref{fig:LU_local_explanation} and Fig. S-7. Similar observations can also be seen in Fig.~\ref{fig:GU_global_explanation} and Fig. S-9, where the variation of the global explanation results is very small in each column. This difference can be attributed to the fact that, on the one hand, \textit{GI} has little effect on discrete features (i.e., features not highlighted in red). On the other hand, for continuous features (i.e., features highlighted in red), \textit{GI} only adjusts regions that are unsmoothed, mutated, or partially violate monotonicity. Consequently, the overall impact of global interpretability on accuracy is small. The stability of both global and local explanations for \textit{MONBM-CG} is verified in Figs. S-10 and S-12 of the \textit{Supplementary Material}, respectively. Compared with the \textit{MONBM-CD} and \textit{MONBM-CL} instantiations, \textit{MONBM-CG} exhibits lower variance in both global shape functions and local feature importance, even near extreme \textit{GI} operating points. Additional discussion of this phenomenon is provided in Section S-IX of the \textit{Supplementary Material}.

Overall, the trade-off between accuracy and global interpretability is weak. The reason behind this phenomenon is that \textit{GI} mainly affects the continuous features, especially by adjusting the regions that are not smooth or violate monotonicity, and these adjustments usually do not significantly change the overall predictive performance of the model. As a result, the performance trade-offs faced by the model in improving global interpretability are less pronounced than those faced in improving fairness or local interpretability.

\paragraph{\textbf{Relationship Between Accuracy, Fairness, Local and Global Interpretability (MONBM-CDLG)}} On the \textit{MONBM-CDLG} instantiation, Table~\ref{tab:MONBM_CDLG_point} shows the performance of the models corresponding to the extreme points on the test set (the performance on the training set is shown in Table S-VIII of the \textit{Supplementary Material}). This table confirms the prior conclusion that global interpretability has the least impact on accuracy, followed by fairness, while local interpretability has the greatest impact. More importantly, Table~\ref{tab:MONBM_CDLG_point} reveals that on the vast majority of datasets (especially \textit{Adult}, \textit{Bank}, \textit{COMPAS}, and \textit{LSAT}), improvements in fairness, local interpretability, or global interpretability individually are accompanied by improvements in fairness and local interpretability, especially fairness. For example, compared to the extreme points of \textit{ACC} (the first row of each dataset in Table~\ref{tab:MONBM_CDLG_point}), the pursuit of \textit{LI} (third row) also positively impacts the \textit{DP} metric. However, the \textit{Dutch} dataset is a notable exception; specifically, the pursuit of \textit{DP} instead worsens \textit{LI}, and the pursuit of \textit{GI} worsens both \textit{DP} and \textit{LI}. Furthermore, the pursuit of \textit{LI} significantly worsens the \textit{DP} metric by up to 0.928, indicating extreme unfairness.

\begin{table*}[h]
    \caption{Comparison of the mean values of the \textit{ACC}, \textit{DP}, \textit{LI}, and \textit{GI} on the test set of the models corresponding to the four extreme points chosen from the final solution set obtained by the \textit{MONBM-CDLG} instantiation over 15 trials. \textit{ACC} values are expressed as percentages ($\%$). The best average is highlighted with a gray background. The difference between it and the best value is shown in parentheses. The ``Majority Baseline'' row represents the accuracy of a constant classifier that consistently predicts the majority class for each dataset.}
    \label{tab:MONBM_CDLG_point}
    \centering
    \resizebox{\textwidth}{!}{
    \begin{tabular}{l|cccc|cccc}
        \toprule
            & \multicolumn{4}{c|}{Adult} & \multicolumn{4}{c}{Bank} \\
            & ACC ($\%$) $\uparrow$ & DP $\downarrow$ & LI $\downarrow$ & GI $\downarrow$ & ACC ($\%$) $\uparrow$ & DP $\downarrow$ & LI $\downarrow$ & GI $\downarrow$ \\
            \midrule
            Majority Baseline & 76.1 & --- & --- & --- & 88.3 & --- & --- & --- \\
            \midrule
            Best CE & \cellcolor{gray!40}85.7 & 0.174 (+0.17) & 1.026 (+0.92) & 0.172 (+0.14) & \cellcolor{gray!40}89.5 & 0.040 (+0.04) & 1.160 (+1.07) & 0.229 (+0.20)\\
            Best DP & 65.6 (-20.1) & \cellcolor{gray!40}0.000 & 0.124 (+0.02) & 0.155 (+0.12) & 88.3 (-1.20) & \cellcolor{gray!40}0.000 & 0.525 (+0.43) & 0.107 (+0.08) \\
            Best LI & 65.7 (-20.0) & 0.000 (+0.00) & \cellcolor{gray!40}0.108 & 0.167 (+0.13) & 83.2 (-6.30) & 0.000 (+0.00) & \cellcolor{gray!40}0.095 & 0.301 (+0.27) \\
            Best GI & 69.7 (-16.0) & 0.008 (+0.01) & 0.463 (+0.36) & \cellcolor{gray!40}0.036 & 88.3 (-1.20) & 0.001 (+0.00) & 0.751 (+0.66) & \cellcolor{gray!40}0.032 \\
        \bottomrule
        \toprule
            & \multicolumn{4}{c|}{COMPAS} & \multicolumn{4}{c}{Default} \\
            & ACC ($\%$) $\uparrow$ & DP $\downarrow$ & LI $\downarrow$ & GI $\downarrow$ & ACC ($\%$) $\uparrow$ & DP $\downarrow$ & LI $\downarrow$ & GI $\downarrow$ \\
            \midrule
            Majority Baseline & 54.5 & --- & --- & --- & 77.9 & --- & --- & --- \\
            \midrule
            Best CE & \cellcolor{gray!40}67.7 & 0.230 (+0.23) & 0.782 (+0.51) & 0.094 (+0.08) & \cellcolor{gray!40}78.4 & 0.024 (+0.02) & 1.212 (+1.07) & 0.706 (+0.65)\\
            Best DP & 51.8 (-15.9) & \cellcolor{gray!40}0.000 & 0.401 (+0.13) & 0.033 (+0.02) & 59.3 (-19.1) & \cellcolor{gray!40}0.000 & 0.524 (+0.38) & 0.241 (+0.18) \\
            Best LI & 52.2 (-15.5) & 0.055 (+0.06) & \cellcolor{gray!40}0.275 & 0.037 (+0.02) & 60.0 (-18.4) & 0.081 (+0.08) & \cellcolor{gray!40}0.143 & 0.563 (+0.51) \\
            Best GI & 60.4 (-7.30) & 0.072 (+0.07) & 0.767 (+0.49) & \cellcolor{gray!40}0.013 & 65.4 (-13.0) & 0.055 (+0.06) & 0.414 (+0.27) & \cellcolor{gray!40}0.057 \\
        \bottomrule
        \toprule
            & \multicolumn{4}{c|}{Dutch} & \multicolumn{4}{c}{LSAT} \\
            & ACC ($\%$) $\uparrow$ & DP $\downarrow$ & LI $\downarrow$ & GI $\downarrow$ & ACC ($\%$) $\uparrow$ & DP $\downarrow$ & LI $\downarrow$ & GI $\downarrow$ \\
            \midrule
            Majority Baseline & 52.4 & --- & --- & --- & 89.0 & --- & --- & --- \\
            \midrule
            Best CE & \cellcolor{gray!40}82.5 & 0.324 (+0.32) & 0.778 (+0.68) & 0.075 (+0.07) & \cellcolor{gray!40}90.0 & 0.187 (+0.19) & 1.109 (+0.96) & 0.174 (+0.14)\\
            Best DP & 50.5 (-32.0) & \cellcolor{gray!40}0.000 & 0.976 (+0.88) & 0.019 (+0.01) & 63.0 (-27.0) & \cellcolor{gray!40}0.000 & 0.171 (+0.02) & 0.171 (+0.14) \\
            Best LI & 66.1 (-16.4) & 0.928 (+0.93) & \cellcolor{gray!40}0.097 & 0.163 (+0.16) & 57.8 (-32.2) & 0.000 (+0.00) & \cellcolor{gray!40}0.154 & 0.187 (+0.16) \\
            Best GI & 70.5 (-12.0) & 0.406 (+0.41) & 0.681 (+0.58) & \cellcolor{gray!40}0.006 & 84.1 (-5.90) & 0.026 (+0.03) & 0.569 (+0.42) & \cellcolor{gray!40}0.029 \\
        \bottomrule
    \end{tabular}}
\end{table*}

\begin{figure*}[h!]
    \centering
    \subfigure[Average local explanation of \textit{MONBM-CDLG} instantiation on the \textit{Adult} dataset]{
        \includegraphics[width=\textwidth]{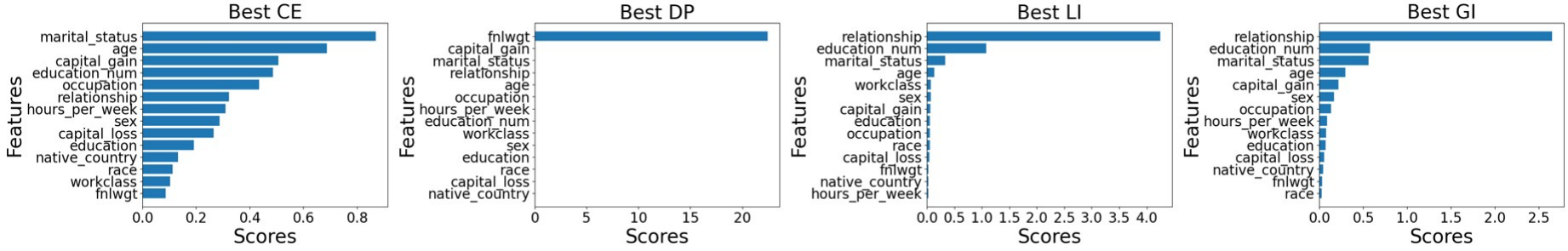}
        \label{fig:adult_4obj_merge_local}
    }
    \subfigure[Average local explanation of \textit{MONBM-CDLG} instantiation on the \textit{Dutch} dataset]{
        \includegraphics[width=\textwidth]{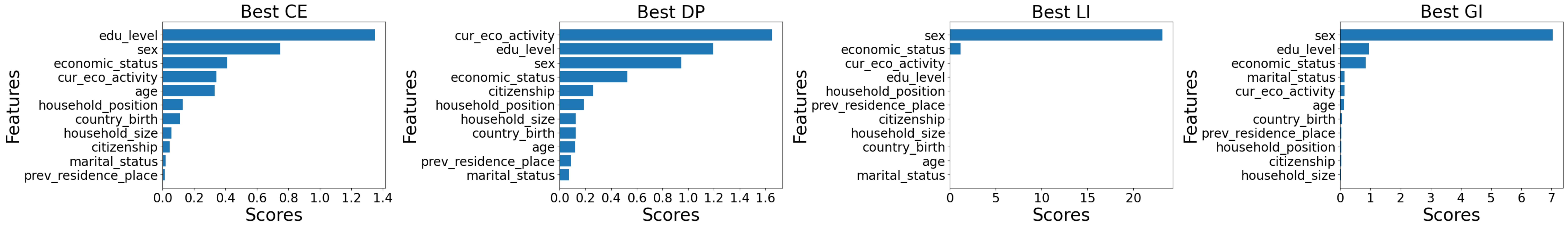}
        \label{fig:dutch_4obj_merge_local}
    }
    \caption{Comparison of the average local explanations of the test set to explore the interrelationships among the four objectives. For each dataset, the four columns represent the average local explanation results of the model corresponding to the best \textit{CE}, best \textit{DP}, best \textit{LI}, and best \textit{GI} points obtained by the \textit{MONBM-CDLG} instantiation in one arbitrary trial, respectively. For each local explanation, the y-axis represents the features and the x-axis represents the average of the absolute values of the shape function outputs corresponding to the features for all data points in the test set.}
    \label{fig:4obj_local_explanation}
\end{figure*}

\begin{figure*}[h]
    \centering
    \subfigure[Distribution of \textit{MONBM-CDLG} solutions on the \textit{Adult} dataset]{
        \includegraphics[width=\textwidth]{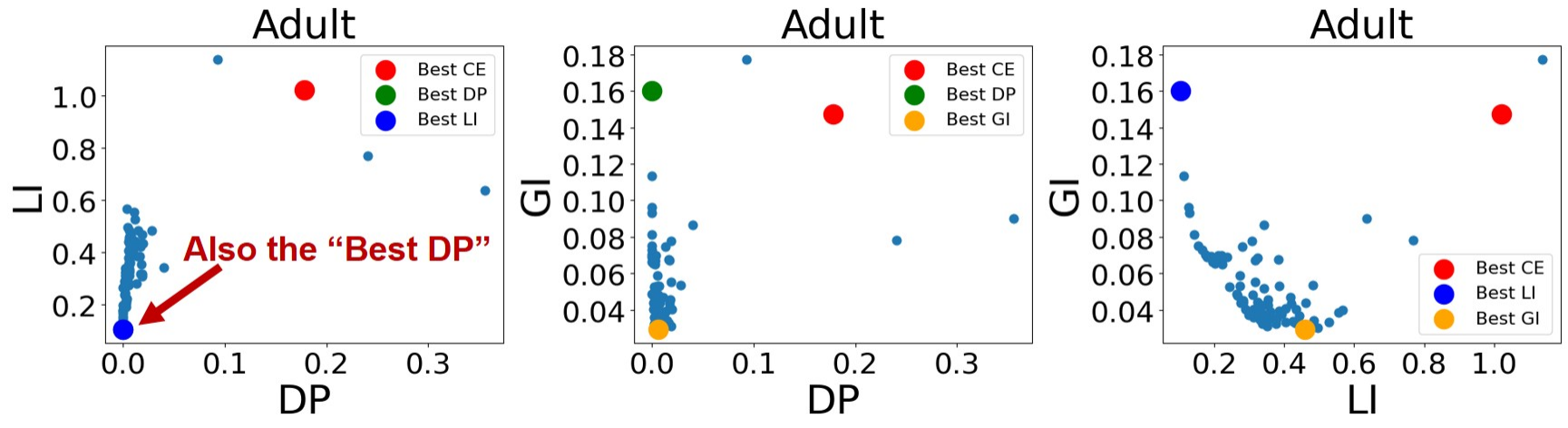}
        \label{fig:adult_CDLG_many_point}
    }
    \subfigure[Distribution of \textit{MONBM-CDLG} solutions on the \textit{Dutch} dataset]{
        \includegraphics[width=\textwidth]{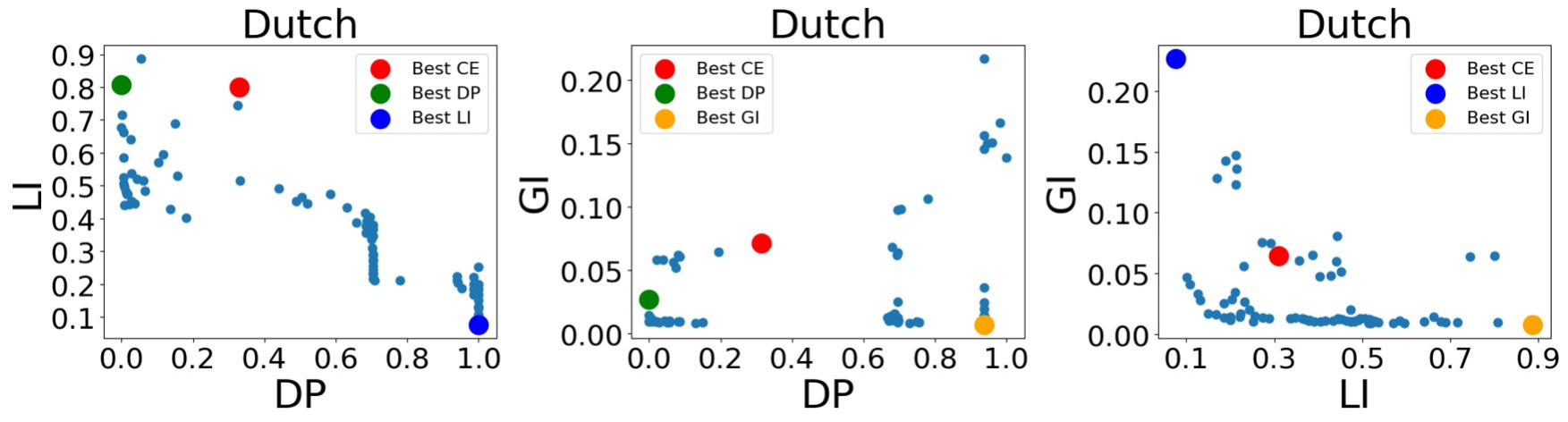}
        \label{fig:dutch_CDLG_many_point}
    }
    \caption{Distribution of all \textit{MONBM-CDLG} non-dominated solutions across pairs of trustworthy objectives to examine their interrelationships. Each subplot visualizes the solutions obtained from one arbitrary run of the \textit{MONBM-CDLG} instantiation on a given dataset, showing pairwise relationships between \textit{DP}, \textit{LI}, and \textit{GI}.}
    \label{fig:CDLG_many_point}
\end{figure*}

To further analyze this phenomenon, we compared the average local explanation results of the models corresponding to the four extreme points. Due to space limitations, we used the \textit{Adult} dataset as a general representative to compare with the \textit{Dutch} dataset, as shown in Fig.~\ref{fig:4obj_local_explanation}. The corresponding results for the remaining datasets are shown in Fig. S-13 of the \textit{Supplementary Material}. 

Beyond the extreme-point analysis, we further visualized the distribution of all non-dominated \textit{MONBM-CDLG} solutions across pairs of trustworthy dimensions (e.g., \textit{DP} vs. \textit{LI}, \textit{DP} vs. \textit{GI}, and \textit{LI} vs. \textit{GI}), as shown in Fig.~\ref{fig:CDLG_many_point} for \textit{Adult} and \textit{Dutch} (with additional datasets provided in Fig. S-15 of the \textit{Supplementary Material}). These distribution plots allow us to examine whether the relationships observed at the extreme points also hold for intermediate trade-off solutions and reduce the risk of conclusions driven by isolated extreme cases. From both the extreme-point analysis and the full-solution distributions, we observe the following:

\begin{itemize}
    \item[(1)] In general, improving \textit{DP} reduces the impact of many features on the decision result, consistent with the phenomenon observed on \textit{MONBM-CD} instantiation. This leads to an improvement in \textit{LI} since only a few features play an important role in the decision. Importantly, the distribution plots of all solutions confirm that this trend is not restricted to extreme solutions: on datasets such as \textit{Adult}, most Pareto solutions form a clear monotonic trend where lower \textit{DP} disparity is generally associated with better \textit{LI}. This suggests that the observed fairness–interpretability synergy reflects a systematic structural relationship rather than an artifact of isolated extreme models. In contrast, the \textit{Dutch} dataset retains many important features, resulting in no improvement in \textit{LI}.
    
    \item[(2)] Additionally, improving local interpretability reduces the impact of the vast majority of features on the decision and retains only a few important features. Thus, in most cases, the influence of the sensitive attribute (and its potential proxy attributes) on the decision is eliminated, thereby improving fairness. However, on the \textit{Dutch} dataset, the sensitive attribute ``sex'' happens to be retained as the only important feature in the pursuit of \textit{LI}, leading to decisions that significantly depend on this attribute and resulting in extreme unfairness. The full-solution distribution further supports this observation: unlike \textit{Adult}, the \textit{Dutch} solutions exhibit a clear negative trend between \textit{DP} and \textit{LI}, indicating a persistent trade-off across the Pareto set rather than only at extreme points. This highlights that fairness–interpretability relationships are strongly dataset-dependent and can reverse depending on feature–label correlations and sensitive attribute structure.
    
    \item[(3)]  Improving global interpretability—particularly smoothness—can moderately reduce the influence of certain features by removing irregular fluctuations in shape functions (as illustrated in Fig.~\ref{fig:GU_global_explanation} and Fig. S-9). This may indirectly support fairness and \textit{LI}. However, the distribution of all solutions shows that the improvement in \textit{LI} associated with better \textit{GI} is generally weak and scattered across datasets, indicating only partial alignment between the two interpretability notions. Conversely, improvements in \textit{LI} rarely lead to noticeable gains in \textit{GI}, confirming an asymmetric interaction. Overall, there is still a clear trade-off between the two. Similarly, improving global interpretability also helps fairness by reducing the influence of certain features, but enhancing fairness does not improve global interpretability. In addition, as can be seen in Fig.~\ref{fig:dutch_4obj_merge_local}, on the \textit{Dutch} dataset, \textit{GI} also encounters a similar situation where the sensitive attribute ``sex'' is taken as the most important feature, leading to more unfair decision-making results. 
    
    To better understand which component of \textit{GI} contributes to this phenomenon, we further decompose global interpretability into smoothness and monotonicity and conduct an ablation study in Section S-XI of the \textit{Supplementary Material}. The results suggest that the observed fairness–global interpretability conflict on the \textit{Dutch} dataset is primarily associated with the smoothness component, whereas the conflict between monotonicity and fairness is significantly weaker. This finding indicates that the relationship between global interpretability and other trustworthy dimensions is not uniform but depends on the specific interpretability mechanism being optimized.
\end{itemize}

Overall, our experimental results reveal the complex trade-offs between accuracy, fairness, local interpretability, and global interpretability. First, local interpretability has the greatest impact on accuracy, with extreme local interpretability often leading to a significant decrease in model performance, while moderate improvements in local interpretability have a minimal effect on accuracy. Second, there is a clear trade-off between fairness and accuracy. Moderate improvements in fairness have a negligible impact on accuracy, but extreme fairness often results in significant accuracy drops, especially on certain datasets, where fairness adjustments can cause substantial variations in performance. Global interpretability has the least impact on accuracy because it only affects the parts of the continuous features that do not satisfy smoothness and monotonicity.

Regarding the relationships between fairness, local interpretability, and global interpretability, our experiments show that these dimensions interact differently depending on the dataset characteristics. In most cases, increases in fairness, local interpretability, and global interpretability reduce the impact of many features on decision making, thus indirectly improving fairness and local interpretability. In contrast, the \textit{Dutch} dataset serves as a valuable counterexample that improving local and global interpretability leads to extremely unfair results, and improving fairness and global interpretability also does not benefit local interpretability, highlighting the context-dependent nature of these relationships. This complex interplay necessitates careful consideration by researchers aiming to optimize models while ensuring fairness and interpretability. To assess the stability of the explanations, the aggregate local explanation results across 15 trials are presented in Fig. S-14 of the \textit{Supplementary Material}, offering additional empirical support for the observed trade-off patterns.

We also note that four-objective instantiation (\textit{MONBM-CDLG}) generally performs worse than two-objective instantiations on the corresponding optimization objectives. This is a known challenge in many-objective evolutionary optimization: as the number of objectives increases, selection pressure weakens, and the search becomes less effective, often resulting in a lower-quality solution set—even with dedicated many-objective selection schemes. 

The reason is that maintaining a proper balance between exploration and convergence becomes increasingly difficult as more objectives are introduced. Therefore, introducing adaptive pressure mechanisms into the optimization process may provide a feasible way to improve search effectiveness. To explore this possibility, we conduct a preliminary investigation by incorporating an adaptive selection pressure strategy into the underlying stochastic ranking procedure, with detailed descriptions and empirical results provided in Section S-XII of the \textit{Supplementary Material}. Although the results indicate that adaptive pressure adjustment can partially mitigate performance degradation in high-dimensional objective spaces, further investigation of more specialized many-objective evolutionary algorithms and adaptive selection operators remains an important direction for future work.

\subsubsection{Comparison with Baseline Methods (\textbf{RQ3})} To validate that \textit{MONBM} solves the proposed problem well and is competitive with existing methods, we compared all the non-dominated solutions obtained by \textit{MONBM-CD} and \textit{MONBM-CL} instantiations with their respective baseline methods on the corresponding optimization objectives. Due to the absence of baseline methods for global interpretability, no comparisons are made for the \textit{MONBM-CG} instantiation. Furthermore, to comprehensively evaluate the framework's capability in addressing complex high-dimensional trade-offs, we present a synthetic evaluation of the \textit{MONBM-CDLG} instantiation against all baseline methods across all four objectives in Section S-XIII of the \textit{Supplementary Material}. This joint analysis, supplemented by visualizations of the complete Pareto fronts for the \textit{MONBM-CG} and \textit{MONBM-CDLG} instantiations in Figs. S-16 and S-17 of the \textit{Supplementary Material}, indicates that our framework produces non-dominated solution sets with strong diversity and competitive performance.

\paragraph{\textbf{MONBM-CD}} We compared the performance of \textit{MONBM-CD} and baseline methods on the test set $\bm{\mathcal{D}}_{\textnormal{\textit{test}}}$ in terms of \textit{ACC} and \textit{DP} metrics, specifically including: (1) comparing performance on \textit{HV}, \textit{MS}, and \textit{PD} metrics with \textit{FairEMOL}~\cite{zhang2022mitigating}; (2) comparing the dominance relationships with all baseline methods; and (3) visualizing the performance of the models obtained by each method.

We first compared the performance of \textit{MONBM-CD} and \textit{FairEMOL}~\cite{zhang2022mitigating} on the \textit{HV}, \textit{MS}, and \textit{PD} metrics, and the results are shown in Table~\ref{tab:FairEMOL_compare}. It is obvious from the table that our method has better convergence and diversity. More importantly, the models obtained by our method are interpretable.

\begin{table*}[h]
    \centering
    \caption{The average of 15 trials of \textit{HV}, \textit{MS}, and \textit{PD} values for the final model set. ``+/$\approx$/-'' indicates that the average \textit{FairEMOL} metric value is better/approximate/worse than the corresponding \textit{MONBM} value on the Wilcoxon rank-sum test, with a significance level of 0.05. The best average value is marked in gray.}
    \resizebox{\textwidth}{!}{
    \begin{tabular}{c|cc|cc|cc}
    \toprule
    & \multicolumn{2}{c|}{HV $\uparrow$} & \multicolumn{2}{c|}{MS $\uparrow$} & \multicolumn{2}{c}{PD $\uparrow$} \\
    & MONBM & FairEMOL & MONBM & FairEMOL & MONBM & FairEMOL \\
    \midrule 
    Adult & \cellcolor{gray!40}0.8849 (0.0183) & 0.8678 (0.0203) - & \cellcolor{gray!40}0.2597 (0.0164) & 0.2302 (0.0096) - & \cellcolor{gray!40}101910 (24450) & 230.48 (27.923) - \\
    Bank & \cellcolor{gray!40}0.9398 (0.0087) & 0.8678 (0.0103) - & \cellcolor{gray!40}0.0866 (0.0058) & 0.0568 (0.0029) - & \cellcolor{gray!40}26207 (7208.7) & 46.04 (9.5900) -\\
    COMPAS & \cellcolor{gray!40}0.6156 (0.0402) & 0.6065 (0.0411) - & 0.2458 (0.0443) & \cellcolor{gray!40}0.2938 (0.0443) + & \cellcolor{gray!40}75825 (21868) & 191.89 (49.975) -\\
    Default & \cellcolor{gray!40}0.8381 (0.0266) & 0.8321 (0.0272) - & 0.0457 (0.0067) & \cellcolor{gray!40}0.0465 (0.0128) $\approx$ & \cellcolor{gray!40}19365 (4974.1) & 30.611 (7.9945) -\\
    Dutch & \cellcolor{gray!40}0.9092 (0.0136) & 0.8951 (0.0160) - & \cellcolor{gray!40}0.8504 (0.3369) & 0.4417 (0.0097) - & \cellcolor{gray!40}266770 (62707) & 334.09 (77.768) -\\
    LSAT & \cellcolor{gray!40}0.8953 (0.0083) & 0.8912 (0.0082) - & \cellcolor{gray!40}0.1825 (0.0149) & 0.1461 (0.0349) - & \cellcolor{gray!40}29425 (7122.7) & 69.26 (13.564) -\\
    \midrule
    +/$\approx$/- & --- & 0/0/6 & --- & 1/1/4 & --- & 0/0/6 \\
    \bottomrule 
    \end{tabular}}
    \label{tab:FairEMOL_compare}
\end{table*}

Moreover, we compared the dominance relationships of \textit{MONBM-CD} with each baseline method in terms of \textit{ACC} and \textit{DP} metrics, as shown in Table~\ref{tab:CD_dominate_compare}. We also compared with the baseline methods in all four objectives, and the results are shown in Section S-XIII of the \textit{Supplementary Material}. It is worth noting that Table~\ref{tab:CD_dominate_compare} compares \textit{MONBM-CD} with single-solution-based (i.e., \textit{Fair-}, \textit{LFR-}, \textit{Re-}, \textit{ROC-}, \textit{LPP-}, \textit{Oracle-}, and \textit{FRAPP{\'E}-NBM}) and population-based baseline methods (i.e., \textit{FairEMOL}) using different dominant metrics. These follow the approaches in~\cite{zhang2022mitigating} and~\cite{wang2024multi} for set-to-point and set-to-set dominance evaluation, respectively. Details are given in Section S-VI of the \textit{Supplementary Material}. However, they have similar meanings. ``Do.'', ``Non-Do.'', and ``Be-Do.'' being larger means that our method has a higher probability of dominating, not dominating each other, and being dominated by the corresponding method, respectively. 

\setlength{\tabcolsep}{0.7pt}
\begin{table*}[h]
    \centering
    \caption{Dominance relationship between \textit{MONBM-CD} and other methods in terms of \textit{ACC} and \textit{DP} metrics. Results are averaged over 15 trials. ``Do.'', ``Non-Do.'', and ``Be-Do.'' mean the probability that \textit{MONBM-CD} dominates, does not dominate each other, and is dominated by the corresponding method. Dominance metrics for single-point and population-based baseline methods follow~\cite{zhang2022mitigating} and~\cite{wang2024multi}, respectively.}
    \resizebox{\textwidth}{!}{
    \begin{tabular}{c|ccc|ccc|ccc|ccc|ccc|ccc|ccc|ccc}
    \toprule
    & \multicolumn{3}{c|}{Fair-NBM} & \multicolumn{3}{c|}{LFR-NBM} & \multicolumn{3}{c|}{Re-NBM} & \multicolumn{3}{c|}{ROC-NBM} & \multicolumn{3}{c}{LPP-NBM} & \multicolumn{3}{c}{Oracle-NBM} & \multicolumn{3}{c}{FRAPP{\'E}-NBM} & \multicolumn{3}{c}{FairEMOL} \\
    & Do. & Non-Do. & Be-Do. & Do. & Non-Do. & Be-Do. & Do. & Non-Do. & Be-Do. & Do. & Non-Do. & Be-Do. & Do. & Non-Do. & Be-Do. & Do. & Non-Do. & Be-Do. & Do. & Non-Do. & Be-Do. & Do. & Non-Do. & Be-Do.\\
    \midrule 
    Adult & 0.53 & 0.97 & 0.01 & 0.93 & 0.66 & 0.00 & 0.33 & 0.95 & 0.03 & 1.00 & 0.00 & 0.00 & 0.13 & 0.84 & 0.14 & 0.07 & 0.95 & 0.04 & 0.07 & 0.86 & 0.14 & 0.40 & 0.60 & 0.00 \\
    Bank & 0.73 & 0.73 & 0.02 & 0.67 & 0.83 & 0.00 & 0.87 & 0.50 & 0.03 & 1.00 & 0.02 & 0.00 & 0.33 & 0.73 & 0.13 & 0.93 & 0.60 & 0.00 & 0.93 & 0.65 & 0.02 & 0.39 & 0.61 & 0.00 \\
    COMPAS & 0.53 & 0.74 & 0.00 & 0.93 & 0.50 & 0.00 & 0.73 & 0.69 & 0.05 & 0.87 & 0.93 & 0.00 & 0.40 & 0.71 & 0.19 & 0.87 & 0.68 & 0.02 & 0.73 & 0.75 & 0.01 & 0.17 & 0.83 & 0.00 \\
    Default & 0.80 & 0.62 & 0.02 & 0.80 & 0.51 & 0.00 & 0.73 & 0.59 & 0.05 & 1.00 & 0.77 & 0.00 & 0.60 & 0.55 & 0.17 & 0.67 & 0.63 & 0.07 & 0.73 & 0.63 & 0.04 & 0.32 & 0.68 & 0.00 \\
    Dutch & 0.00 & 0.96 & 0.04 & 0.80 & 0.82 & 0.01 & 0.40 & 0.98 & 0.01 & 1.00 & 0.75 & 0.00 & 0.20 & 0.94 & 0.05 & 0.00 & 0.96 & 0.04 & 0.00 & 0.94 & 0.06 & 0.13 & 0.87 & 0.00 \\
    LSAT & 1.00 & 0.83 & 0.00 & 0.80 & 0.86 & 0.02 & 0.20 & 0.90 & 0.07 & 0.00 & 0.97 & 0.03 & 0.40 & 0.80 & 0.10 & 0.47 & 0.89 & 0.05 & 0.27 & 0.74 & 0.16 & 0.05 & 0.90 & 0.05 \\
    \midrule
    \midrule
    Mean & 0.60 & 0.81 & 0.01 & 0.82 & 0.70 & 0.04 & 0.54 & 0.77 & 0.04 & 0.81 & 0.57 & 0.01 & 0.34 & 0.76 & 0.13 & 0.50 & 0.79 & 0.04 & 0.46 & 0.76 & 0.07 & 0.24 & 0.75 & 0.01 \\
    \bottomrule 
    \end{tabular}}
    \label{tab:CD_dominate_compare}
\end{table*}

From Table~\ref{tab:CD_dominate_compare}, we observe that (1) \textit{MONBM-CD} has quite a few solutions that can dominate other methods, achieving better performance in terms of both accuracy and fairness; (2) it also yields a wide range of mutually non-dominating solutions, serving as a valuable complement to existing methods; (3) apart from \textit{LPP-NBM}, the probability of \textit{MONBM-CD} being dominated by other methods remains very low. Although \textit{LPP-NBM} may dominate some solutions generated by \textit{MONBM-CD}, our method shows a higher probability of producing solutions that dominate those of \textit{LPP-NBM}, further demonstrating its competitive advantage.

\begin{figure*}[h]
    \centering
    \includegraphics[width=\textwidth]{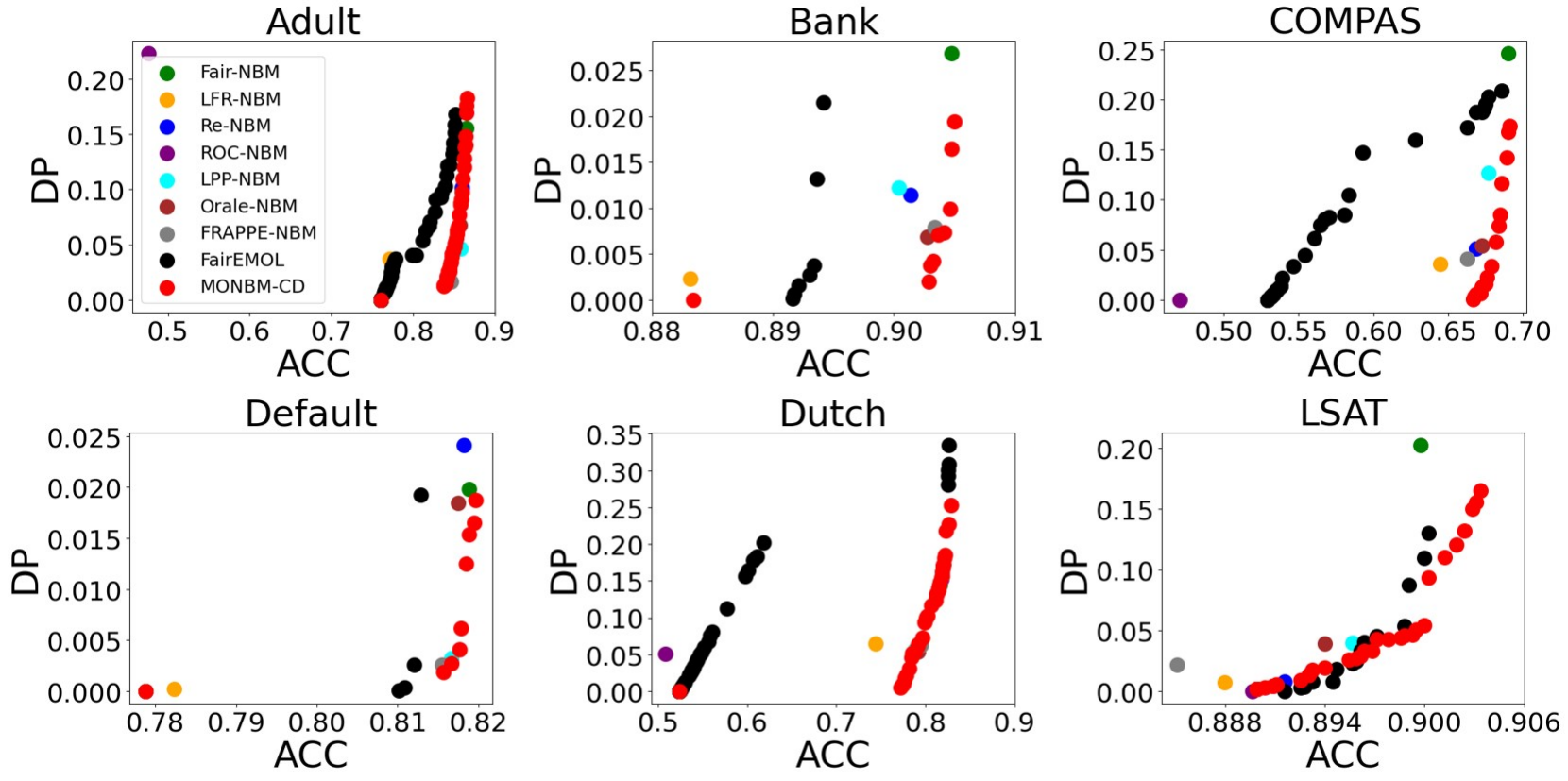}
    \caption{Comparison of the performance of \textit{MONBM-CD} and baseline methods on model accuracy and fairness metrics on one arbitrary trial.}
    \label{fig:test_BCE_DP_distribution_result}
\end{figure*}

Finally, we visualized the performance of the models obtained by each method, as shown in Fig.~\ref{fig:test_BCE_DP_distribution_result}. Notably, the performance of \textit{ROC-NBM} on the \textit{Bank} and \textit{Default} datasets is too poor to be displayed. Fig.~\ref{fig:test_BCE_DP_distribution_result} reveals that \textit{Fair-NBM} has limited improvement in fairness, while \textit{ROC-NBM} has obvious compromises in accuracy. \textit{LFR-NBM} and \textit{Re-NBM} can achieve a compromise solution but lack competitiveness. \textit{LPP-NBM}, \textit{Oracle-NBM}, and \textit{FRAPP{\'E}-NBM} can obtain a single solution with both good accuracy and fairness. In contrast, \textit{MONBM-CD} and \textit{FairEMOL} obtain a set of models with different trade-offs. Moreover, our method is interpretable and much closer to the bottom right corner of the plot, indicating better accuracy and fairness. This demonstrates the advantages of our approach.

\paragraph{\textbf{MONBM-CL}} Similarly, we compared the performance of \textit{MONBM-CL} and baseline methods on the test set $\bm{\mathcal{D}}_{\textnormal{\textit{test}}}$ in terms of \textit{ACC} and \textit{LI} metrics, specifically including: (1) comparing the dominance relationship with all baseline methods and (2) visualizing the performance of the models obtained by each method. We also compared the mutual dominance relationships with the baseline methods for all four objectives in Section S-XIII of the \textit{Supplementary Material}.

\begin{figure*}[htbp]
    \centering
    \includegraphics[width=\textwidth]{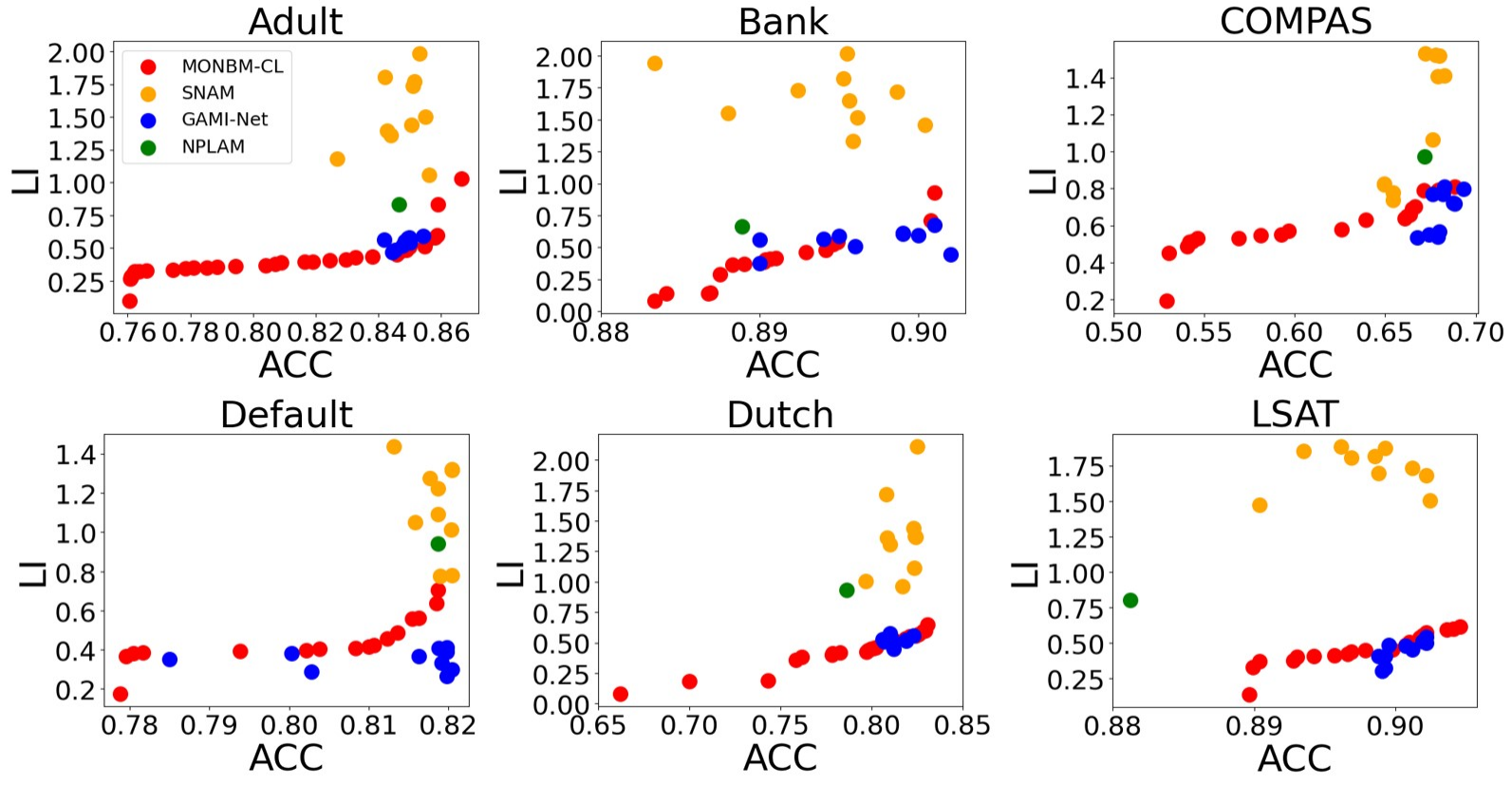}
    \caption{Comparison of the performance of \textit{MONBM-CL} and baseline methods on model accuracy and \textit{LI} metrics on one arbitrary trial.}
    \label{fig:test_BCE_LU_distribution_result}
\end{figure*}

For \textit{SNAM} and \textit{GAMI-Net}, we controlled the degree of feature selection by adjusting their hyper-parameters to generate ten models with different degrees of \textit{LI}. The dominance relationships between \textit{MONBM-CL} and the baseline methods in terms of \textit{ACC} and \textit{LI} metrics are shown in Table~\ref{tab:CL_dominate_compare}. From Table~\ref{tab:CL_dominate_compare}, it can be seen that \textit{MONBM-CL} significantly outperforms \textit{SNAM} and \textit{NPLAM}. However, the comparison with \textit{GAMI-Net} varies across datasets: \textit{MONBM-CL} outperforms \textit{GAMI-Net} on the \textit{Adult}, \textit{Bank}, \textit{COMPAS}, and \textit{Dutch} datasets while underperforming on the \textit{Default} and \textit{LSAT} datasets. Overall, \textit{MONBM-CL} is more advantageous. Fig.~\ref{fig:test_BCE_LU_distribution_result} further corroborates the above conclusion. 

\begin{table}[h]
    \centering
    \caption{Dominance relationship between \textit{MONBM-CL} and other methods in terms of \textit{ACC} and \textit{LI} metrics. Results are averaged over 15 trials. ``Do.'', ``Non-Do.'', and ``Be-Do.'' mean the probability that \textit{MONBM-CL} dominates, does not dominate each other, and is dominated by the corresponding method. Dominance metrics for all baseline methods follow~\cite{zhang2022mitigating}.}
    \begin{tabular}{c|ccc|ccc|ccc}
    \toprule
    & \multicolumn{3}{c|}{SNAM} & \multicolumn{3}{c}{GAMI-Net} & \multicolumn{3}{c}{NPLAM}\\
    & Do. & Non-Do. & Be-Do. & Do. & Non-Do. & Be-Do. & Do. & Non-Do. & Be-Do.\\
    \midrule 
    Adult & 1.000 & 0.788 & 0.000 & 0.573 & 0.914 & 0.316 & 0.933 & 0.775 & 0.000 \\
    Bank & 0.967 & 0.541 & 0.000 & 0.433 & 0.793 & 0.108 & 1.000 & 0.412 & 0.000 \\
    COMPAS & 0.953 & 0.659 & 0.000 & 0.507 & 0.779 & 0.110 & 0.733 & 0.657 & 0.004 \\
    Default & 0.653 & 0.831 & 0.005 & 0.080 & 0.374 & 0.612 & 0.600 & 0.861 & 0.006 \\
    Dutch & 0.993 & 0.586 & 0.000 & 0.773 & 0.854 & 0.007 & 1.000 & 0.573 & 0.000 \\
    LSAT & 0.929 & 0.608 & 0.000 & 0.207 & 0.706 & 0.263 & 0.600 & 0.710 & 0.014 \\
    \midrule
    \midrule
    Mean & 0.916 & 0.669 & 0.001 & 0.429 & 0.737 & 0.236 & 0.811 & 0.660 & 0.004 \\
    \bottomrule 
    \end{tabular}
    \label{tab:CL_dominate_compare}
\end{table}

More importantly, since the three baseline methods are not specifically designed to improve \textit{LI}, \textit{MONBM-CL} presents two major additional advantages: (1) The baseline methods exhibit high stochasticity, making it challenging to control \textit{LI} performance by adjusting the degree of feature selection. For example, stronger feature selection may instead result in a model that performs worse on \textit{LI}. In contrast, \textit{MONBM-CL} generates a set of models with different trade-offs between accuracy and \textit{LI} in a single run to choose from; (2) Compared to our method, the three baseline methods have difficulty in obtaining solutions with low \textit{LI}, which can also be seen in Fig.~\ref{fig:test_BCE_LU_distribution_result}.

Overall, the comparison with existing methods validates the effectiveness of our method in producing models with competitive performance across accuracy, fairness, and interpretability. Beyond numerical performance, our method also provides a diverse set of solutions. As illustrated in Figs.~\ref{fig:test_BCE_DP_distribution_result} and~\ref{fig:test_BCE_LU_distribution_result}, and further demonstrated by the Pareto front visualizations for the \textit{MONBM-CG} and \textit{MONBM-CDLG} instantiations in Figs. S-16 and S-17 of the \textit{Supplementary Material}, our framework exhibits excellent diversity in the objective space. This enables stakeholders to select models that offer distinct and controllable trade-offs across all optimized dimensions based on application-specific priorities—an advantage that single-point baselines inherently lack.

\subsection{Case Study: Threshold-Based Model Selection}

To illustrate how stakeholders may identify a deployable solution from a broad Pareto front with diverse trade-offs, we present a case study based on a single run of the \textit{Adult} dataset using the \textit{MONBM-CD} instantiation. Starting from the initial population of 100 solutions generated by \textit{MONBM}, we first extracted the non-dominated Pareto front, resulting in 43 candidate solutions. We then further selected knee points from this Pareto set. Knee points correspond to regions with relatively strong local trade-off characteristics and, in the absence of explicit preferences, are commonly regarded as representative compromise solutions within their neighborhoods~\cite{zhang2014knee}. Following this procedure, 13 knee-point solutions were retained as the candidate set for subsequent decision making.

Next, to simulate a realistic deployment scenario, we assume that a stakeholder specifies two threshold constraints according to operational requirements: (i) prediction accuracy must remain within 2\% of the highest accuracy achieved among the candidate solutions; and (ii) the \textit{DP} metric must remain below 0.05 to satisfy fairness requirements. Applying these constraints progressively filters the candidate set and ultimately retains two feasible deployment candidates:

\begin{itemize}
    \item \textit{Model 0}: $\text{ACC} = 86.21\%$, $\text{DP} = 0.0433$
    \item \textit{Model 1}: $\text{ACC} = 85.13\%$, $\text{DP} = 0.0302$
\end{itemize}

The final deployment decision can then be made according to application-specific priorities. For example, a stakeholder emphasizing predictive performance may prefer \textit{Model 0}, whereas scenarios with stricter fairness requirements may consider \textit{Model 1} more appropriate. Importantly, this case study is intended only to illustrate a simple a posteriori decision-making process rather than prescribe a universal selection strategy.

Overall, this example shows that the solution set generated by \textit{MONBM} can be progressively narrowed through the screening of knee points and intuitive threshold constraints, allowing stakeholders to efficiently move from a large Pareto set to a small set of practically deployable candidates while preserving flexibility in final decision making.

\section{Discussion}
\label{sec:discussion}

This section aims to clarify the scope of the application and the limitations of the conclusions obtained in this paper, especially those exploring the relationships between dimensions. In terms of interpretability, this paper employs NN-based GAM as a self-interpretable model to explore the relationship between its interpretability and other dimensions. The corresponding conclusions can be generalized to other GAMs, which also need to consider the interpretability of global and local explanations~\cite{zhong2024exploring,maindonald2010smoothing}. Further, widely recognized self-interpretable models include traditional linear models, decision trees, and decision rules. The interpretability of the explanations they provide depends on the number of terms in the linear model, the depth of the decision tree, and the number of decision rules, respectively~\cite{arrieta2020explainable,linardatos2020explainable}. This is very similar to the local interpretability of NN-based GAMs considered in this paper. Therefore, the conclusions considering the relationship between interpretability and other dimensions in these models can be referred to the corresponding conclusions of this paper on local interpretability.

Regarding fairness, this paper utilizes the widely used group fairness metric \textit{DP} to assess the fairness of the models. However, researchers have proposed numerous metrics to assess the fairness of ML models, which portray different aspects of fairness~\cite{zhang2022mitigating}. Furthermore, existing studies have categorized these metrics, grouping highly correlated metrics into the same category~\cite{anahideh2021choice,zhang2022mitigating}. Therefore, we believe that the corresponding conclusions obtained can be extended to fairness metrics within the same category, including two other widely used fairness metrics, equal opportunity (EOP) and \textit{EOD}, as well as metrics like predictive equality (PE), discovery difference (DD). To empirically support this generalizability, we conducted a sensitivity analysis on \textit{EOD} in Section S-VIII of the \textit{Supplementary Material}. The results indicate that the observed trade-off patterns and decision-logic adjustments under \textit{EOD} are largely consistent with those obtained using \textit{DP}. It should be acknowledged that the conclusions presented here require further validation when applied to other categories of fairness metrics. Nonetheless, they apply to three widely used fairness metrics: \textit{DP}, \textit{EOP}, and \textit{EOD}. In the future, we plan to consider accuracy, interpretability, and multiple fairness metrics simultaneously to analyze their relationships.

Furthermore, we clarify that the current instantiation of the \textit{MONBM} framework is designed specifically for tabular datasets, where each feature has a distinct and relatively independent semantic meaning. In contrast, for high-dimensional structured data such as images, the interpretation of univariate shape functions becomes less straightforward because individual input units (e.g., pixels) typically do not carry meaningful semantics in isolation and are strongly influenced by spatial redundancy and translation invariance (e.g., the same visual pattern appearing at different positions within a region of interest). As a result, the current \textit{MONBM} instantiation—and more broadly, standard GAM formulations operating directly on input features—is not directly applicable to raw pixel-level representations, since interpreting isolated pixel values provides limited explanatory value without surrounding context. Nevertheless, this limitation does not prevent its extension to structured modalities. As demonstrated in the original NBM framework~\cite{radenovic2022neural}, structured inputs can first be transformed into high-level semantic concepts through a concept extraction stage, allowing additive basis functions to operate on human-interpretable representations rather than raw inputs. Since \textit{MONBM} follows a model-agnostic multi-objective optimization paradigm, such concept-level additive architectures may provide a feasible pathway for extending the framework beyond tabular data. In this setting, the optimization objectives could remain unchanged while the underlying interpretable representation is adapted to the target modality. Exploring and empirically validating such extensions on large-scale structured datasets remains an important direction for future work.

Beyond these scope considerations, we also examined the phenomenon of overfitting. In our experiments for \textit{RQ2}, the selected solutions showed consistent results between their performance on the training set (Tables S-V-S-VIII in the \textit{Supplementary Material}) and on the test set (Tables~\ref{tab:MONBM-CD_point}–\ref{tab:MONBM_CDLG_point}), indicating that overfitting was limited. Nonetheless, non-dominance inconsistencies may still occur. In future work, we plan to use a validation set to mitigate this risk further.

\section{Conclusion}
\label{sec:conclusion}

In this paper, we introduced two explicit quantitative interpretability metrics, smoothness and monotonicity, specifically designed for NN-based GAMs and experimentally confirmed that they guide the model toward desirable global explanations. We then formulated the construction of accurate, fair, and interpretable GAMs as a new multi-objective learning problem and proposed the \textit{MONBM} framework, the first to bring multi-objective evolutionary learning to GAMs while jointly optimizing accuracy, fairness, and both local and global interpretability. Our instantiation retrains only the ``feature specific'' layer of a pre-trained NBM, making EMO practical for large-scale networks.

Using \textit{MONBM}, we reveal the complex relationships among the dimensions and the reasons behind them. Among them, improvements in fairness, local interpretability, and global interpretability each result in trade-offs with accuracy, though to varying extents. Moreover, the relationships between fairness, local interpretability, and global interpretability are complex, sometimes complementary, and sometimes conflicting, depending on the particular dataset and problem characteristics. These findings also illustrate how EMO can be applied to self-interpretable models to explore and interpret the complex relationships among multiple objectives, thereby offering a new perspective to analyze the trade-offs between various trustworthiness objectives. Finally, the effectiveness of the proposed metrics and the competitiveness of \textit{MONBM} are verified by visualizing the explanation results and comparing them with baseline methods.

In the future, we plan to (1) systematically investigate non-uniform smoothness control by integrating localized expert knowledge and adaptive weighting strategies, (2) validate the drawn conclusions on additional self-interpretable models and different categories of fairness metrics, (3) conduct further empirical studies to evaluate the framework's scalability and performance on high-dimensional problems as more suitable datasets with sensitive attributes become available, and (4) explore potential trade-off relationships with other trustworthy AI dimensions, such as robustness and security.

\section*{Data availability}

We used publicly available/open-source datasets for our experiments. The details and access information for these datasets are provided in the manuscript.

\section*{CRediT authorship contribution statement}

\textbf{Ziming Wang:} Writing – original draft, Methodology, Data curation, Validation; \textbf{Changwu Huang:} Conceptualization, Writing – review \& editing, Supervision, Funding acquisition; \textbf{Ke Tang:} Writing – review \& editing, Supervision, Funding acquisition; \textbf{Yew-Soon Ong:} Writing – review \& editing, Supervision, Funding acquisition; \textbf{Xin Yao:} Project administration, Resources, Writing – review \& editing, Supervision, Funding acquisition;

\section*{Declaration of competing interest}

The authors declare that they have no known competing financial interests or personal relationships that could have appeared to influence the work reported in this paper.

\section*{Acknowledgements}

This work was supported by the National Natural Science Foundation of China (Grant No. 62250710682), the Guangdong Major Project of Basic and Applied Basic Research (Grant No. 2023B0303000010), the Guangdong Provincial Key Laboratory (Grant No. 2020B121201001), BNBU Research Grant with No. of R0700158-26 at Beijing Normal-Hong Kong Baptist University, internal grants of Lingnan University, and the IEEE Computational Intelligence Society Graduate Student Research Grant 2024.

\bibliographystyle{elsarticle-num} 
\bibliography{ref}

\renewcommand\thesection{S-\Roman{section}}
\renewcommand\thetable{S-\Roman{table}}
\renewcommand\thefigure{S-\arabic{figure}}

\setcounter{figure}{0}
\setcounter{table}{0}
\setcounter{section}{0}

\section{Dataset Under Consideration}

In this paper, we consider six datasets that are widely used in the field of fairness~\cite{le2022survey}, with specific information shown in Table~\ref{tab:dataset}.

\begin{table}[htbp]
    \centering
    \label{tab:dataset}
    \caption{Six datasets used in this paper. “$|\bm{\mathcal{D}}|$'', “$|\textit{\textbf{X}}|$'', “$|\textit{\textbf{X}}_{disc.}|$'', “$|\textit{\textbf{X}}_{num.}|$'', and “$S$'' denote the number of samples, the number of features, the number of discrete features, the number of numerical features, and the sensitive attribute under consideration, respectively.}
    \begin{tabular}{ccccccc}
        \toprule
            Dataset & $|\bm{\mathcal{D}}|$ & $|\textit{\textbf{X}}|$ & $|\textit{\textbf{X}}_{disc.}|$ & $|\textit{\textbf{X}}_{num.}|$ & $S$\\
        \midrule
        Adult & \num[group-separator={,}]{48842} & 14 & 8 & 6 & Sex \\
        Bank & \num[group-separator={,}]{45211} & 16 & 9 & 7 & Marital \\
        COMAPS & \num[group-separator={,}]{6172} & 7 & 3 & 4 & Race \\
        Default & \num[group-separator={,}]{30000} & 23 & 9 & 14 & Sex \\
        Dutch & \num[group-separator={,}]{60420} & 11 & 9 & 2 & Sex \\
        LSAT & \num[group-separator={,}]{20798} & 11 & 5 & 6 & Race \\
        \bottomrule
    \end{tabular}
\end{table}

\section{Algorithmic and Computational Properties of \textit{MONBM}}

This section provides a comprehensive analysis of the proposed \textit{MONBM} framework, focusing on its computational budget, algorithmic complexity, generalization behavior, robustness, and practical execution time. These analyses aim to clarify the computational characteristics and practical scalability of the proposed framework.

\subsection{Computational Budget Analysis}

For the baseline methods compared with \textit{MONBM-CD}, all single-point methods (e.g., \textit{Fair-NBM}~\cite{kraus2024interpretable}, \textit{LFR-NBM}~\cite{zemel2013learning}, \textit{Re-NBM}~\cite{kamiran2012data}, \textit{ROC-NBM}~\cite{kamiran2012decision}, \textit{LPP-NBM}~\cite{xian2023fair}, \textit{Oracle-NBM}~\cite{xian2024unified}, and \textit{FRAPP{\'E}-NBM}) were trained for 1,200 epochs. \textit{FairEMOL}~\cite{zhang2022mitigating}, a population-based multi-objective method, was trained for 200 generations on a deep neural network with approximately 68,000 parameters.

For the baseline methods compared with \textit{MONBM-CL}, \textit{SNAM}~\cite{xu2023sparse} and \textit{GAMI-Net}~\cite{yang2021gami} were trained for 1,200 and 5,000 epochs, respectively.

In contrast, our method applies \textit{MONBM} to retrain only the ``feature specific'' layer of a pre-trained NBM model, which contains approximately 1,500 parameters. We conduct 200 generations of evolutionary optimization using this efficient retraining strategy on the pre-trained NBM for 1,000 epochs.

Overall, although each generation in \textit{MONBM} involves updating a population of 100 models, the per-model computational cost is relatively low, as only $\sim$2\% of the NBM’s total parameters are updated. Additionally, unlike single-point methods that yield only a single solution per run, \textit{MONBM} produces a diverse set of trade-off solutions in a single run, supporting downstream analysis and flexible deployment.

\subsection{Algorithmic Complexity}

For a multi-objective problem with $k$ objectives, a population size of $\tau$, a data dimension of $D$, and a data size of $n$, the computational complexity analysis of the \textit{MONBM} instantiation (Algorithm 2) is as follows:

\begin{itemize}
    \item \textbf{Time Complexity:} Each generation involves reproduction ($O(\tau \cdot D)$), partial training for $N$ individuals ($O(k \cdot N \cdot n \cdot D)$), and indicator-based fitness evaluation. Specifically, accuracy and fairness evaluations take $O(n \cdot D)$, \textit{GI} takes $O(D)$, while \textit{LI} requires $O(n \cdot D \log D)$ due to feature ranking. The SRA-based environmental selection takes $O(k \cdot \tau^2)$~\cite{li2016stochastic}. Combining these components, the total time complexity per generation is $O(k \tau^2 + \tau \cdot n \cdot D \log D)$, indicating quadratic dependence on population size and near-linear scaling with dataset size and dimensionality.

    \item \textbf{Space Complexity:} The framework maintains the population parameters $O(\tau \cdot D)$ and the training dataset $O(n \cdot D)$. Since the shared bases are frozen and shared, the overall space complexity is $O((\tau + n)D)$, which scales linearly with the feature dimension $D$.
\end{itemize}

\subsection{Robustness and Generalization Analysis}

The generalization behavior of \textit{MONBM} can be understood from both its structural design and its explicit functional regularization mechanisms. Fundamentally, as a class of GAMs, \textit{MONBM} tends to exhibit stable generalization by limiting high-order feature interactions. As established in the bias–variance analysis of additive models~\cite{lou2012intelligible}, reducing complex feature couplings can lower model variance, which helps mitigate overfitting commonly observed in highly interactive models and supports reliable performance across diverse practical scenarios.

From the perspective of statistical learning theory~\cite{bartlett2002rademacher}, generalization performance is closely related to the capacity of the hypothesis space. By restricting evolutionary optimization to approximately 2\% of the total model parameters—specifically the linear combination weights over frozen nonlinear basis functions—the effective hypothesis space of \textit{MONBM} is substantially smaller than that of fully trainable deep models. This controlled parameterization helps regulate model capacity and reduces the risk of overfitting while preserving sufficient expressive power.

Furthermore, the explicit smoothness objective ($\mathcal{M}_s$) acts as a functional regularization mechanism. By discouraging excessive oscillations in the learned shape functions, it helps prevent fitting high-frequency noise and encourages stable global trends. Empirically, this behavior is reflected in the consistent performance observed across 15 independent trials on both training and test sets (Tables 2 to 5 and Tables ~\ref{tab:MONBM-CD_point_train} to~\ref{tab:MONBM_CDLG_point}). The relatively small performance differences between training and testing results suggest stable generalization performance when optimizing multiple objectives simultaneously.

\subsection{Wall-Clock Time Performance}

To provide a precise assessment of the practical scalability and efficiency of the framework, we measured the average wall-clock time for the \textit{MONBM-CD} instantiation across all six benchmark datasets. All experiments were conducted on a Linux server equipped with dual Intel Xeon Platinum 8358 processors (totaling 64 physical cores) and 512GB of RAM. To ensure a fair and consistent comparison, all computational tasks were restricted to a single-threaded execution mode. Specifically, we measured the total runtime required for a full \textit{MONBM} execution (evolving a population of $\tau=100$ models for 200 generations) and compared it with the time required to train a single standard \textit{NBM} model until convergence.

\begin{table}[htbp]
    \caption{Comparison of average wall-clock time (seconds) between a single \textit{NBM} training run and the \textit{MONBM-CD} instantiation across six datasets.}
    \label{tab:wall_clock_time}
    \centering
    \begin{tabular}{ccc}
        \toprule
        Dataset & NBM & MONBM \\
        \midrule
        Adult & 1,681.89 & 19,650.14 \\
        Bank & 1,531.17 & 19,273.12 \\
        COMPAS & 203.473 & 5,359.28\\
        Default & 1,515.16 & 20,792.25 \\
        Dutch & 2,087.10 & 22,009.63 \\
        LSAT & 863.613 & 10,367.60\\
        \midrule
        \midrule
        Mean & 1,313.73 & 16,242.00 \\
        \bottomrule
    \end{tabular}
\end{table}

As summarized in Table~\ref{tab:wall_clock_time}, the total execution time of \textit{MONBM} is on average approximately 12.4 times that of a single NBM training run. However, unlike single-model training, \textit{MONBM} produces a diverse set of trade-off solutions in one execution. Obtaining a comparable approximation of the Pareto front using scalarized optimization would typically require many independent runs with different weight settings, leading to substantially higher cumulative cost. These results indicate that the partial-retraining strategy effectively reduces the computational burden of population-based optimization, making \textit{MONBM} practically feasible for exploring complex trade-offs among accuracy, fairness, and interpretability.

\section{Hyperparameters of Pre-trained NBM and Baseline Methods}
\label{sec:par_NBM}

\subsection{Hyperparameters of Pre-trained NBM}

For training the pre-trained NBM, we used the Adam with decoupled weight decay (AdamW) optimizer~\cite{loshchilov2018decoupled} and trained for 1000 epochs with a batch size of 1024. The learning rate was decayed from its initial value to zero using cosine annealing~\cite{loshchilov2017sgdr}. The shared MLP architecture included three hidden layers with 256, 128, and 128 nodes, and the basis output parameter $B = 100$. We adjusted the starting learning rate in the continuous interval [1e-5, 1.0), weight decay within [1e-10, 1.0), the output penalty coefficients in the range [1e-7, 100), and the dropout and feature dropout coefficients in the discrete set $\{0, 0.05, 0.1, 0.2, 0.3, 0.4, 0.5, 0.6, 0.7, 0.8, 0.9\}$. The optimal hyper-parameters were determined using the validation set and random search. Table~\ref{tab:par_NBM} lists the optimal hyper-parameters for NBM across all datasets.

\begin{table}[htbp]
    \caption{The parameter settings of the NBM on six datasets in this paper.}
    \label{tab:par_NBM}
    \centering
    \resizebox{\textwidth}{!}{
    \arrayrulecolor{black}
    \begin{tabular}{cccccccc}
        \toprule
        Dataset & Number of epochs & Batch size & Learning rate & Weight decay & Dropout & Basis dropout & Output penalty \\
        \midrule
        Adult & 1000 & 1024 & 0.02525 & 0.02234 & 0.0 & 0.05 & 0.00219 \\
        Bank & 1000 & 1024 & 0.02078 & 0.09175 & 0.0 & 0.1 & 0.00803 \\
        COMPAS & 1000 & 1024 & 0.01612 & 0.07796 & 0.2 & 0.05 & 0.00045 \\
        Default & 1000 & 1024 & 0.00995 & 0.07137 & 0.2 & 0.1 & 0.02432 \\
        Dutch & 1000 & 1024 & 0.00037 & 0.02814 & 0.2 & 0.0 & 0.01620 \\
        LSAT & 1000 & 1024 & 0.01452 & 0.04858 & 0.1 & 0.05 & 0.02428 \\
        \bottomrule
    \end{tabular}
    }
\end{table}

\subsection{Hyperparameters of Baseline Methods}

For the pre-process techniques \textit{LFR}~\cite{zemel2013learning} and \textit{reweighting}~\cite{kamiran2012data} and the post-process techniques \textit{ROC}~\cite{kamiran2012decision}, \textit{LPP}~\cite{xian2023fair}, \textit{Oracle}~\cite{xian2024unified}, and \textit{FRAPP{\'E}}~\cite{ctifrea2024frappe} for improving fairness, we used the implementation of the AI Fairness 360 toolkit~\cite{bellamy2018ai} or the source code provided by the authors with recommended parameter settings. \textit{FairEMOL}~\cite{zhang2022mitigating} was also configured with its recommended parameters, with the MLP architecture modified to include 256, 128, and 128 nodes in three hidden layers, instead of the original single hidden layer with 64 nodes for fair comparison.

For the \textit{SNAM}~\cite{xu2023sparse} and \textit{GAMI-Net}~\cite{yang2021gami} methods, the parameters $\lambda$ and $\eta$ control the strength of feature selection, respectively, thus indirectly controlling the local interpretability. Referring to the settings in~\cite{xu2023sparse} and~\cite{yang2021gami}, we set $\lambda$ and $\eta$ to ten different parameter values, i.e., $\lambda=$\{0.01, 0.03, 0.05, 0.08, 0.1, 0.3, 0.5, 0.7, 0.9, 1\} and $\eta=$\{5\textnormal{e}-4, 1\textnormal{e}-3, 3\textnormal{e}-3, 5\textnormal{e}-3, 8\textnormal{e}-3, 0.01, 0.02, 0.03, 0.04, 0.05\}, to obtain models with varying degrees of local interpretability.

\section{Sensitivity Analysis of the Smoothness Scaling Factor $\eta$}

To systematically evaluate the impact of the hyperparameter $\eta$ (and the resulting tolerance threshold $k$) on model interpretability, we conducted a targeted sensitivity analysis. In this section, to isolate the effects of smoothness control and prevent interference from other interpretability dimensions, we specifically simplified the optimization problem to a two-objective configuration: accuracy (CE) and smoothness ($\mathcal{M}_s$), while excluding the monotonicity metric ($\mathcal{M}_m$). We focused this analysis on the \textit{Adult} dataset, exploring a range of $\eta$ values including $\frac{1}{2}, \frac{2}{3}, 1.0, \frac{4}{3}$, and $\frac{3}{2}$. The resulting non-dominated solution sets obtained under these different configurations are visualized in Fig.~\ref{fig:dif_k_distribution_adult} below.

\begin{figure*}[h]
    \centering
    \includegraphics[width=0.7\textwidth]{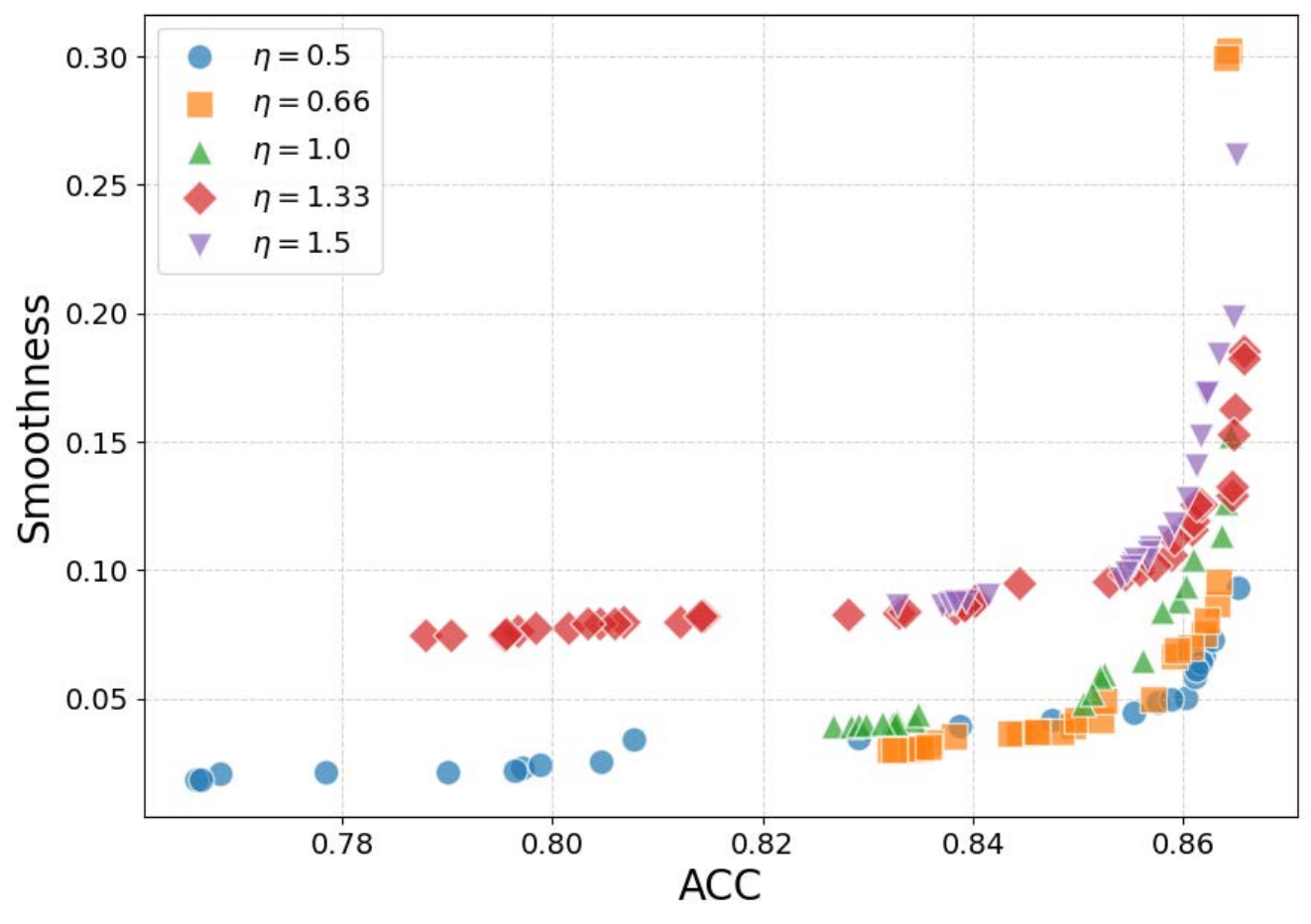}
    \caption{Pareto front distributions of accuracy (ACC) and smoothness ($\mathcal{M}_s$) across different scaling factors $\eta$ on the \textit{Adult} dataset in one arbitrary trial.}
    \label{fig:dif_k_distribution_adult}
\end{figure*}

As intuitively expected, the visualization demonstrates that as $\eta$ decreases, the smoothness constraint becomes progressively more stringent (corresponding to a smaller tolerance threshold $k$). This is evidenced by the shift of the Pareto front toward lower $\mathcal{M}_s$ values, yielding significantly more regularized and smoother shape functions. Such results underscore the practical flexibility of the \textit{MONBM} framework: it empowers users and stakeholders to explicitly calibrate $\eta$ according to their specific requirements, enabling a controllable transition between high-accuracy predictive modeling and highly interpretable, smooth global explanations.

\section{Required Monotonic Features}

\label{sec:monotonicity_metric_setup}
On each dataset, the monotonically increasing and decreasing features required in this paper are shown in Table~\ref{tab:monotonicity_features}.

\begin{table}[htbp]
    \caption{The required monotonically increasing and decreasing features on each dataset in this paper.}
    \label{tab:monotonicity_features}
    \centering
    \resizebox{\textwidth}{!}{
    \begin{tabular}{ccc}
        \toprule
        Dataset & Monotonically Increasing Features & Monotonically Dncreasing Features \\
        \midrule
        Adult & ``capital\_gain'', ``education\_num'', ``hours\_per\_week'' & ``capital\_loss'' \\
        Bank & ``balance'', ``duration'', ``previous'' & ------ \\
        COMPAS & ``priors\_count'' & ------ \\
        Default & ------ & ------ \\
        Dutch & ``edu\_level'' & ------ \\
        LSAT & ``decile1b'', ``decile3'', ``lsat'', ``ugpa'', ``zfygpa'', ``zgpa'' & ------ \\
        \bottomrule
    \end{tabular}
    }
\end{table}

For example, consider the ``hours\_per\_week'' feature from the \textit{Adult} dataset, where the task is to predict whether an individual earns more than \$50,000 annually. The ``hours\_per\_week'' feature describes the number of hours worked weekly. We considered that the probability of his/her annual income exceeding \$50,000 should increase monotonically as the number of hours worked increases, thus setting this feature to need to satisfy monotonically increasing. It is worth noting that the setting is based on our experience and is only intended to validate the proposed metrics and methodology. In practice, the settings should be determined according to specific needs.

\section{Evaluation Criteria}
\label{sec:detail_evaluation_criteria}

We used the comparative methods in~\cite{wang2024multi} and~\cite{zhang2022mitigating} to compare the mutual dominance relationships between \textit{MONBM} and population-based and single solution-based methods, respectively.

For population-based methods, we referred to the comparison method of~\cite{wang2024multi} to compare the mutual dominance relationships between all pairs of solutions in two populations. Specifically, assuming that \textit{MONBM} and the baseline method each produce 100 solutions, we performed pairwise comparisons, resulting in $100 \times 100=10,000$ total comparisons. For example, if in 3000 of these comparisons the corresponding solution of the \textit{MONBM} dominates the corresponding solution of the baseline method, the dominance probability of \textit{MONBM} over the baseline is $3000/10,000 = 0.300$, i.e., ``Do.'' is 0.300.

For single solution-based methods, we referred to the three metrics of~\cite{zhang2022mitigating} to compare the mutual dominance relationships of the two methods. Given a set of models $s$ and a set of model sets $P$ generated by the baseline method and \textit{MONBM} in all $n$ trials, respectively, the \textit{Dominate} metric evaluates whether \textit{MONBM} can find solutions to dominate the baseline method as follows:

\begin{equation}
    \textnormal{Dominate}(P,s)=\frac{1}{n}\sum_{i=1}^n  \textnormal{sign}\left[\sum_{j=1}^{|P^i|}\textnormal{sign}[P_j^i\prec s^i]>0\right],
    \label{eq:zhang_do}
\end{equation}
where $P^i$ and $s^i$ are the model set and model in the $i$-th trial, $P_j^i$ is the $j$-th model in $P^i$,    $P_j^i\prec s^i$ means $P_j^i$ dominates $s^i$, and $\textnormal{sign}[\cdot]$ is equal to 1 if $[\cdot]$ is true, otherwise 0. Larger values of the \textit{Dominate} metric indicate that in more trials the solutions obtained by the baseline method are dominated by some of the solutions in the model set obtained by \textit{MONBM}.

The \textit{Incomparable} metric calculates the average proportion of solutions $P$ obtained by \textit{MONBM} that are incomparable to solutions $s$ obtained by the baseline method across all $n$ trials, as follows:

\begin{equation}
    \textnormal{Incomparable}(P,s)=\frac{1}{n}\sum_{i=1}^n\frac{1}{|P^i|}  \sum_{j=1}^{|P^i|}\textnormal{sign}[(P_j^i\not\prec s^i)\&(s^i\not\prec P_j^i)]>0,
    \label{eq:zhang_non_do}
\end{equation}
where $P_j^i\not\prec s^i$ means $P_j^i$ does not dominate $s^i$. A larger \textit{Incomparable} metric means that more models obtained by \textit{MONBM} are incomparable with $s$. 

The \textit{Dominated} metric calculates the average proportion of solutions obtained by \textit{MONBM} are dominated by solutions $s$ obtained by the baseline method over all $n$ trials, as follows:

\begin{equation}
    \textnormal{Dominated}(P,s)=\frac{1}{n}\sum_{i=1}^n\frac{1}{|P^i|}  \sum_{j=1}^{|P^i|}\textnormal{sign}[s^i\prec P_j^i].
    \label{eq:zhang_be_do}
\end{equation}

A larger \textit{Dominated} values means that more models obtained by \textit{MONBM} are dominated by $s$. In fact, the \textit{Incomparable} and \textit{Dominated} metrics are consistent with the population-based approach to comparing the proportion of mutually non-dominated and dominated~\cite{wang2024multi}.

\section{Results on the Training Set}

This section supplements the performance of the solutions selected in Section 4.2.2 of the main text on the training set to discuss the overfitting phenomenon.

\begin{table*}[htbp]
    \caption{Comparison of the mean values of the \textit{ACC} and \textit{DP} on the training set of the models corresponding to the three points chosen from the final solution set obtained by the \textit{MONBM-CD} instantiation over 15 trials. \textit{ACC} values are expressed as percentages ($\%$). The best average is highlighted with a gray background. The difference between it and the best value is shown in parentheses.}
    \label{tab:MONBM-CD_point_train}
    \resizebox{\textwidth}{!}{
    \centering
    \begin{tabular}{l|cc|cc|cc}
        \toprule
        & \multicolumn{2}{c|}{Adult} & \multicolumn{2}{c|}{Bank} & \multicolumn{2}{c}{COMPAS} \\
        & ACC ($\%$) $\uparrow$ & DP $\downarrow$ & ACC ($\%$) $\uparrow$ & DP $\downarrow$ & ACC ($\%$) $\uparrow$ & DP $\downarrow$ \\
        \midrule
        Best CE & \cellcolor{gray!40}86.57 & 0.1727 (+0.1721) & \cellcolor{gray!40}90.47 & 0.0360 (+0.0352) & \cellcolor{gray!40}69.03 & 0.1964 (+0.1955) \\
        ``Middle DP'' & 85.83 (-0.74) & 0.0865 (+0.0859) & 90.44 (-0.03) & 0.0171 (+0.0162) & 68.51 (-0.52) & 0.0998 (+0.0990) \\
        Best DP & 76.60 (-9.97) & \cellcolor{gray!40}0.0005 & 88.73 (-1.49) & \cellcolor{gray!40}0.0008 & 53.94 (-15.10) & \cellcolor{gray!40}0.0008 \\
        \bottomrule
        \toprule
        & \multicolumn{2}{c|}{Default} & \multicolumn{2}{c|}{Dutch} & \multicolumn{2}{c}{LSAT} \\
        & ACC ($\%$) $\uparrow$ & DP $\downarrow$ & ACC ($\%$) $\uparrow$ & DP $\downarrow$ & ACC ($\%$) $\uparrow$ & DP $\downarrow$ \\
        \midrule
        Best CE & \cellcolor{gray!40}82.01 & 0.0240 (+0.0240) & \cellcolor{gray!40}83.42 & 0.3336 (+0.3311) & \cellcolor{gray!40}90.23 & 0.1644 (+0.1636) \\
        ``Middle DP'' & 82.00 (-0.02) & 0.0155 (+0.0155) & 82.08 (-1.34) & 0.1678 (+0.1653) & 89.93 (-0.30) & 0.0851 (+0.0843) \\
        Best DP & 77.88 (-4.13) & \cellcolor{gray!40}0.0000 & 56.13 (-27.29) & \cellcolor{gray!40}0.0024 & 88.99 (-1.24) & \cellcolor{gray!40}0.0009 \\
        \bottomrule
    \end{tabular}}
\end{table*}

\begin{table*}[htbp]
    \caption{Comparison of the mean values of the \textit{ACC} and \textit{LI} on the training set of the models corresponding to the three points chosen from the final solution set obtained by the \textit{MONBM-CL} instantiation over 15 trials. \textit{ACC} values are expressed as percentages ($\%$). The best average is highlighted with a gray background. The difference between it and the best value is shown in parentheses.}
    \label{tab:MONBM_CL_point}
    \resizebox{\textwidth}{!}{
    \centering
    \begin{tabular}{l|cc|cc|cc}
        \toprule
            & \multicolumn{2}{c|}{Adult} & \multicolumn{2}{c|}{Bank} & \multicolumn{2}{c}{COMPAS} \\
            & ACC ($\%$) $\uparrow$ & LI $\downarrow$ & ACC ($\%$) $\uparrow$ & LI $\downarrow$ & ACC ($\%$) $\uparrow$ & LI $\downarrow$ \\
            \midrule
            Best CE & \cellcolor{gray!40}86.60 & 1.0300 (+0.9205) & \cellcolor{gray!40}90.28 & 1.2031 (+1.1246) & \cellcolor{gray!40}67.41 & 0.7382 (+0.4508) \\
            ``Middle LI'' & 84.89 (-1.71) & 0.5724 (+0.4629) & 89.85 (-0.43) & 0.6326 (+0.5541) & 62.03 (-5.38) & 0.5129 (+0.2255) \\
            Best LI & 65.70 (-20.90) & \cellcolor{gray!40}0.1095 & 83.26 (-7.02) & \cellcolor{gray!40}0.0785 & 54.03 (-13.38) & \cellcolor{gray!40}0.2874 \\
        \bottomrule
        \toprule
            & \multicolumn{2}{c|}{Default} & \multicolumn{2}{c|}{Dutch} & \multicolumn{2}{c}{LSAT} \\
            & ACC ($\%$) $\uparrow$ & LI $\downarrow$ & ACC ($\%$) $\uparrow$ & LI $\downarrow$ & ACC ($\%$) $\uparrow$ & LI $\downarrow$ \\
            \midrule
            Best CE & \cellcolor{gray!40}82.01 & 1.1217 (+0.9689) & \cellcolor{gray!40}83.07 & 0.7317 (+0.6090) & \cellcolor{gray!40}90.00 & 0.9130 (+0.7760) \\
            ``Middle LI'' & 81.13 (-0.88) & 0.6351 (+0.4823) & 80.34 (-2.73) & 0.4174 (+0.2947) & 89.80 (-0.20) & 0.5450 (+0.4080) \\
            Best LI & 64.59 (-17.42) & \cellcolor{gray!40}0.1528 & 68.16 (-14.91) & \cellcolor{gray!40}0.1227 & 89.00 (-1.00) & \cellcolor{gray!40}0.1370 \\
        \bottomrule
    \end{tabular}}
\end{table*}

\begin{table*}[htbp]
    \caption{Comparison of the mean values of the \textit{ACC} and \textit{GI} on the training set of the models corresponding to the three points chosen from the final solution set obtained by the \textit{MONBM-CG} instantiation over 15 trials. \textit{ACC} values are expressed as percentages ($\%$). The best average is highlighted with a gray background. The difference between it and the best value is shown in parentheses.}
    \label{tab:MONBM_CG_point}
    \centering
    \resizebox{\textwidth}{!}{
    \begin{tabular}{l|cc|cc|cc}
        \toprule
            & \multicolumn{2}{c|}{Adult} & \multicolumn{2}{c|}{Bank} & \multicolumn{2}{c}{COMPAS} \\
            & ACC ($\%$) $\uparrow$ & GI $\downarrow$ & ACC ($\%$) $\uparrow$ & GI $\downarrow$ & ACC ($\%$) $\uparrow$ & GI $\downarrow$ \\
            \midrule
            Best CE & \cellcolor{gray!40}85.94 & 0.1713 (+0.1411) & 89.52 (-0.55) & 0.2287 (+0.1864) & 68.79 (-0.02) & 0.0129 (+0.0019) \\
            ``Middle GI'' & 85.89 (-0.04) & 0.0915 (+0.0612) & \cellcolor{gray!40}90.07 & 0.0715 (+0.0291) & \cellcolor{gray!40}68.82 & 0.0120 (+0.0009) \\
            Best GI & 80.62 (-5.31) & \cellcolor{gray!40}0.0303 & 89.70 (-0.37) & \cellcolor{gray!40}0.0423 & 68.42 (-0.40) & \cellcolor{gray!40}0.0111 \\
        \bottomrule
        \toprule
            & \multicolumn{2}{c|}{Default} & \multicolumn{2}{c|}{Dutch} & \multicolumn{2}{c}{LSAT} \\
            & ACC ($\%$) $\uparrow$ & GI $\downarrow$ & ACC ($\%$) $\uparrow$ & GI $\downarrow$ & ACC ($\%$) $\uparrow$ & GI $\downarrow$ \\
            \midrule
            Best CE & \cellcolor{gray!40}78.44 & 0.6897 (+0.6354) & \cellcolor{gray!40}82.68 & 0.0658 (+0.0608) & \cellcolor{gray!40}90.17 & 0.0314 (+0.0042) \\
            ``Middle GI'' & 78.05 (-0.38) & 0.4003 (+0.3460) & 81.78 (-0.90) & 0.0481 (+0.0431) & 90.17 (-0.00) & 0.0292 (+0.0020) \\
            Best GI & 78.35 (-0.09) & \cellcolor{gray!40}0.0544 & 82.49 (-0.19) & \cellcolor{gray!40}0.0050 & 90.14 (-0.03) & \cellcolor{gray!40}0.0272 \\
        \bottomrule
    \end{tabular}}
\end{table*}

\begin{table*}[htpb]
    \caption{Comparison of the mean values of the \textit{ACC}, \textit{DP}, \textit{LI}, and \textit{GI} on the training set of the models corresponding to the four extreme points chosen from the final solution set obtained by the \textit{MONBM-CDLG} instantiation over 15 trials. \textit{ACC} values are expressed as percentages ($\%$). The best average is highlighted with a gray background. The difference between it and the best value is shown in parentheses. The ``Majority Baseline'' row represents the accuracy of a constant classifier that consistently predicts the majority class for each dataset.}
    \label{tab:MONBM_CDLG_point}
    \centering
    \resizebox{\textwidth}{!}{
    \begin{tabular}{l|cccccccc}
        \toprule
            & \multicolumn{4}{c|}{Adult} & \multicolumn{4}{c}{Bank} \\
            & ACC ($\%$) $\uparrow$ & DP $\downarrow$ & LI $\downarrow$ & GI $\downarrow$ & ACC ($\%$) $\uparrow$ & DP $\downarrow$ & LI $\downarrow$ & GI $\downarrow$ \\
            \midrule
            Majority Baseline & 76.1 & --- & --- & --- & 88.3 & --- & --- & --- \\
            \midrule
            Best CE & \cellcolor{gray!40}85.9 & 0.178 (+0.18) & 1.026 (+0.92) & 0.159 (+0.12) & \cellcolor{gray!40}89.5 & 0.041 (+0.04) & 1.160 (+1.07) & 0.226 (+0.19)\\
            Best DP & 69.1 (-16.8) & \cellcolor{gray!40}0.000 & 0.130 (+0.02) & 0.145 (+0.11) & 88.3 (-1.20) & \cellcolor{gray!40}0.000 & 0.525 (+0.43) & 0.111 (+0.08) \\
            Best LI & 65.7 (-20.2) & 0.000 (+0.00) & \cellcolor{gray!40}0.108 & 0.163 (+0.13) & 83.2 (-6.30) & 0.000 (+0.00) & \cellcolor{gray!40}0.095 & 0.295 (+0.26) \\
            Best GI & 69.7 (-16.2) & 0.009 (+0.01) & 0.464 (+0.36) & \cellcolor{gray!40}0.036 & 88.3 (-1.20) & 0.001 (+0.00) & 0.751 (+0.66) & \cellcolor{gray!40}0.032 \\
        \bottomrule
        \toprule
            & \multicolumn{4}{c|}{COMPAS} & \multicolumn{4}{c}{Default} \\
            & ACC ($\%$) $\uparrow$ & DP $\downarrow$ & LI $\downarrow$ & GI $\downarrow$ & ACC ($\%$) $\uparrow$ & DP $\downarrow$ & LI $\downarrow$ & GI $\downarrow$ \\
            \midrule
            Majority Baseline & 54.5 & --- & --- & --- & 77.9 & --- & --- & --- \\
            \midrule
            Best CE & \cellcolor{gray!40}68.5 & 0.224 (+0.22) & 0.780 (+0.51) & 0.091 (+0.08) & \cellcolor{gray!40}78.5 & 0.031 (+0.03) & 1.211 (+1.07) & 0.686 (+0.63)\\
            Best DP & 51.8 (-16.8) & \cellcolor{gray!40}0.000 & 0.401 (+0.13) & 0.033 (+0.02) & 59.3 (-19.2) & \cellcolor{gray!40}0.000 & 0.524 (+0.38) & 0.240 (+0.18) \\
            Best LI & 52.4 (-16.1) & 0.056 (+0.06) & \cellcolor{gray!40}0.275 & 0.037 (+0.02) & 60.0 (-18.5) & 0.082 (+0.08) & \cellcolor{gray!40}0.143 & 0.584 (+0.53) \\
            Best GI & 60.8 (-7.70) & 0.057 (+0.06) & 0.767 (+0.49) & \cellcolor{gray!40}0.013 & 65.4 (-13.1) & 0.056 (+0.06) & 0.415 (+0.27) & \cellcolor{gray!40}0.057 \\
        \bottomrule
        \toprule
            & \multicolumn{4}{c|}{Dutch} & \multicolumn{4}{c}{LSAT} \\
            & ACC ($\%$) $\uparrow$ & DP $\downarrow$ & LI $\downarrow$ & GI $\downarrow$ & ACC ($\%$) $\uparrow$ & DP $\downarrow$ & LI $\downarrow$ & GI $\downarrow$ \\
            \midrule
            Majority Baseline & 52.4 & --- & --- & --- & 89.0 & --- & --- & --- \\
            \midrule
            Best CE & \cellcolor{gray!40}82.5 & 0.324 (+0.32) & 0.778 (+0.68) & 0.075 (+0.07) & \cellcolor{gray!40}90.0 & 0.187 (+0.19) & 1.109 (+0.96) & 0.174 (+0.14)\\
            Best DP & 50.5 (-32.0) & \cellcolor{gray!40}0.000 & 0.976 (+0.88) & 0.019 (+0.01) & 63.0 (-27.0) & \cellcolor{gray!40}0.000 & 0.171 (+0.02) & 0.171 (+0.14) \\
            Best LI & 66.1 (-16.4) & 0.928 (+0.93) & \cellcolor{gray!40}0.097 & 0.163 (+0.16) & 57.8 (-32.2) & 0.000 (+0.00) & \cellcolor{gray!40}0.154 & 0.187 (+0.16) \\
            Best GI & 70.5 (-12.0) & 0.406 (+0.41) & 0.681 (+0.58) & \cellcolor{gray!40}0.006 & 84.1 (-5.90) & 0.026 (+0.03) & 0.569 (+0.42) & \cellcolor{gray!40}0.029 \\
        \bottomrule
    \end{tabular}}
\end{table*}

\newpage

\section{Sensitivity Analysis to Fairness Metrics: Results on Equalized Odds (EOD)}

To address the concern that the observed trade-off trends might be specific to the \textit{DP} metric, we conducted a sensitivity analysis by optimizing the equalized odds (EOD) metric—a label-aware fairness definition that constrains the model to maintain equal true positive and false positive rates across subgroups. We implemented the \textit{MONBM-CE} instantiation, considering both accuracy and \textit{EOD} objectives.

Table~\ref{tab:MONBM-EOD_point} summarizes the performance of the models at the extreme and ``middle \textit{EOD}'' points. The empirical results reveal a trade-off pattern remarkably consistent with that observed for \textit{DP} (see Table 2 in the main text). Specifically, moderate \textit{EOD} improvement (defined as a 50\% reduction in the \textit{EOD} metric relative to its extremes, represented by the “middle \textit{EOD}” point) has a marginal impact on accuracy, with an average \textit{ACC} decrease of only 0.64\% across six datasets. However, pursuing extreme fairness (\textit{EOD} $\approx$ 0) leads to substantial accuracy declines on the \textit{Adult}, \textit{COMPAS}, and \textit{Dutch} datasets, while its impact remains moderate on the \textit{Bank}, \textit{Default}, and \textit{LSAT} datasets. This confirms that the inherent difficulty of balancing accuracy and fairness is a robust phenomenon across different fairness definitions.

\begin{table*}[h]
    \caption{Comparison of the mean values of the \textit{ACC} and \textit{EOD} on the test set of the models corresponding to the three points chosen from the final solution set obtained by the \textit{MONBM-CE} instantiation over 15 trials. \textit{ACC} values are expressed as percentages ($\%$). The best average is highlighted with a gray background. The difference between it and the best value is shown in parentheses.}
    \label{tab:MONBM-EOD_point}
    \centering
    \resizebox{\textwidth}{!}{
    \begin{tabular}{l|cc|cc|cc}
        \toprule
        & \multicolumn{2}{c|}{Adult} & \multicolumn{2}{c|}{Bank} & \multicolumn{2}{c}{COMPAS} \\
        & ACC ($\%$) $\uparrow$ & EOD $\downarrow$ & ACC ($\%$) $\uparrow$ & EOD $\downarrow$ & ACC ($\%$) $\uparrow$ & EOD $\downarrow$ \\
        \midrule
        Best CE & \cellcolor{gray!40}86.37 & 0.0485 (+0.0482) & \cellcolor{gray!40}90.32 & 0.0181 (+0.0178) & \cellcolor{gray!40}68.60 & 0.1280 (+0.1280) \\
        ``Middle EOD'' & 85.38 (-0.99) & 0.0236 (+0.0233) & 90.23 (-0.09) & 0.0094 (+0.0091) & 68.19 (-0.41) & 0.0643 (+0.0643) \\
        Best EOD & 76.07 (-10.3) & \cellcolor{gray!40}0.0003 & 88.62 (-1.70) & \cellcolor{gray!40}0.0003 & 52.94 (-15.66) & \cellcolor{gray!40}0.0000 \\
        \bottomrule
        \toprule
        & \multicolumn{2}{c|}{Default} & \multicolumn{2}{c|}{Dutch} & \multicolumn{2}{c}{LSAT} \\
        & ACC ($\%$) $\uparrow$ & EOD $\downarrow$ & ACC ($\%$) $\uparrow$ & EOD $\downarrow$ & ACC ($\%$) $\uparrow$ & EOD $\downarrow$ \\
        \midrule
        Best CE & \cellcolor{gray!40}81.94 & 0.0143 (+0.0141) & \cellcolor{gray!40}81.86 & 0.0784 (+0.0784) & \cellcolor{gray!40}90.00 & 0.1786 (+0.1780) \\
        ``Middle EOD'' & 80.94 (-1.00) & 0.0093 (+0.0091) & 80.75 (-1.11) & 0.0380 (+0.0380) & 89.74 (-0.26) & 0.0894 (+0.0888) \\
        Best EOD & 77.92 (-4.02) & \cellcolor{gray!40}0.0002 & 52.39 (-29.47) & \cellcolor{gray!40}0.0000 & 88.97 (-1.03) & \cellcolor{gray!40}0.0006 \\
        \bottomrule
    \end{tabular}}
\end{table*}

\begin{figure*}[htbp]
    \centering
    \subfigure[Global explanation results of \textit{MONBM-CE} instantiation on the \textit{Adult} dataset]{
        \includegraphics[width=\textwidth]{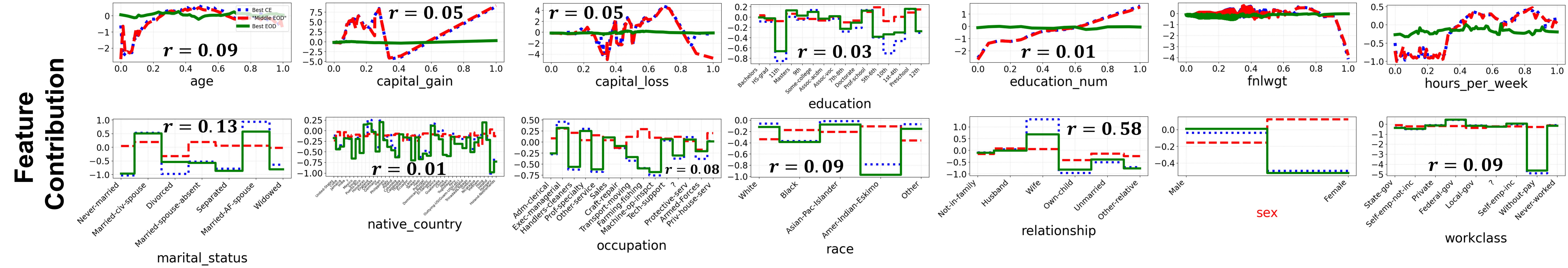}
        \label{fig:adult_EOD_merge_global}
    }
    \subfigure[Global explanation results of \textit{MONBM-CE} instantiation on the \textit{Bank} dataset]{
        \includegraphics[width=\textwidth]{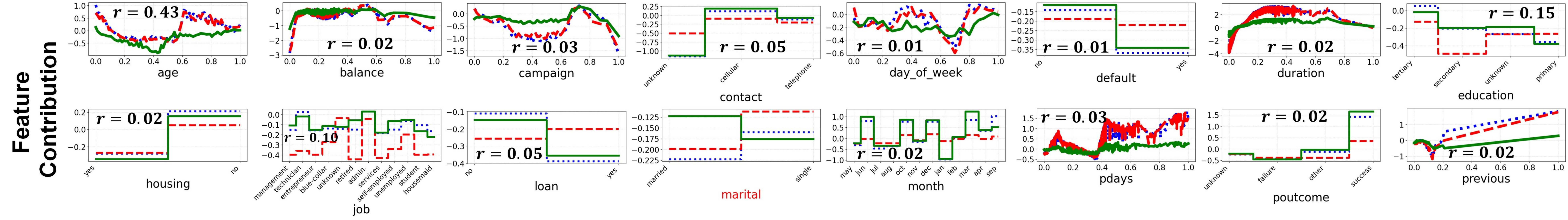}
        \label{fig:bank_EOD_merge_global}
    }
    \subfigure[Global explanation results of \textit{MONBM-CE} instantiation on the \textit{COMPAS} dataset]{
        \includegraphics[width=0.6\textwidth]{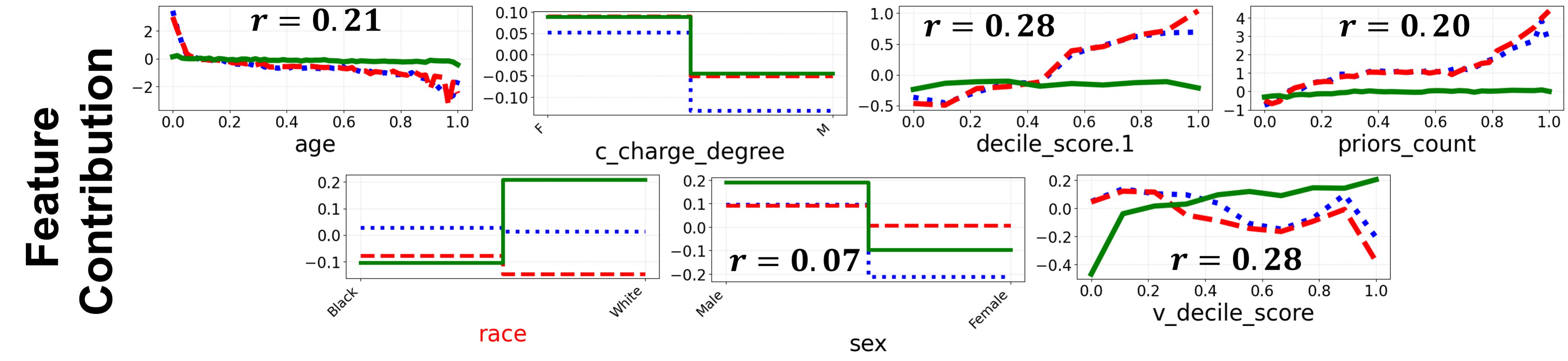}
        \label{fig:compas_EOD_merge_global}
    }
    \subfigure[Global explanation results of \textit{MONBM-CE} instantiation on the \textit{Default} dataset]{
        \includegraphics[width=\textwidth]{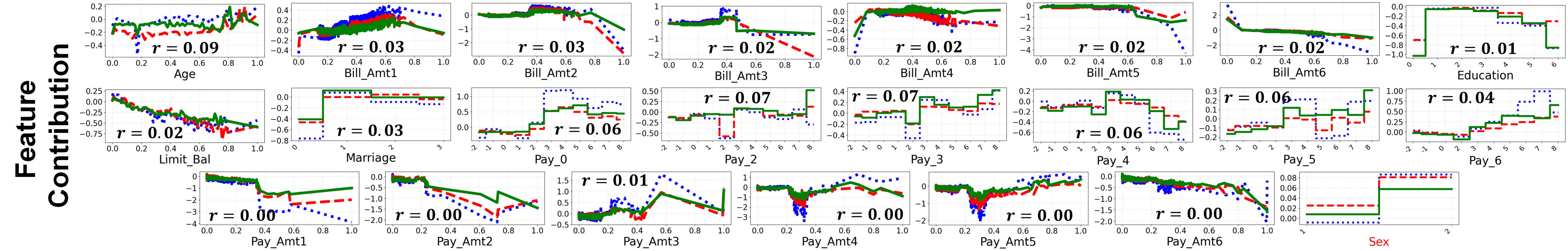}
        \label{fig:default_EOD_merge_global}
    }
    \subfigure[Global explanation results of \textit{MONBM-CE} instantiation on the \textit{Dutch} dataset]{
        \includegraphics[width=\textwidth]{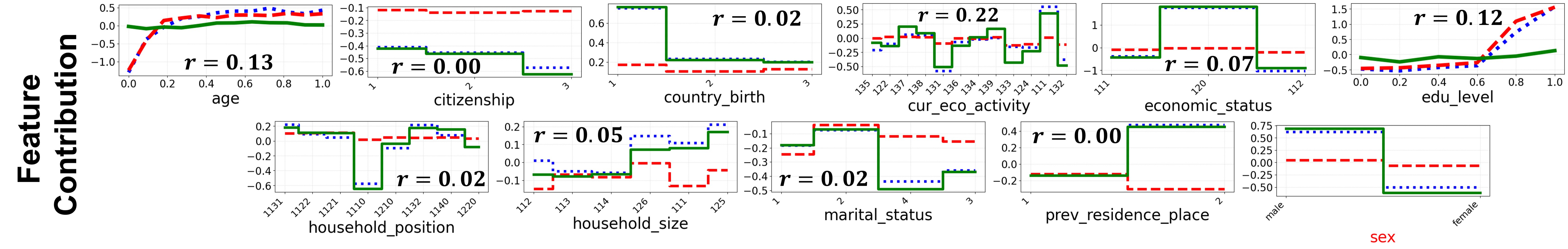}
        \label{fig:dutch_EOD_merge_global}
    }
    \subfigure[Global explanation results of \textit{MONBM-CE} instantiation on the \textit{LSAT} dataset]{
        \includegraphics[width=\textwidth]{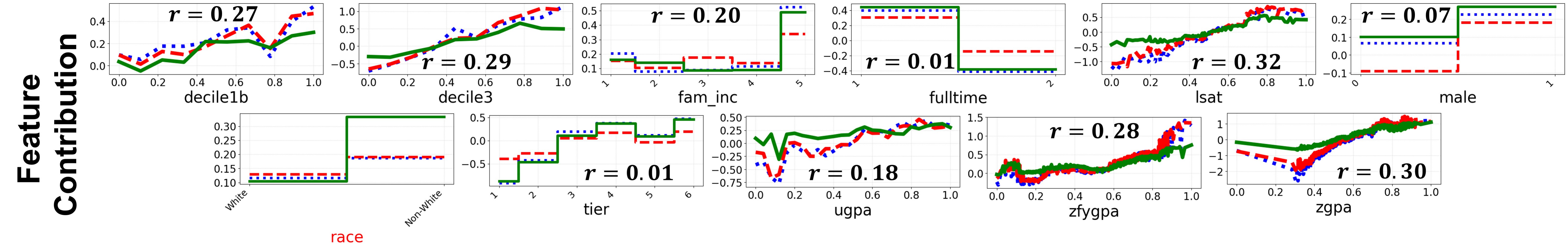}
        \label{fig:lsat_EOD_merge_global}
    }
    \caption{Comparison of global explanations to assess the impact of improved fairness on each feature. For each dataset, the blue, red, and green lines represent the global explanation results of the model corresponding to the best \textit{CE}, ``middle \textit{EOD}'', and best \textit{EOD} points obtained by the \textit{MONBM-CE} instantiation in one arbitrary trial, respectively. Each subfigure corresponds to the global explanation of a feature, with the x-axis representing feature values and the y-axis showing the outputs of the shape functions. The sensitive attribute is highlighted in red. The Pearson correlation coefficient between each attribute and the sensitive attribute is annotated in the respective subplots.}
    \label{fig:EOD_global_explanation_sup}
\end{figure*}

Furthermore, Fig.~\ref{fig:EOD_global_explanation_sup} visualizes the global explanations for the \textit{MONBM-CE} instantiation. A cross-comparison with the \textit{DP}-optimized results (Fig. 5 and Fig. S-2) shows that the shape functions exhibit highly similar trends and feature-importance shifts. Notably, to satisfy \textit{EOD} constraints, the framework still assigns distinct weights to different values of the sensitive attribute (e.g., sex or race). This observation suggests that, given the intrinsic bias present in these datasets, even label-aware metrics like \textit{EOD} necessitate significant adjustments to the sensitive attribute's contribution to compensate for disparate error distributions among groups. These results reinforce the conclusion that the interpretability profiles and trade-off insights provided by \textit{MONBM} are generalizable and not limited to a single fairness metric.

\section{Stability and Reliability of Explanations Across Independent Trials}

A systematic cross-analysis of the 15 independent trials (Figs.~\ref{fig:DP_global_explanation_avg_sup},~\ref{fig:LU_local_explanation_sup_avg},~\ref{fig:GI_global_explanation_avg_sup},~\ref{fig:GU_local_explanation_sup_avg}, and~\ref{fig:4obj_local_explanation_sup_avg}) reveals a clear relationship between predictive performance and explanatory stability. Specifically, the standard deviation of feature importance remains relatively low near the ``Best CE'' solutions but increases at the ``Middle'' and ``Extreme'' operating points along the fairness and local interpretability dimensions. Two main factors contribute to this behavior:

\begin{enumerate}
    \item When strong fairness or interpretability constraints are imposed, predictive accuracy often decreases. Under such conditions, the optimization landscape becomes less sharply defined, allowing multiple feature configurations to satisfy the constraints. Consequently, the learned decision logic becomes more sensitive to variations in the sampled training data.

    \item The ``middle points'' are selected as knee points from the non-dominated set of each independent run. Since the Pareto front shifts across different training samples, the ``middle point'' represents a varying compromise solution rather than a fixed parameter set, naturally contributing to the fluctuations in aggregate results.
\end{enumerate}

Overall, while the main trade-off trends identified by \textit{MONBM} remain consistent across trials, these analyses indicate that the stability of explanations is closely linked to predictive performance and the position of the model on the trade-off surface.

It is important to note that the observed variance in per-feature importance does not imply that the method itself is unreliable. Since GAM-based explanations are 100\% faithful to the underlying model, differences in shape functions across runs primarily reflect variations in the training samples rather than shortcomings of the framework.

\section{Additional Explanation Results}

This section supplements the global and local explanation results for datasets not shown on the various instantiations in Section 4.2 of the main text.

\begin{figure*}[htbp]
    \centering
    \subfigure[Global explanation results for continuous features of \textit{MONBM-CG} instantiation on the \textit{COMPAS} dataset]{
        \includegraphics[width=\textwidth]{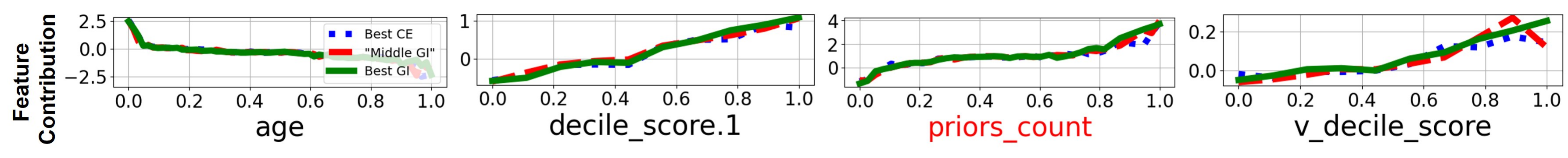}
        \label{fig:compas_GU_merge_global_con}
    }
    \subfigure[Global explanation for continuous features of \textit{MONBM-CG} instantiation on the \textit{Default} dataset]{
        \includegraphics[width=\textwidth]{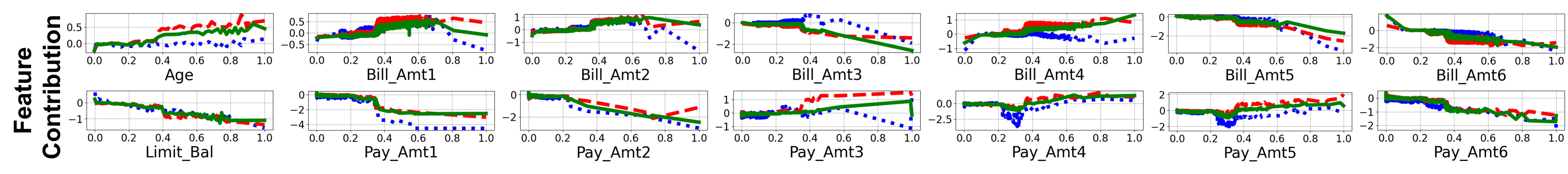}
        \label{fig:default_GU_merge_global_con}
    }
    \subfigure[Global explanation for continuous features of \textit{MONBM-CG} instantiation on the \textit{Dutch} dataset]{
        \includegraphics[width=0.6\textwidth]{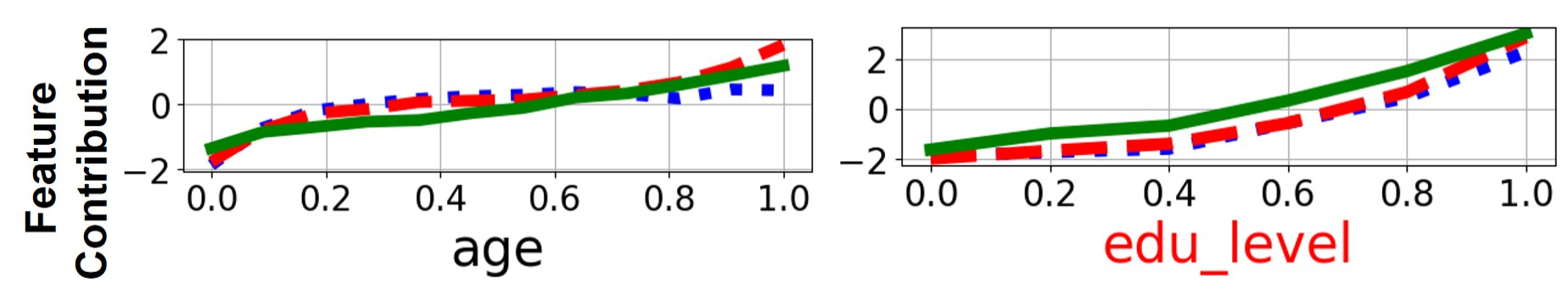}
        \label{fig:dutch_GU_merge_global_con}
    }
    \subfigure[Global explanation for continuous features of \textit{MONBM-CG} instantiation on the \textit{LSAT} dataset]{
        \includegraphics[width=\textwidth]{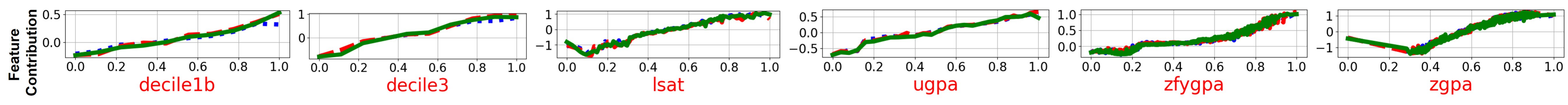}
        \label{fig:LSAT_GU_merge_global_con}
    }
    \caption{Comparison of global explanations of continuous features to validate the proposed smoothness and monotonicity metrics. For each dataset, the blue, red, and green lines represent the global explanation results of the continuous features by the model corresponding to the best \textit{CE}, ``middle \textit{GI},'' and best \textit{GI} points obtained by the \textit{MONBM-CG} instantiation in one arbitrary trial, respectively. Each subfigure corresponds to the global explanation of a feature, with the x-axis representing feature values and the y-axis showing the outputs of the shape functions. Each continuous feature has a smoothness requirement, and features highlighted in red also have a monotonicity requirement.}
    \label{fig:GU_merge_global_con_sup}
\end{figure*}

\begin{figure*}[htbp]
    \centering
    \subfigure[Distribution of shape functions for continuous features across non-dominated knee-point solutions for the \textit{MONBM-CG} instantiation on the \textit{Adult} dataset.]{
        \includegraphics[width=0.8\textwidth]{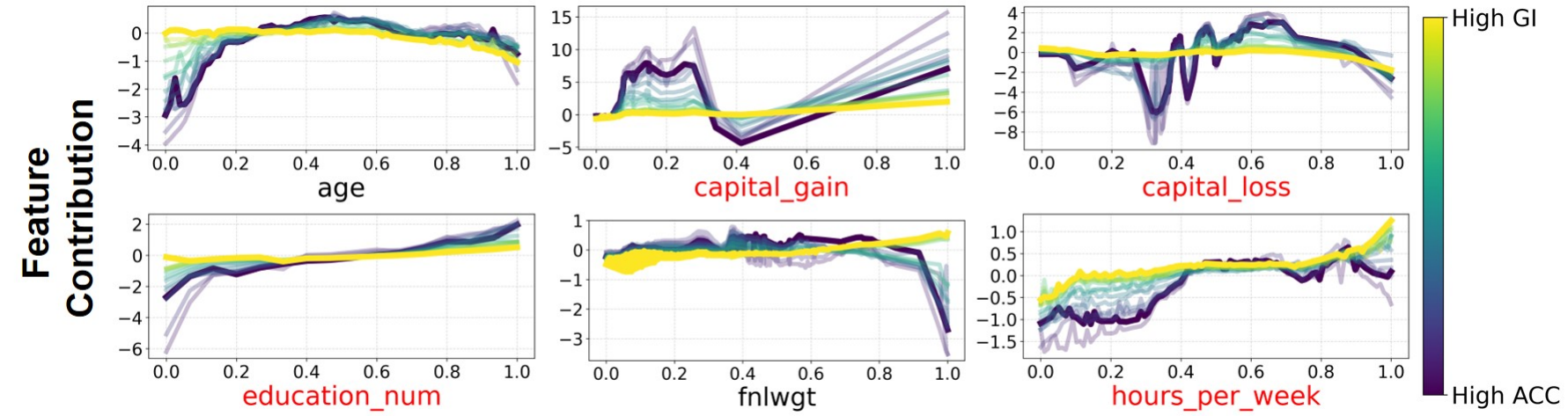}
        \label{fig:adult_GU_merge_global_con_trend}
    }
    \subfigure[Distribution of shape functions for continuous features across non-dominated knee-point solutions for the \textit{MONBM-CG} instantiation on the \textit{Bank} dataset.]{
        \includegraphics[width=\textwidth]{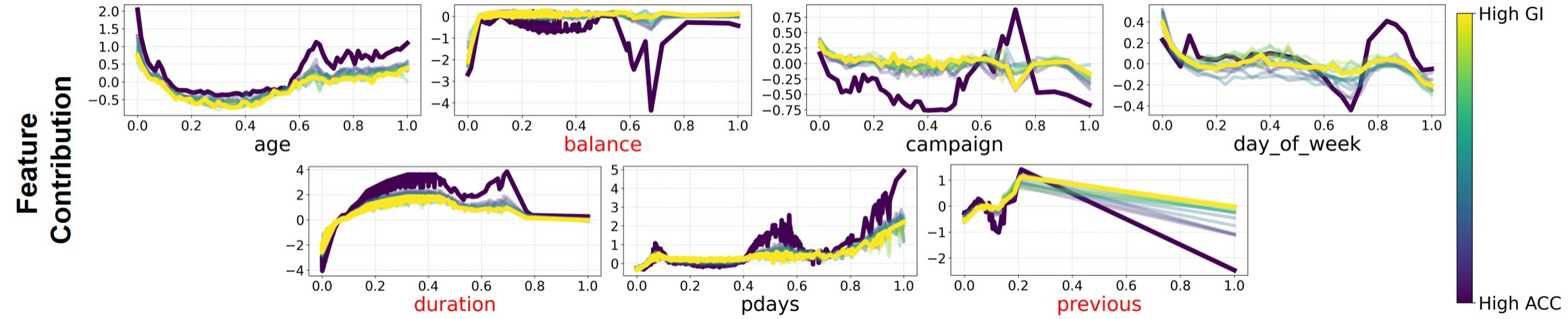}
        \label{fig:bank_GU_merge_global_con_trend}
    }
    \subfigure[Distribution of shape functions for continuous features across non-dominated knee-point solutions for the \textit{MONBM-CG} instantiation on the \textit{COMPAS} dataset.]{
        \includegraphics[width=\textwidth]{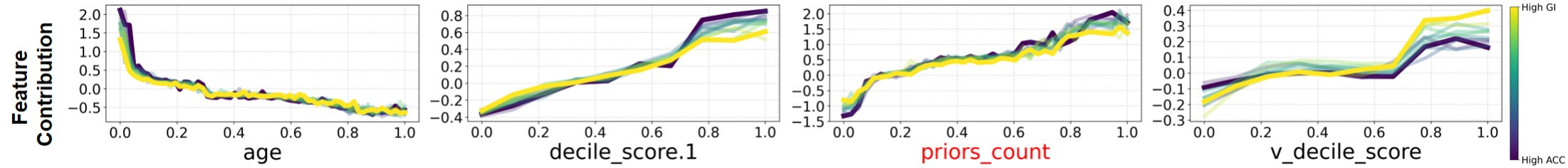}
        \label{fig:compas_GU_merge_global_con_trend}
    }
    \subfigure[Distribution of shape functions for continuous features across non-dominated knee-point solutions for the \textit{MONBM-CG} instantiation on the \textit{Default} dataset.]{
        \includegraphics[width=\textwidth]{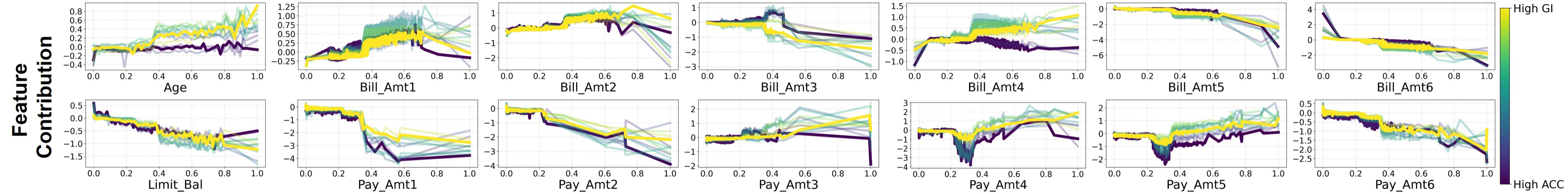}
        \label{fig:default_GU_merge_global_con_trend}
    }
    \subfigure[Distribution of shape functions for continuous features across non-dominated knee-point solutions for the \textit{MONBM-CG} instantiation on the \textit{Dutch} dataset.]{
        \includegraphics[width=0.6\textwidth]{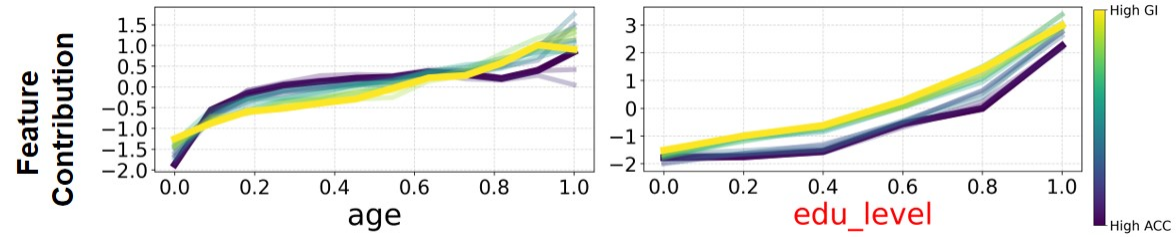}
        \label{fig:dutch_GU_merge_global_con_trend}
    }
    \subfigure[Distribution of shape functions for continuous features across non-dominated knee-point solutions for the \textit{MONBM-CG} instantiation on the \textit{LSAT} dataset.]{
        \includegraphics[width=\textwidth]{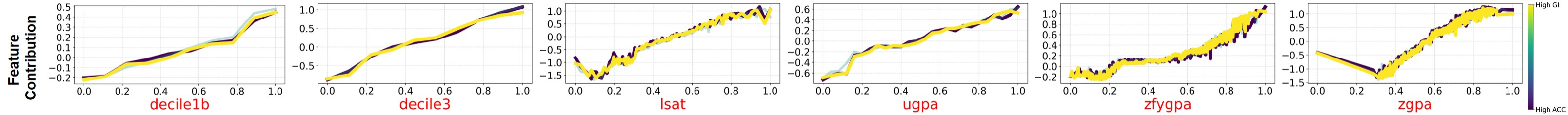}
        \label{fig:lsat_GU_merge_global_con_trend}
    }
    \caption{Variation of global explanations for continuous features among non-dominated knee-point solutions. In each subfigure, the collection of curves represents the shape functions of continuous features for the set of non-dominated knee-point solutions obtained from the final generation of the \textit{MONBM-CG} instantiation in one arbitrary trial. The color gradient, ranging from deep purple (High \textit{ACC}, prioritizing predictive accuracy) to bright yellow (High \textit{GI}, emphasizing smoothness and monotonicity), illustrates the gradual transition in the continuous shape functions across the trade-off surface.}
    \label{fig:GU_merge_global_con_show_trend}
\end{figure*}

\begin{figure*}[htbp]
    \centering
    \subfigure[Global explanation results of \textit{MONBM-CD} instantiation on the \textit{Adult} dataset]{
        \includegraphics[width=\textwidth]{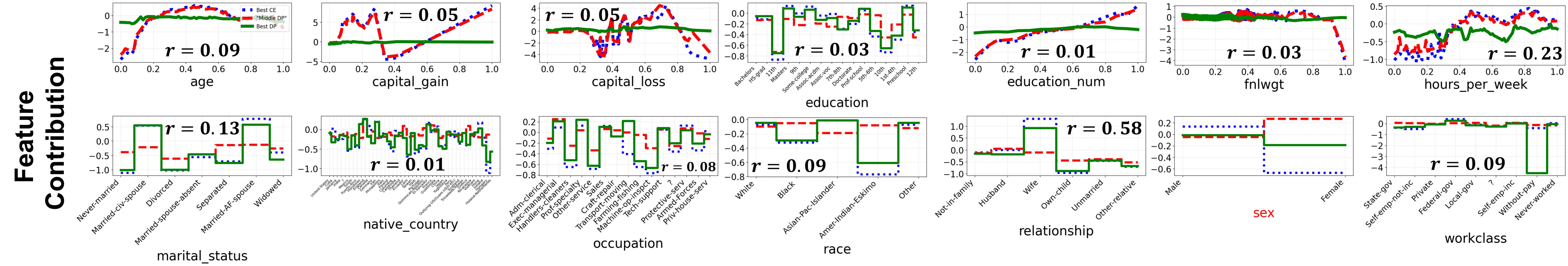}
        \label{fig:adult_DP_global}
    }
    \subfigure[Global explanation results of \textit{MONBM-CD} instantiation on the \textit{Bank} dataset]{
        \includegraphics[width=\textwidth]{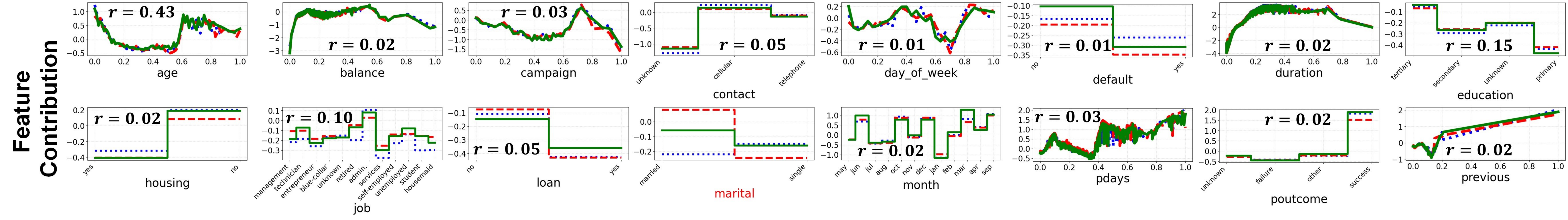}
        \label{fig:bank_DP_global}
    }
    \subfigure[Global explanation results of \textit{MONBM-CD} instantiation on the \textit{Default} dataset]{
        \includegraphics[width=\textwidth]{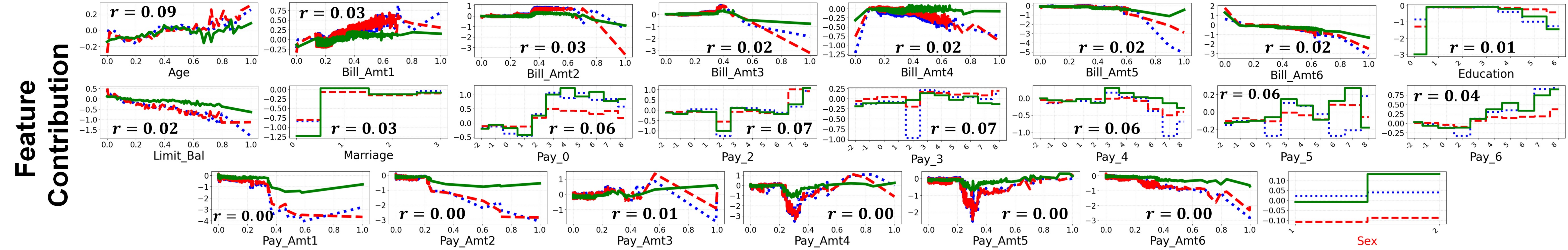}
        \label{fig:default_DP_global}
    }
    \subfigure[Global explanation results of \textit{MONBM-CD} instantiation on the \textit{Dutch} dataset]{
        \includegraphics[width=\textwidth]{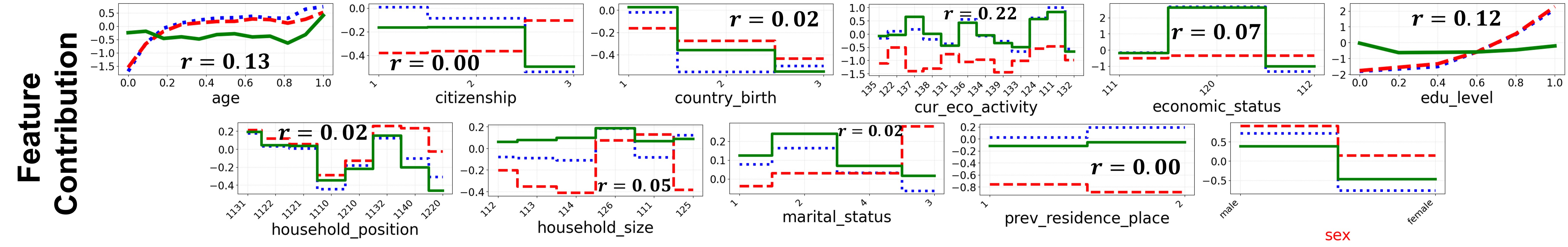}
        \label{fig:dutch_DP_global}
    }
    \caption{Comparison of global explanations to assess the impact of improved fairness on each feature. For each dataset, The blue, red, and green lines represent the global explanation results of the model corresponding to the best \textit{CE}, ``middle \textit{DP}'', and best \textit{DP} points obtained by the \textit{MONBM-CD} instantiation in one arbitrary trial, respectively. Each subfigure corresponds to the global explanation of a feature, with the x-axis representing feature values and the y-axis showing the outputs of the shape functions. The sensitive attribute is highlighted in red. The Pearson correlation coefficient between each attribute and the sensitive attribute is annotated in the respective subplots.}
    \label{fig:DP_global_explanation_sup}
\end{figure*}

\begin{figure*}[htbp]
    \centering
    \subfigure[Mean global explanation results of \textit{MONBM-CD} instantiation over 15 independent trials on the \textit{Adult} dataset]{
        \includegraphics[width=\textwidth]{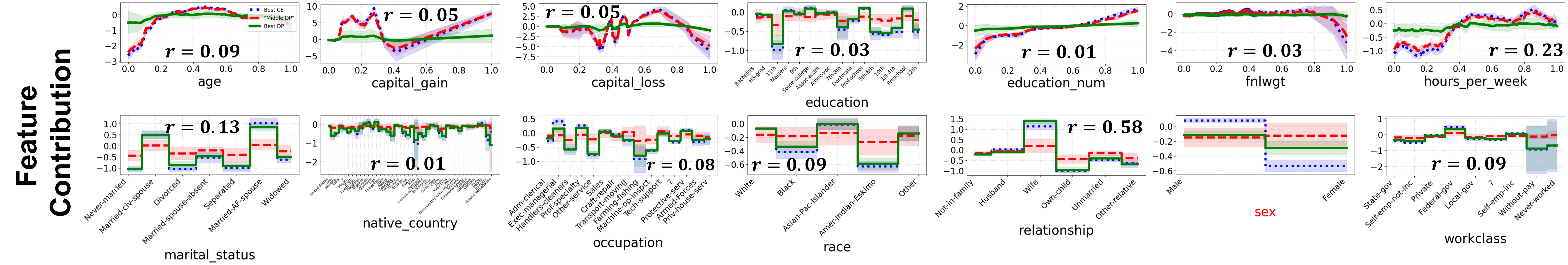}
        \label{fig:adult_DP_merge_global_15avg}
    }
    \subfigure[Mean global explanation results of \textit{MONBM-CD} instantiation over 15 independent trials on the \textit{Bank} dataset]{
        \includegraphics[width=\textwidth]
        {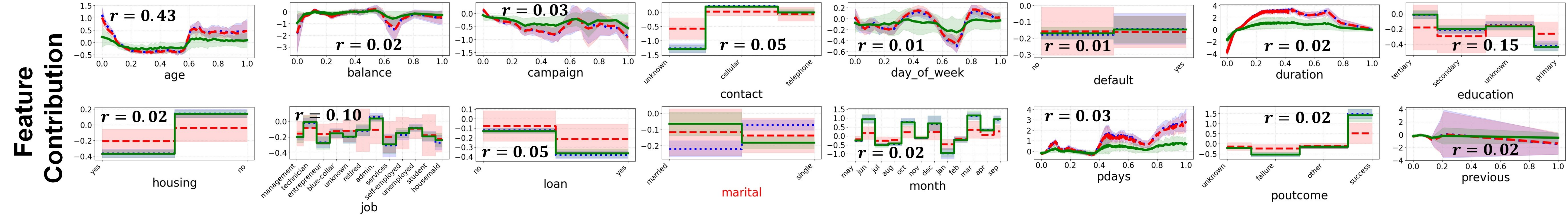}
        \label{fig:bank_DP_merge_global_15avg}
    }
    \subfigure[Mean global explanation results of \textit{MONBM-CD} instantiation over 15 independent trials on the \textit{COMPAS} dataset]{
        \includegraphics[width=0.6\textwidth]{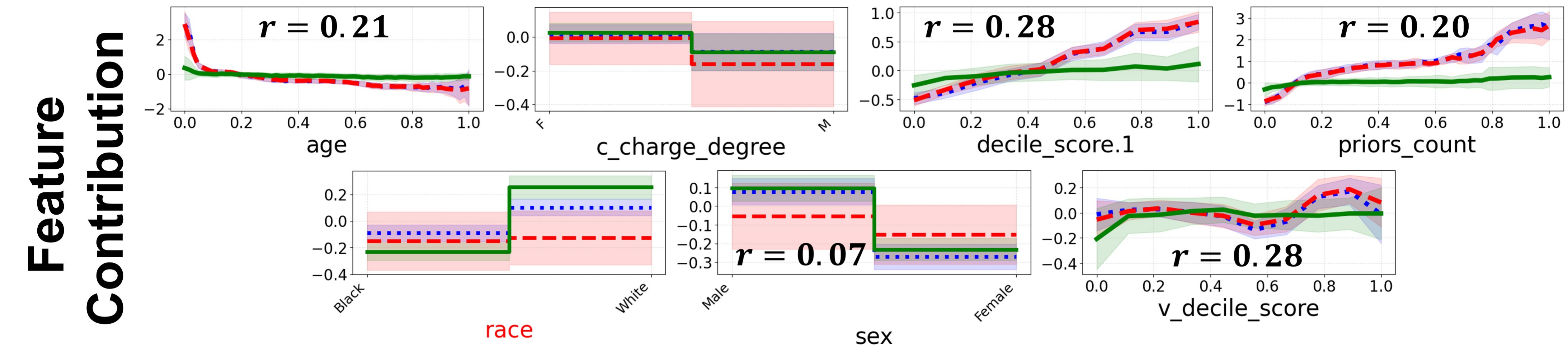}
        \label{fig:compas_DP_merge_global_15avg}
    }
    \subfigure[Mean global explanation results of \textit{MONBM-CD} instantiation over 15 independent trials on the \textit{Default} dataset]{
        \includegraphics[width=\textwidth]{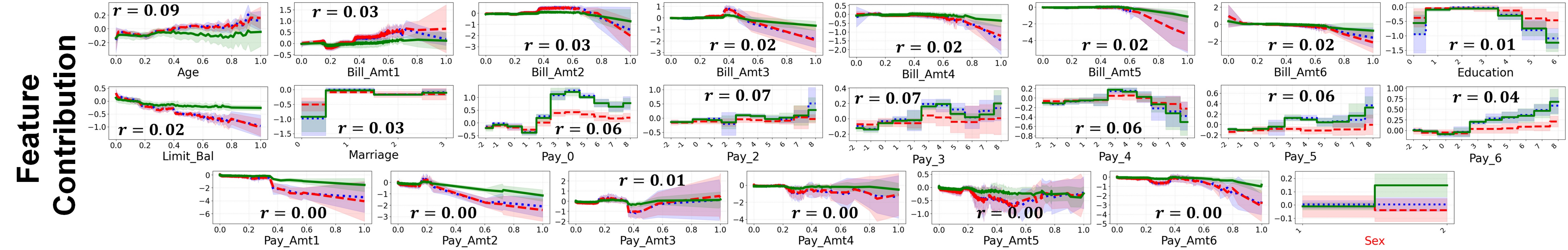}
        \label{fig:default_DP_merge_global_15avg}
    }
    \subfigure[Mean global explanation results of \textit{MONBM-CD} instantiation over 15 independent trials on the \textit{Dutch} dataset]{
        \includegraphics[width=\textwidth]{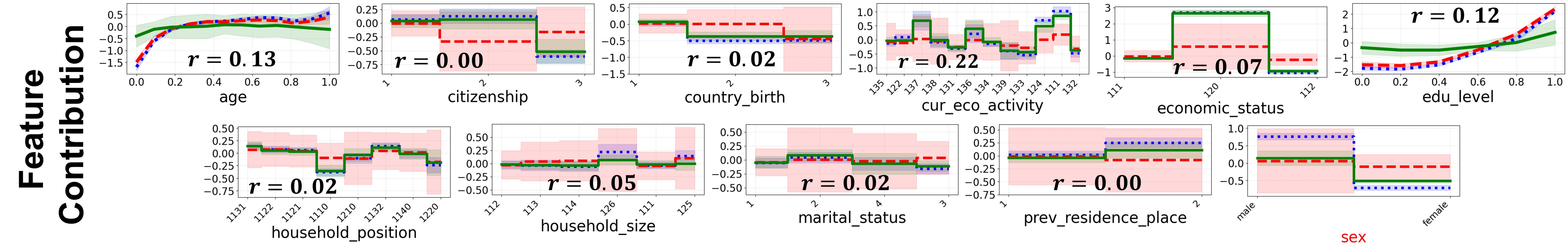}
        \label{fig:dutch_DP_merge_global_15avg}
    }
    \subfigure[Mean global explanation results of \textit{MONBM-CD} instantiation over 15 independent trials on the \textit{LSAT} dataset]{
        \includegraphics[width=\textwidth]{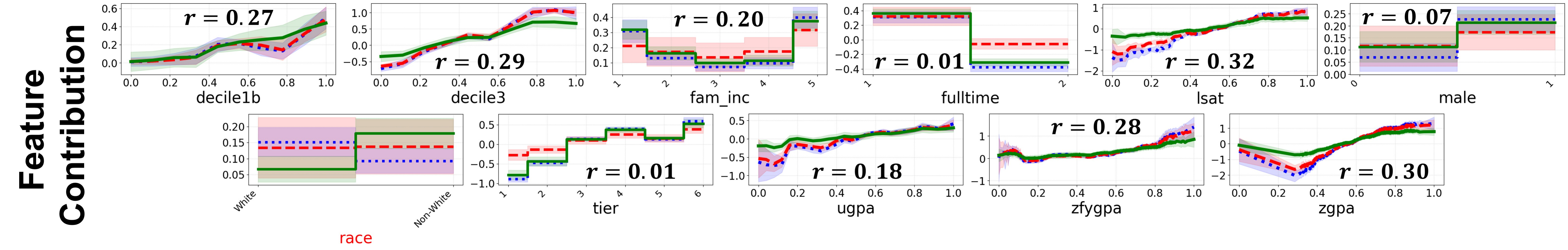}
        \label{fig:lsat_DP_merge_global_15avg}
    }
    \caption{Comparison of global explanations to assess the impact of improved fairness on each feature. For each dataset, the blue, red, and green lines and their corresponding shaded areas represent the mean global explanation results and standard deviations of the models corresponding to the best \textit{CE}, ``middle \textit{DP}'', and best \textit{DP} points obtained by the \textit{MONBM-CD} instantiation averaged over 15 independent trials, respectively. Each subfigure corresponds to the global explanation of a feature, with the x-axis representing feature values and the y-axis showing the outputs of the shape functions. The sensitive attribute is highlighted in red. The Pearson correlation coefficient between each attribute and the sensitive attribute is annotated in the respective subplots.}
    \label{fig:DP_global_explanation_avg_sup}
\end{figure*}

\begin{figure}[htbp]
    \centering
    \subfigure[Average local explanation of \textit{MONBM-CL} instantiation on the \textit{Adult} dataset]{
        \includegraphics[width=0.48\textwidth]{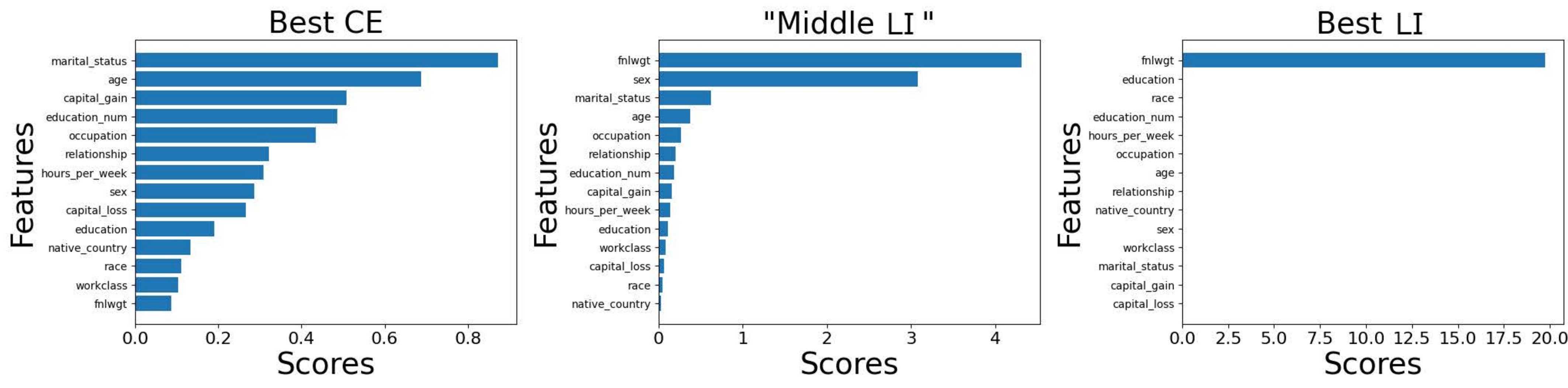}
        \label{fig:adult_LU_merge_local}
    }
    \subfigure[Average local explanation of \textit{MONBM-CD} instantiation on the \textit{Adult} dataset]{
        \includegraphics[width=0.485\textwidth]{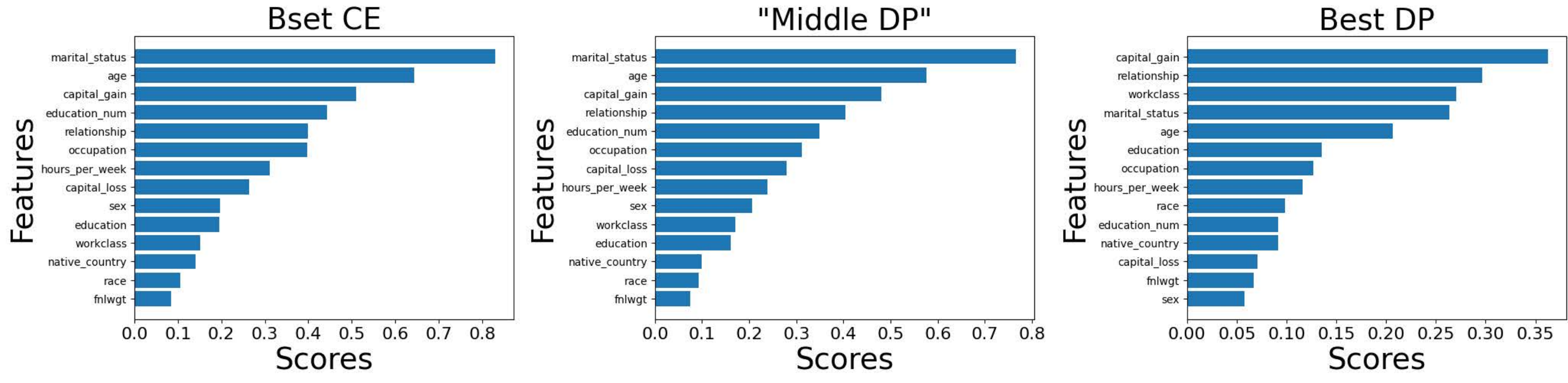}
        \label{fig:adult_DP_merge_local}
    }
    \subfigure[Average local explanation of \textit{MONBM-CL} instantiation on the \textit{COMPAS} dataset]{
        \includegraphics[width=0.48\textwidth]{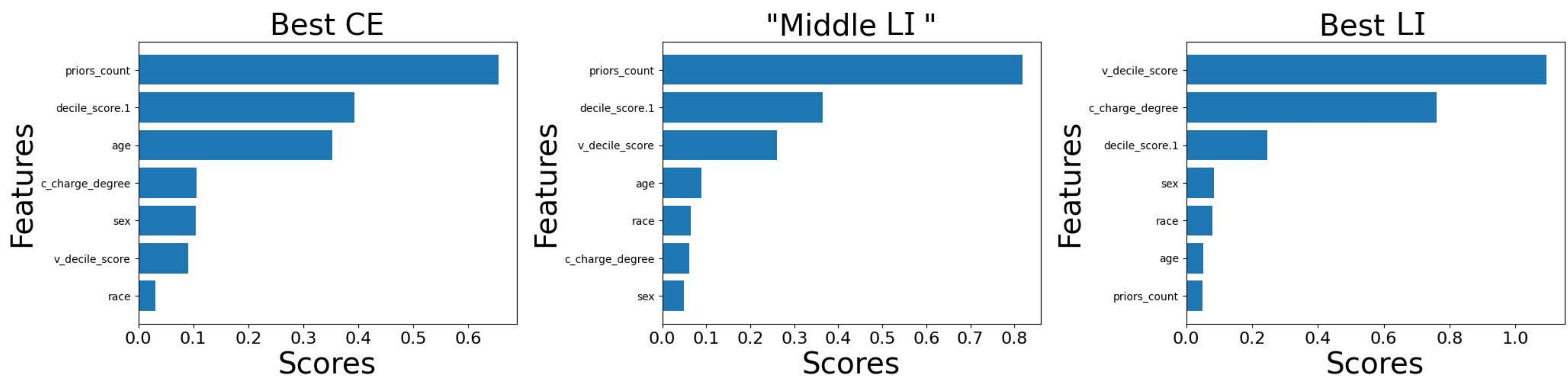}
        \label{fig:compas_LU_merge_local}
    }
    \subfigure[Average local explanation of \textit{MONBM-CD} instantiation on the \textit{COMPAS} dataset]{
        \includegraphics[width=0.48\textwidth]{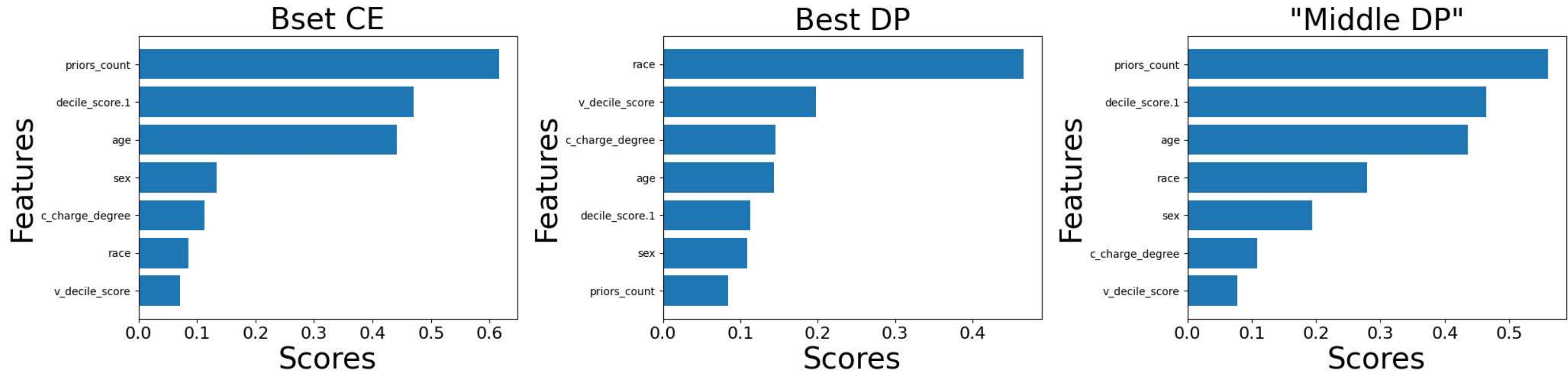}
        \label{fig:compas_DP_merge_local}
    }
    \subfigure[Average local explanation of \textit{MONBM-CL} instantiation on the \textit{Default} dataset]{
        \includegraphics[width=0.48\textwidth]{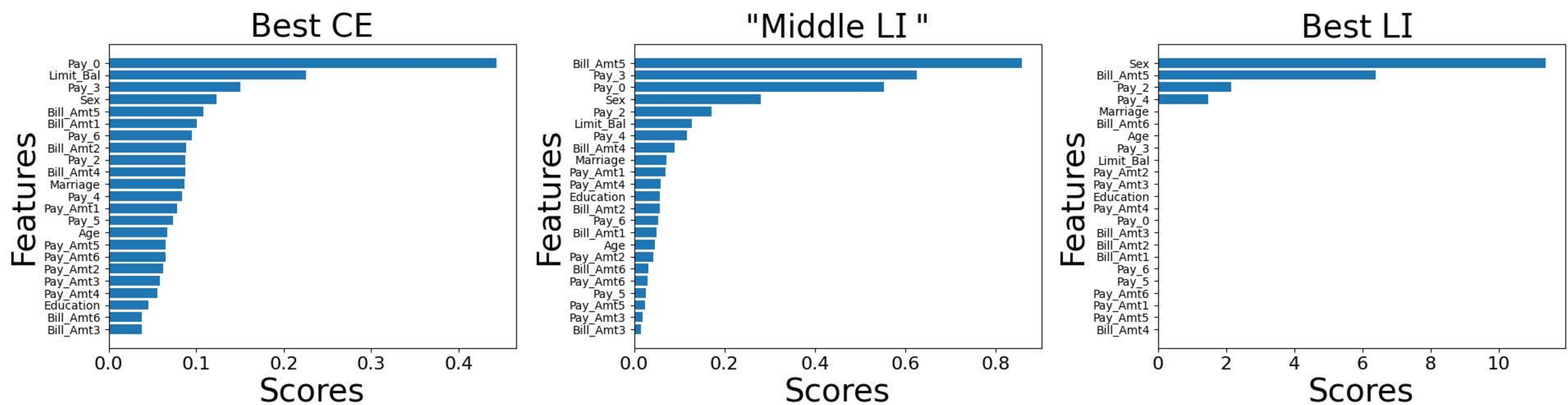}
        \label{fig:default_LU_merge_local}
    }
    \subfigure[Average local explanation of \textit{MONBM-CD} instantiation on the \textit{Default} dataset]{
        \includegraphics[width=0.48\textwidth]{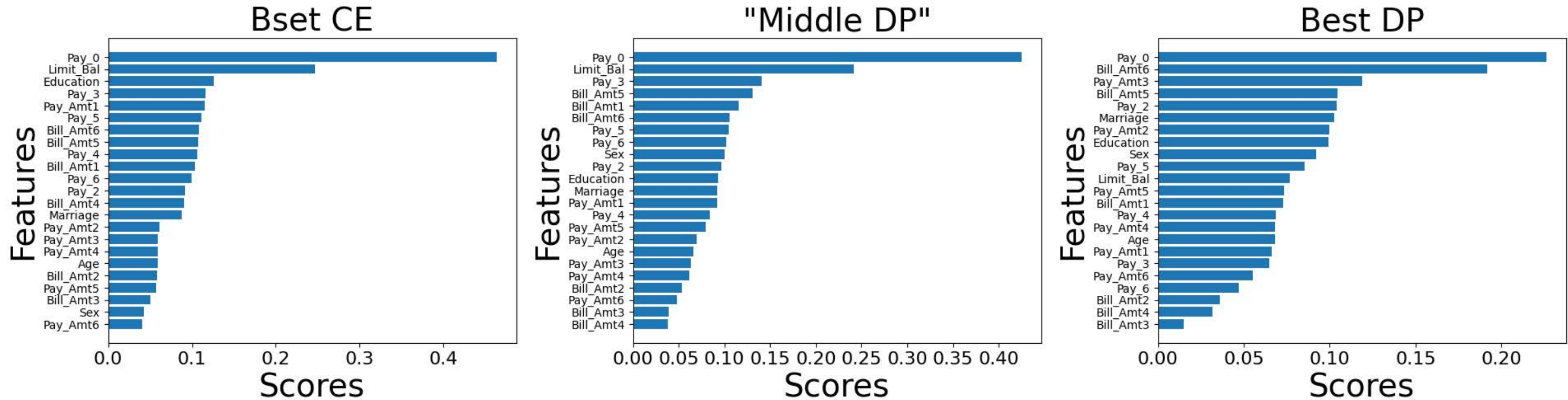}
        \label{fig:default_DP_merge_local}
    }
    \subfigure[Average local explanation of \textit{MONBM-CL} instantiation on the \textit{Dutch} dataset]{
        \includegraphics[width=0.48\textwidth]{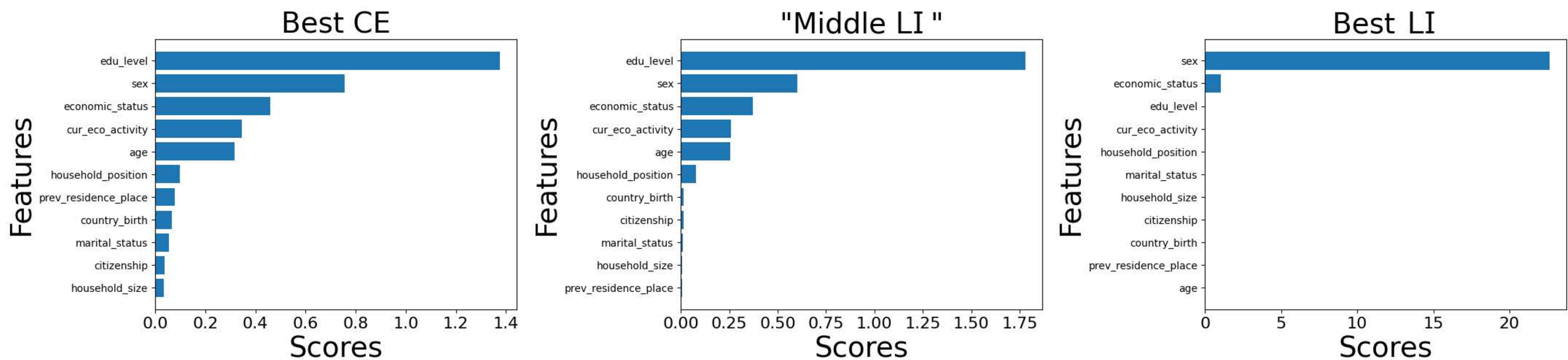}
        \label{fig:dutch_LU_merge_local}
    }
    \subfigure[Average local explanation of \textit{MONBM-CD} instantiation on the \textit{Dutch} dataset]{
        \includegraphics[width=0.48\textwidth]{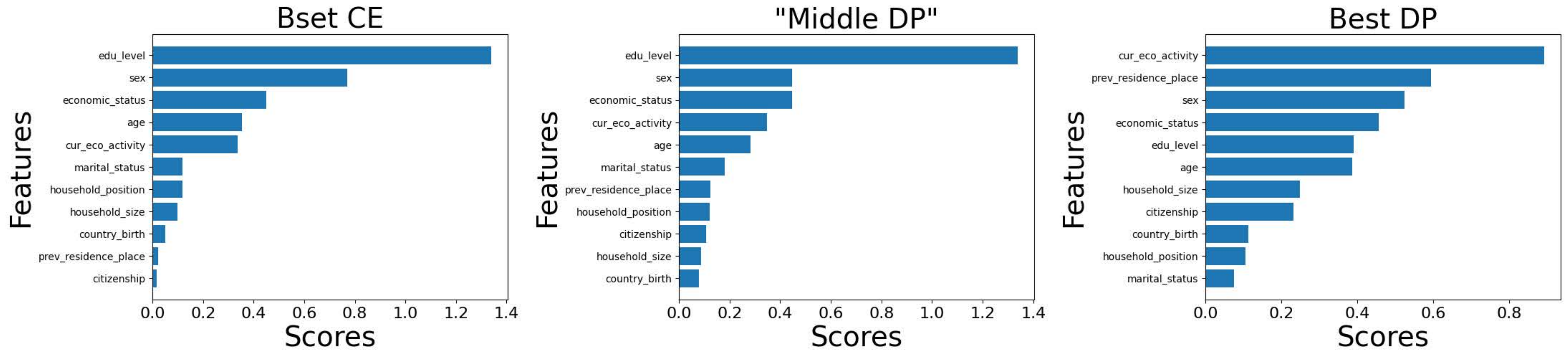}
        \label{fig:dutch_DP_merge_local}
    }
    \subfigure[Average local explanation of \textit{MONBM-CL} instantiation on the \textit{LSAT} dataset]{
        \includegraphics[width=0.48\textwidth]{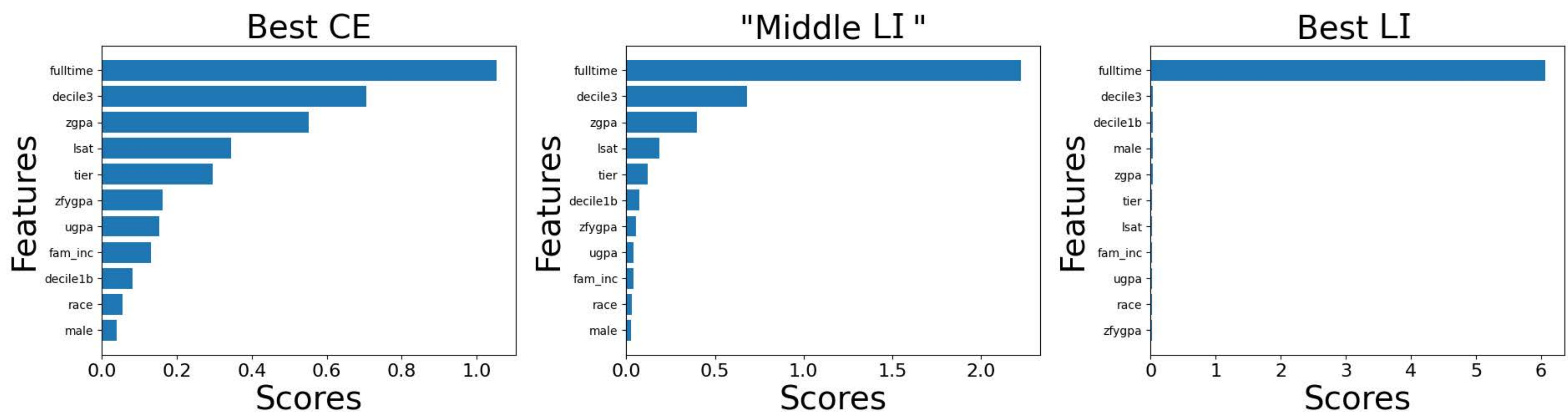}
        \label{fig:LSAT_LU_merge_local}
    }
    \subfigure[Average local explanation of \textit{MONBM-CD} instantiation on the \textit{LSAT} dataset]{
        \includegraphics[width=0.48\textwidth]{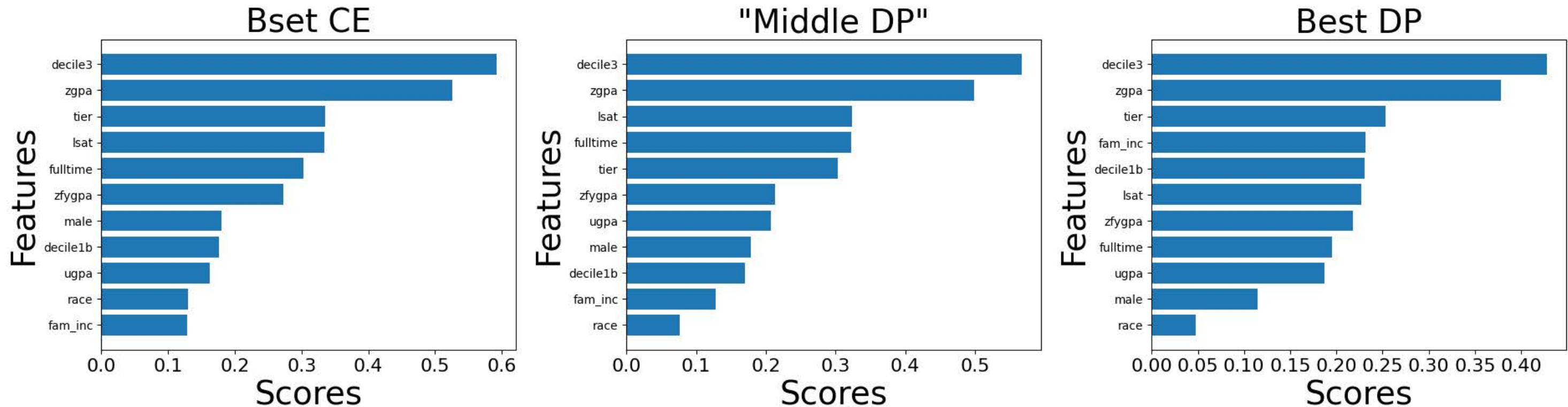}
        \label{fig:LSAT_DP_merge_local}
    }
    \caption{Comparison of the average local explanations to assess the impact of improved \textit{LI} on each feature. For each dataset, the three columns respectively display the average local explanation results of the model corresponding to: (1) the best \textit{CE} point, (2) the ``middle \textit{LI}'' or ``middle \textit{DP}'' point, and (3) the best \textit{LI} or best \textit{DP} point, obtained by the \textit{MONBM-CL} or \textit{MONBM-CD} instantiation in one arbitrary trial. For each local explanation, the y-axis represents the features and the x-axis represents the average of the absolute values of the shape function outputs corresponding to the features for all data points in the test set.}
    \label{fig:LU_local_explanation_sup}
\end{figure}

\begin{figure}[htbp]
    \centering
    \subfigure[Mean local explanation and standard deviation \textit{MONBM-CL} instantiation over 15 independent trials on the \textit{Adult} dataset]{
        \includegraphics[width=0.48\textwidth]{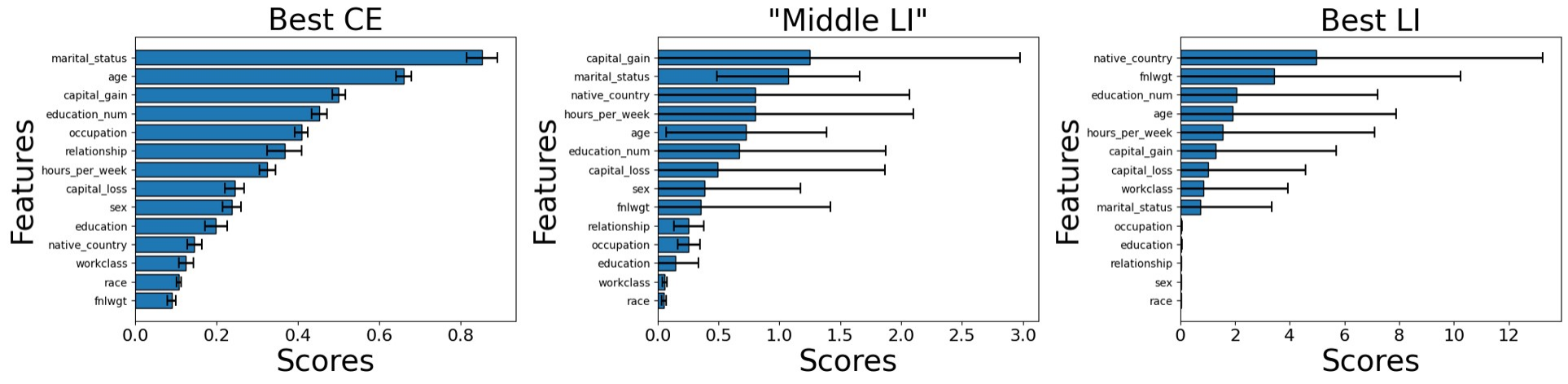}
        \label{fig:Adult_LU_merge_local_avg}
    }
    \subfigure[Mean local explanation and standard deviation \textit{MONBM-CD} instantiation over 15 independent trials on the \textit{Adult} dataset]{
        \includegraphics[width=0.485\textwidth]{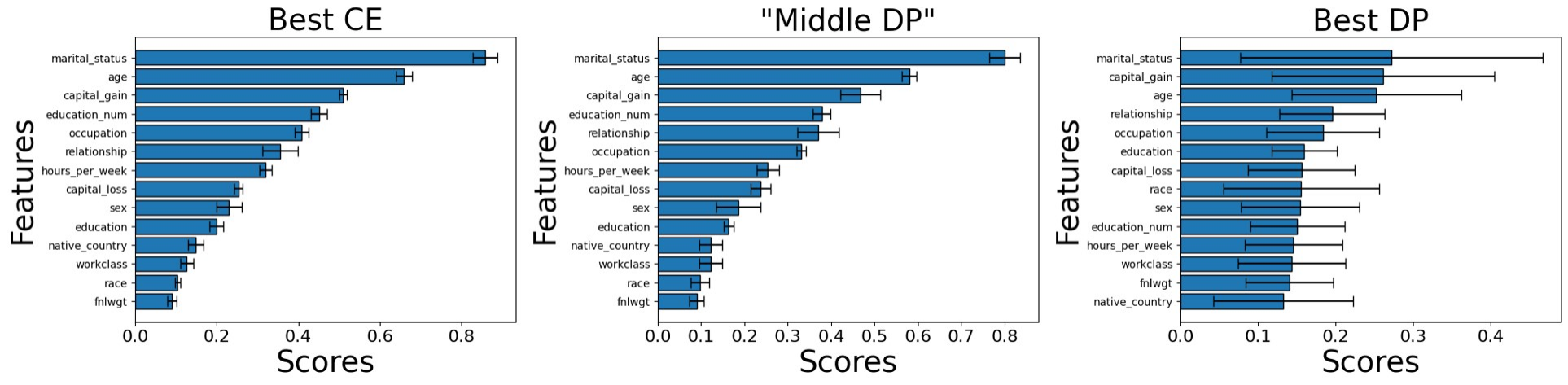}
        \label{fig:adult_DP_merge_local_avg}
    }
    \subfigure[Mean local explanation and standard deviation \textit{MONBM-CL} instantiation over 15 independent trials on the \textit{Bank} dataset]{
        \includegraphics[width=0.48\textwidth]{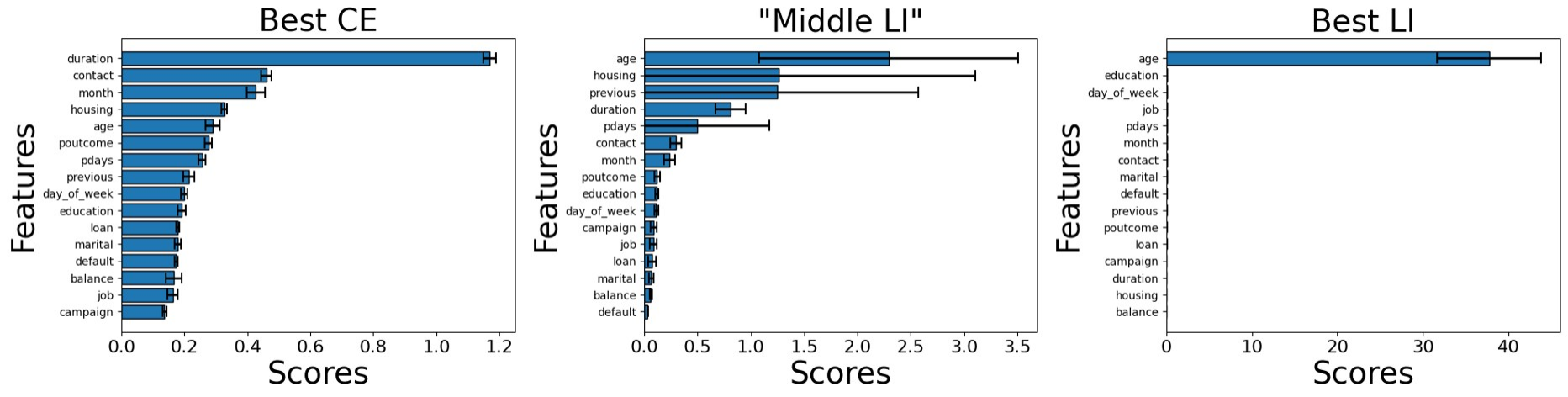}
        \label{fig:bank_LU_merge_local_avg}
    }
    \subfigure[Mean local explanation and standard deviation \textit{MONBM-CD} instantiation over 15 independent trials on the \textit{Bank} dataset]{
        \includegraphics[width=0.485\textwidth]{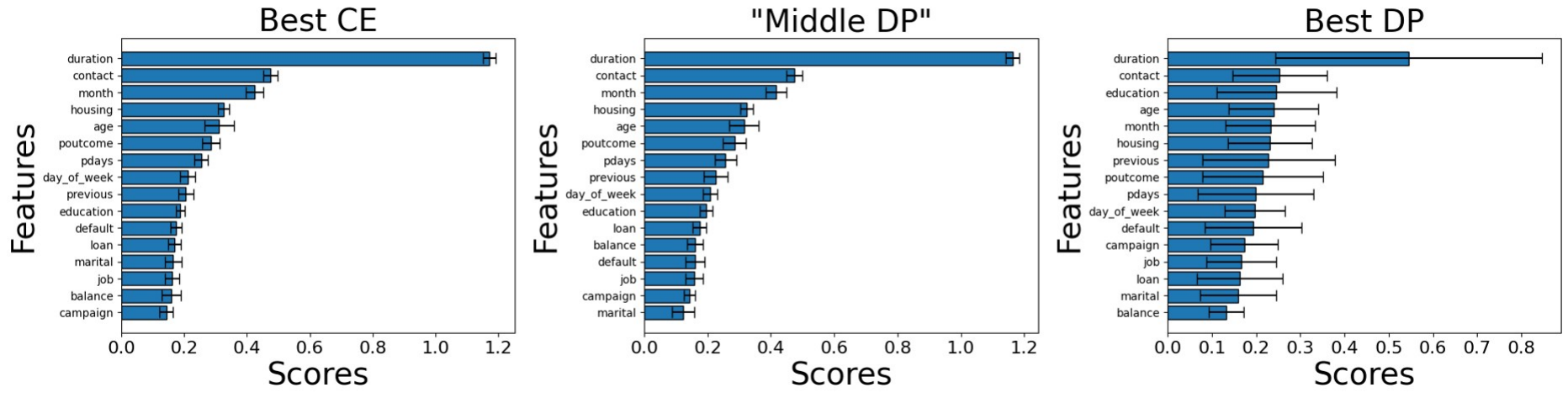}
        \label{fig:bank_DP_merge_local_avg}
    }
    \subfigure[Mean local explanation and standard deviation \textit{MONBM-CL} instantiation over 15 independent trials on the \textit{COMPAS} dataset]{
        \includegraphics[width=0.48\textwidth]{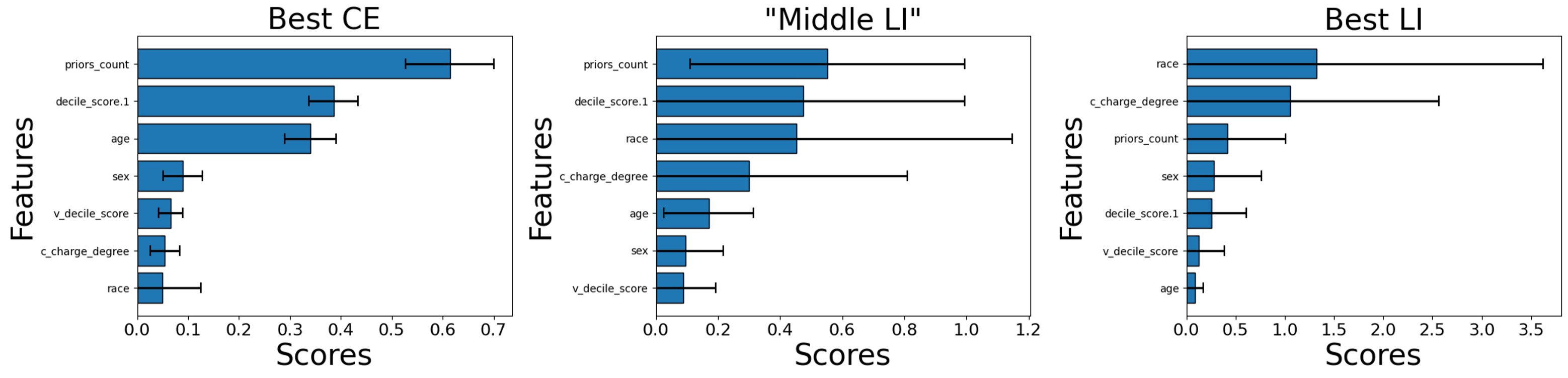}
        \label{fig:compas_LU_merge_local_avg}
    }
    \subfigure[Mean local explanation and standard deviation \textit{MONBM-CD} instantiation over 15 independent trials on the \textit{COMPAS} dataset]{
        \includegraphics[width=0.48\textwidth]{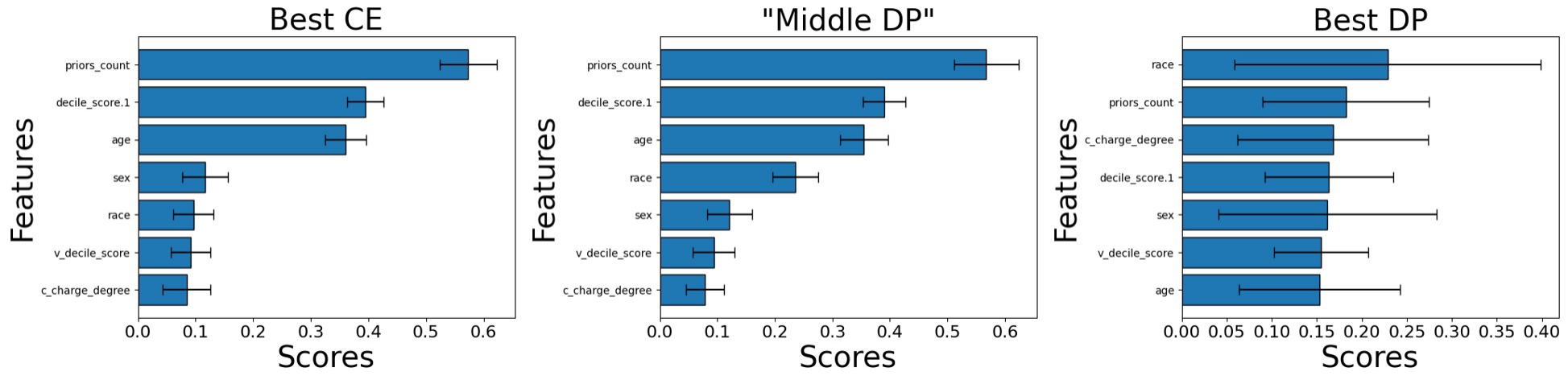}
        \label{fig:compas_DP_merge_local_avg}
    }
    \subfigure[Mean local explanation and standard deviation \textit{MONBM-CL} instantiation over 15 independent trials on the \textit{Default} dataset]{
        \includegraphics[width=0.48\textwidth]{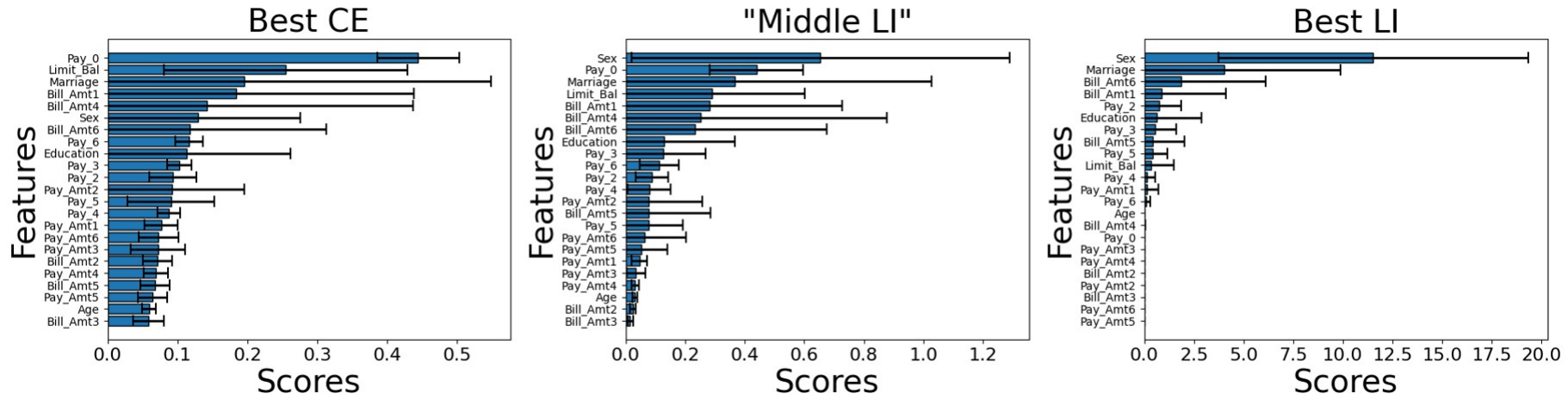}
        \label{fig:default_LU_merge_local_avg}
    }
    \subfigure[Mean local explanation and standard deviation \textit{MONBM-CD} instantiation over 15 independent trials on the \textit{Default} dataset]{
        \includegraphics[width=0.48\textwidth]{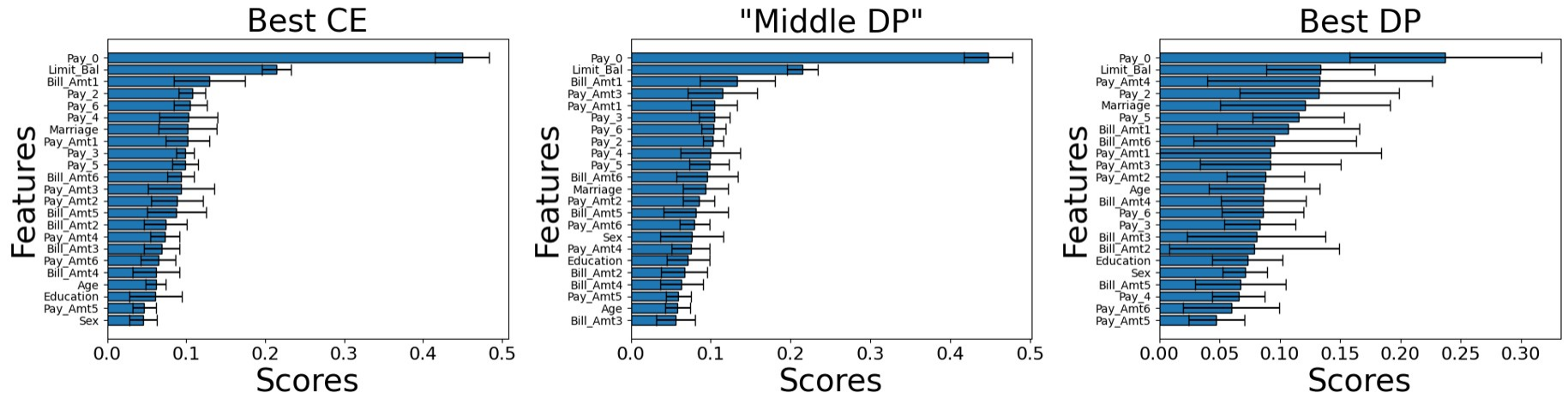}
        \label{fig:default_DP_merge_local_avg}
    }
    \subfigure[Mean local explanation and standard deviation \textit{MONBM-CL} instantiation over 15 independent trials on the \textit{Dutch} dataset]{
        \includegraphics[width=0.48\textwidth]{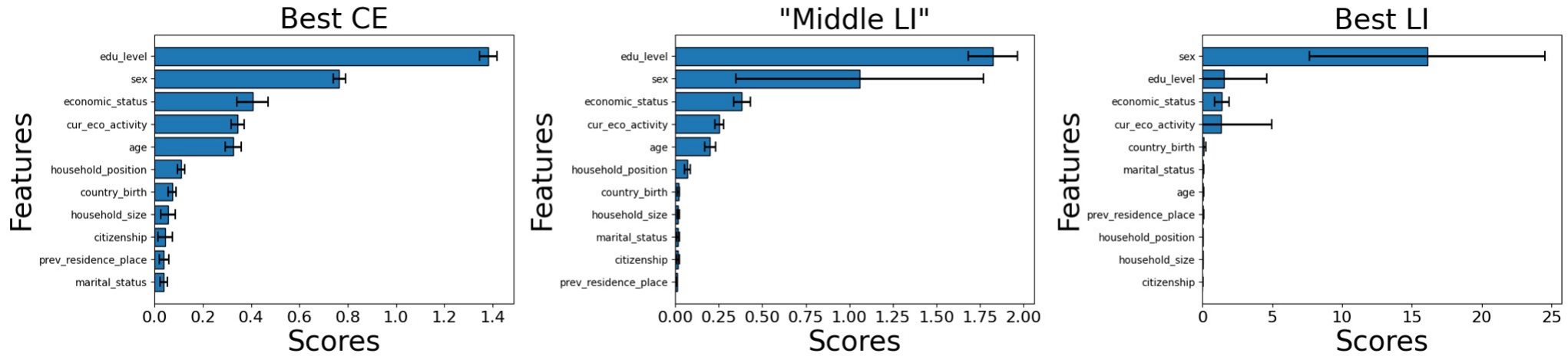}
        \label{fig:dutch_LU_merge_local_avg}
    }
    \subfigure[Mean local explanation and standard deviation \textit{MONBM-CD} instantiation over 15 independent trials on the \textit{Dutch} dataset]{
        \includegraphics[width=0.48\textwidth]{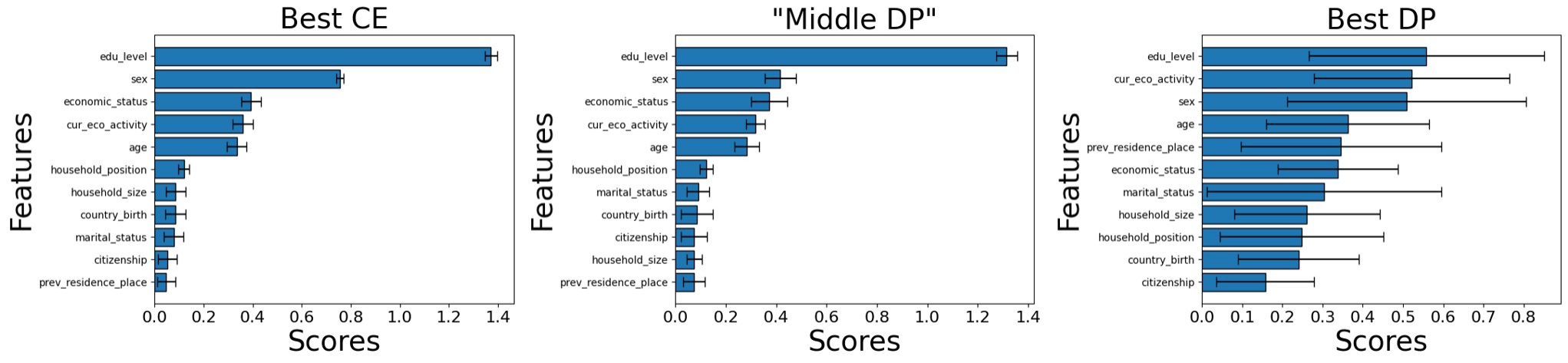}
        \label{fig:dutch_DP_merge_local_avg}
    }
    \subfigure[Mean local explanation and standard deviation \textit{MONBM-CL} instantiation over 15 independent trials on the \textit{LSAT} dataset]{
        \includegraphics[width=0.48\textwidth]{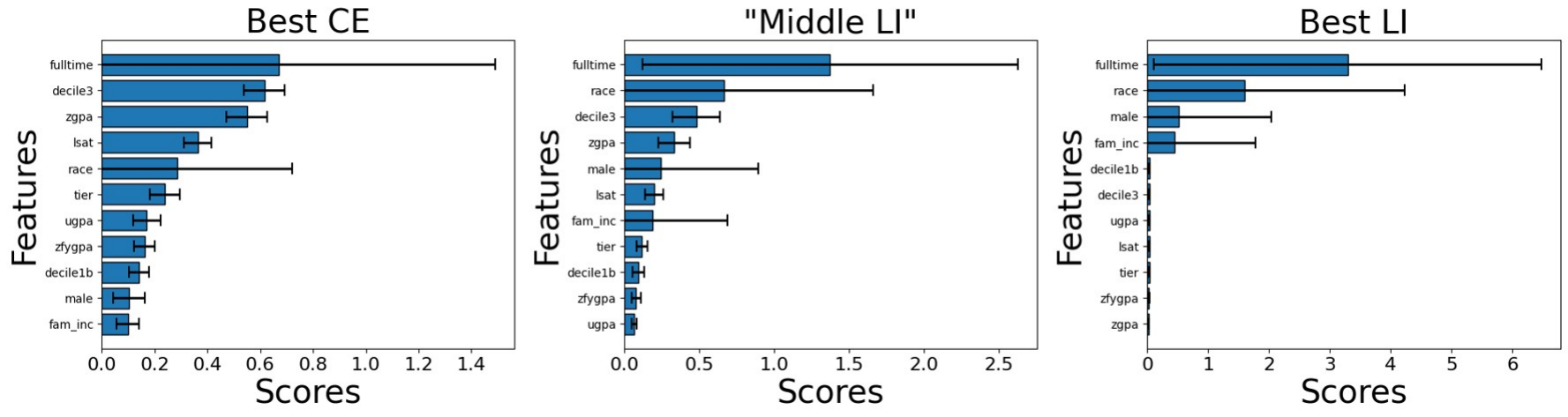}
        \label{fig:LSAT_LU_merge_local_avg}
    }
    \subfigure[Mean local explanation and standard deviation \textit{MONBM-CD} instantiation over 15 independent trials on the \textit{LSAT} dataset]{
        \includegraphics[width=0.48\textwidth]{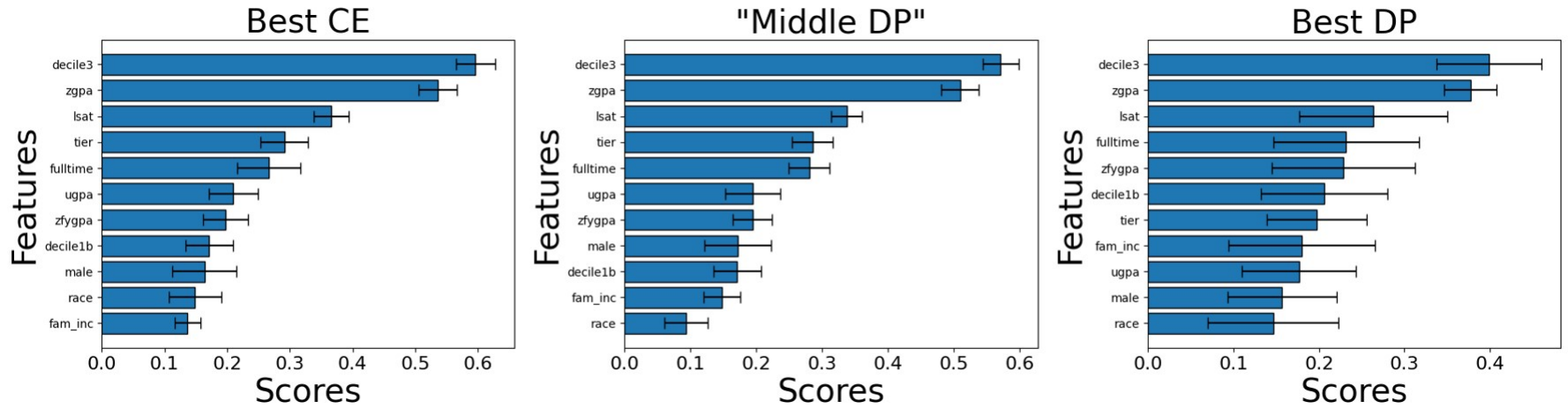}
        \label{fig:LSAT_DP_merge_local_avg}
    }
    \caption{Comparison of the average local explanations to assess the impact of improved \textit{LI} on each feature. For each dataset, the three columns respectively display the mean and standard deviation of the average local explanation results of the model corresponding to: (1) the best \textit{CE} point, (2) the ``middle \textit{LI}'' or ``middle \textit{DP}'' point, and (3) the best \textit{LI} or best \textit{DP} point, obtained by the \textit{MONBM-CL} or \textit{MONBM-CG} instantiation averaged over 15 independent trials. For each local explanation, the y-axis represents the features and the x-axis represents the average of the absolute values of the shape function outputs corresponding to the features for all data points in the test set, with error bars representing the standard deviation across the 15 trials.}
    \label{fig:LU_local_explanation_sup_avg}
\end{figure}

\begin{figure}[htbp]
    \centering
    \subfigure[Global explanation results of \textit{MONBM-CG} instantiation on the \textit{Adult} dataset]{
    \includegraphics[width=\textwidth]{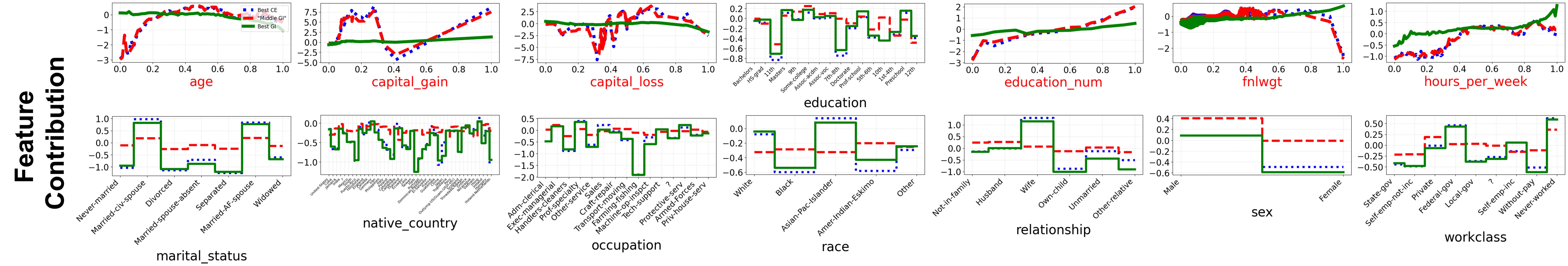}
    \label{fig:adult_GU_merge_global}
    }
    \subfigure[Global explanation results of \textit{MONBM-CG} instantiation on the \textit{COMPAS} dataset]{
    \includegraphics[width=\textwidth]{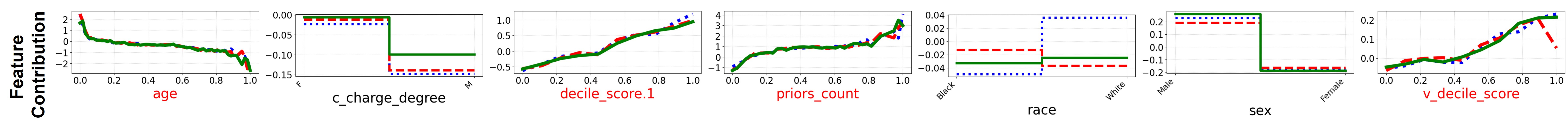}
    \label{fig:compas_GU_merge_global}
    }
    \subfigure[Global explanation results of \textit{MONBM-CG} instantiation on the \textit{Default} dataset]{
    \includegraphics[width=\textwidth]{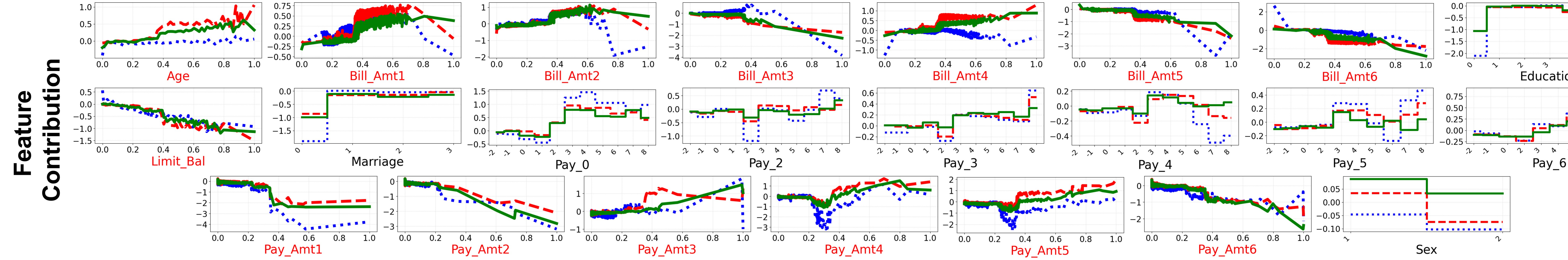}
    \label{fig:default_GU_merge_global}
    }
    \subfigure[Global explanation results of \textit{MONBM-CG} instantiation on the \textit{Dutch} dataset]{
    \includegraphics[width=\textwidth]{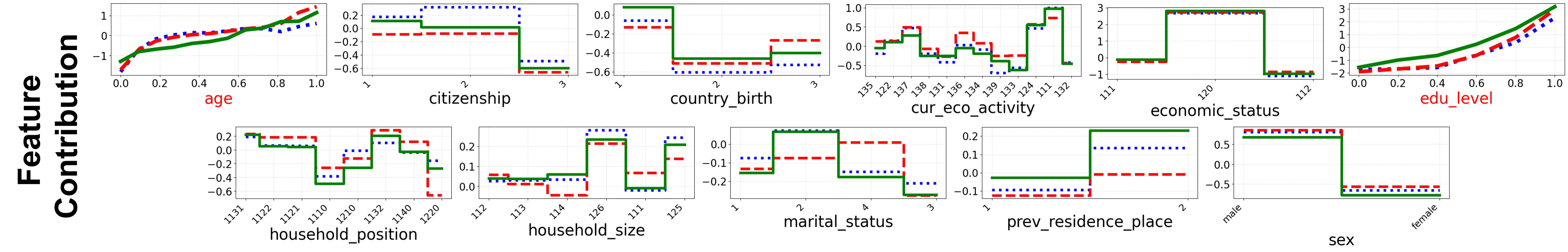}
    \label{fig:dutch_GU_merge_global}
    }
    \subfigure[Global explanation results of \textit{MONBM-CG} instantiation on the \textit{LSAT} dataset]{
    \includegraphics[width=\textwidth]{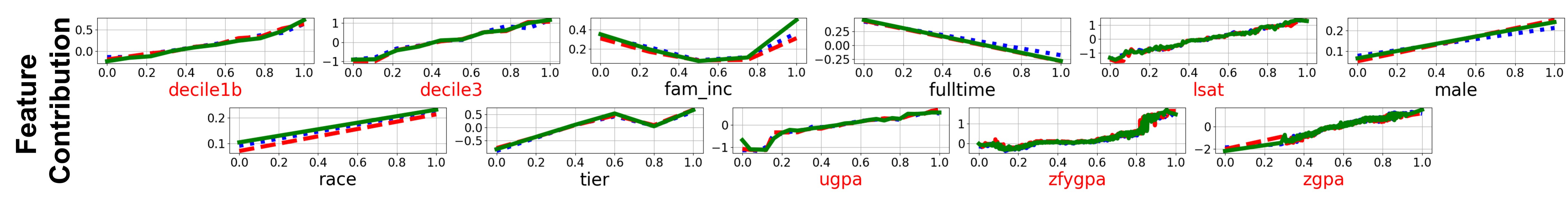}
    \label{fig:LSAT_GU_merge_global}
    }
    \caption{Comparison of global explanations to assess the impact of improved \textit{GI} on each feature. For each dataset, the blue, red, and green lines represent the global explanation results of the model corresponding to the best \textit{CE}, ``middle \textit{GI},'' and best \textit{GI} points obtained by the \textit{MONBM-CG} instantiation in one arbitrary trial, respectively. Each subfigure corresponds to the global explanation of a feature, with the x-axis representing feature values and the y-axis showing the outputs of the shape functions. The continuous features are highlighted in red.}
    \label{fig:GU_global_explanation_sup}
\end{figure}

\begin{figure*}[htbp]
    \centering
    \subfigure[Mean global explanation results of \textit{MONBM-CG} instantiation over 15 independent trials on the \textit{Adult} dataset]{
        \includegraphics[width=\textwidth]{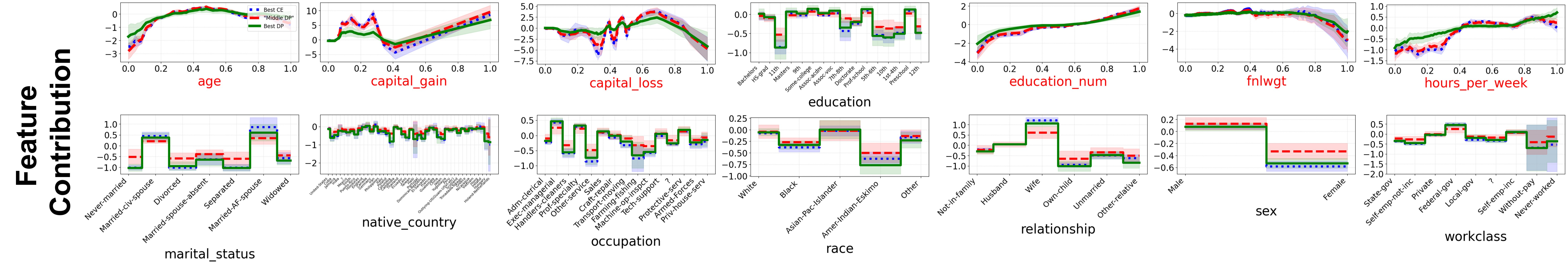}
        \label{fig:adult_GI_merge_global_15avg}
    }
    \subfigure[Mean global explanation results of \textit{MONBM-CG} instantiation over 15 independent trials on the \textit{Bank} dataset]{
        \includegraphics[width=\textwidth]{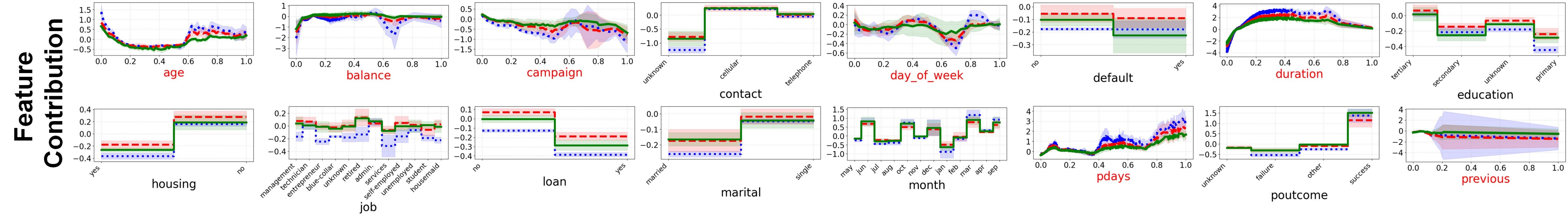}
        \label{fig:bank_GI_merge_global_15avg}
    }
    \subfigure[Mean global explanation results of \textit{MONBM-CG} instantiation over 15 independent trials on the \textit{COMPAS} dataset]{
        \includegraphics[width=0.6\textwidth]{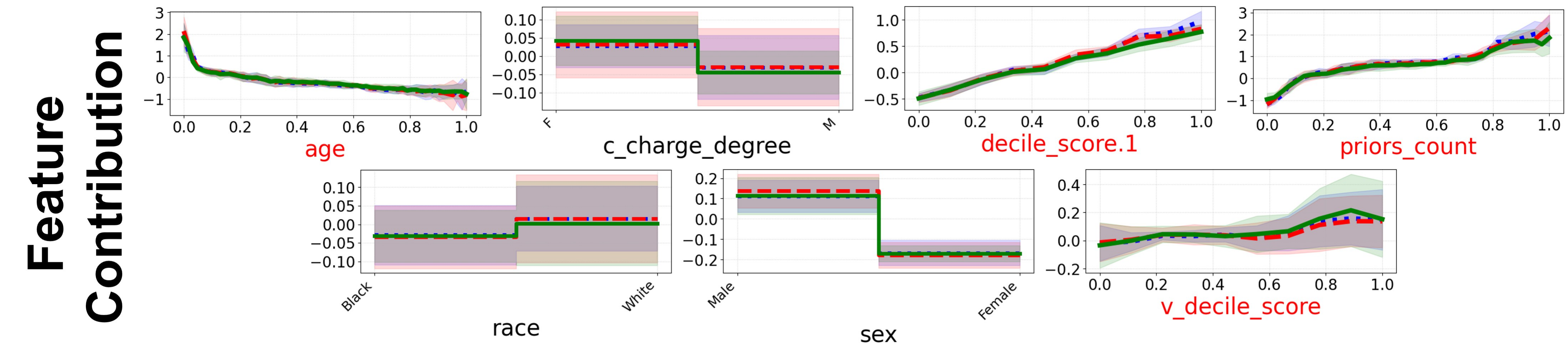}
        \label{fig:compas_GI_merge_global_15avg}
    }
    \subfigure[Mean global explanation results of \textit{MONBM-CG} instantiation over 15 independent trials on the \textit{Default} dataset]{
        \includegraphics[width=\textwidth]{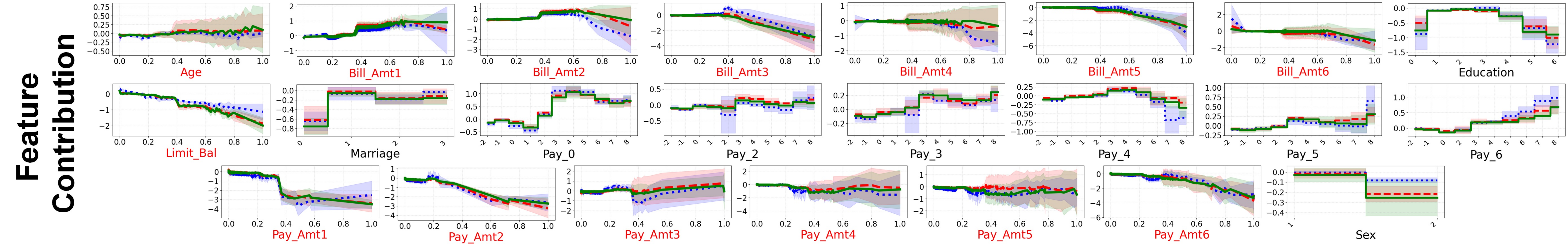}
        \label{fig:default_GI_merge_global_15avg}
    }
    \subfigure[Mean global explanation results of \textit{MONBM-CG} instantiation over 15 independent trials on the \textit{Dutch} dataset]{
        \includegraphics[width=\textwidth]{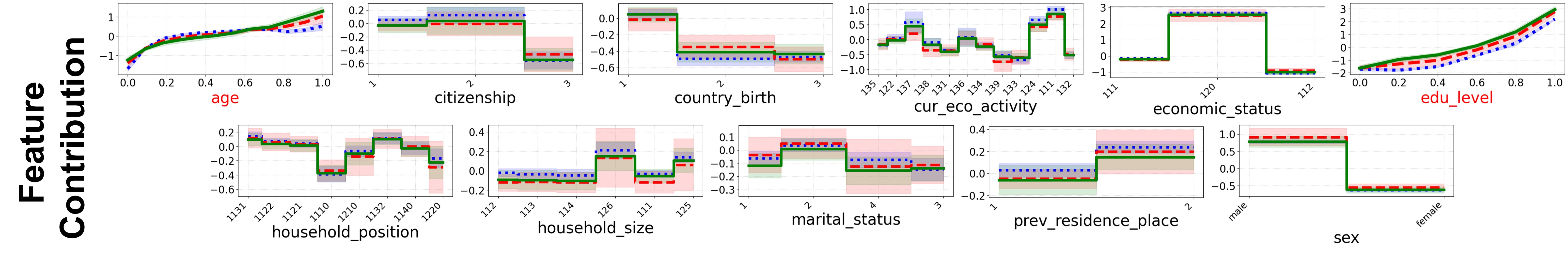}
        \label{fig:dutch_GI_merge_global_15avg}
    }
    \subfigure[Mean global explanation results of \textit{MONBM-CG} instantiation over 15 independent trials on the \textit{LSAT} dataset]{
        \includegraphics[width=\textwidth]{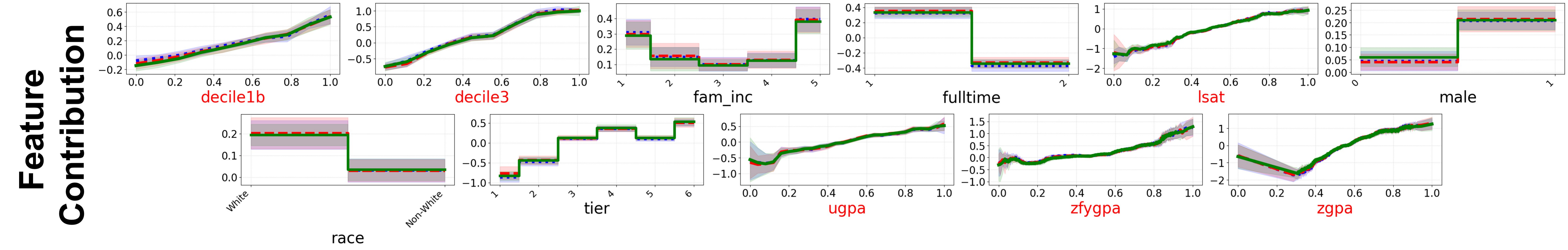}
        \label{fig:lsat_GI_merge_global_15avg}
    }
    \caption{Comparison of global explanations to assess the impact of improved fairness on each feature. For each dataset, the blue, red, and green lines and their corresponding shaded areas represent the mean global explanation results and standard deviations of the models corresponding to the best \textit{CE}, ``middle \textit{GI}'', and best \textit{GI} points obtained by the \textit{MONBM-CG} instantiation averaged over 15 independent trials, respectively. Each subfigure corresponds to the global explanation of a feature, with the x-axis representing feature values and the y-axis showing the outputs of the shape functions. The continuous features are highlighted in red.}
    \label{fig:GI_global_explanation_avg_sup}
\end{figure*}

\begin{figure}[htbp]
    \centering
    \subfigure[Average local explanation of \textit{MONBM-CG} instantiation on \textit{Adult} dataset]{
    \includegraphics[width=0.48\textwidth]{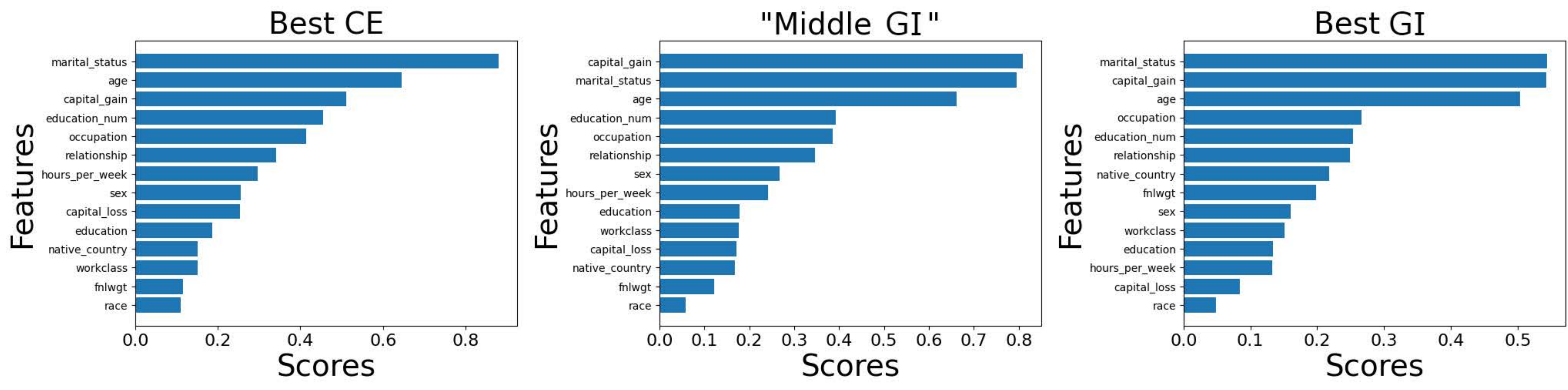}
    \label{fig:adult_GU_merge_local}
    }
    \subfigure[Average local explanation of \textit{MONBM-CG} instantiation on \textit{COMPAS} dataset]{
    \includegraphics[width=0.48\textwidth]{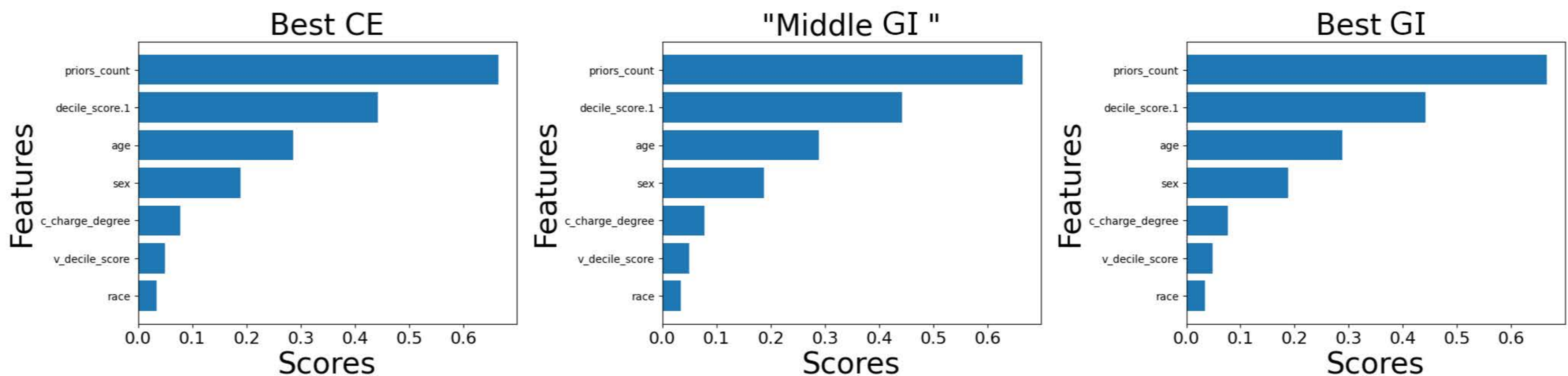}
    \label{fig:compas_GU_merge_local}
    }
    \subfigure[Average local explanation of \textit{MONBM-CG} instantiation on \textit{Default} dataset]{
    \includegraphics[width=0.48\textwidth]{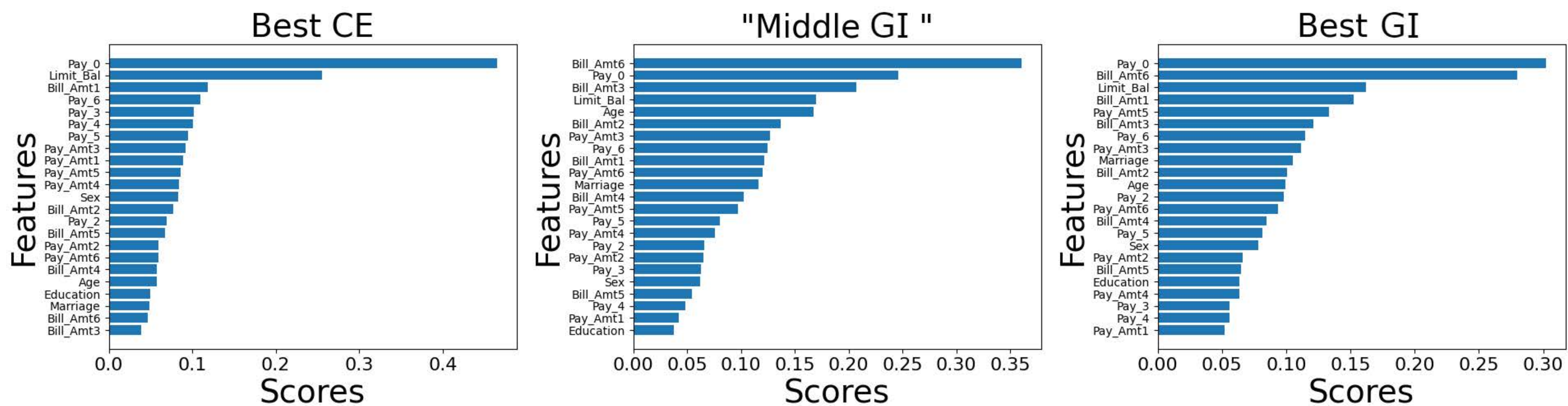}
    \label{fig:default_GU_merge_local}
    }
    \subfigure[Average local explanation of \textit{MONBM-CG} instantiation on \textit{Dutch} dataset]{
    \includegraphics[width=0.48\textwidth]{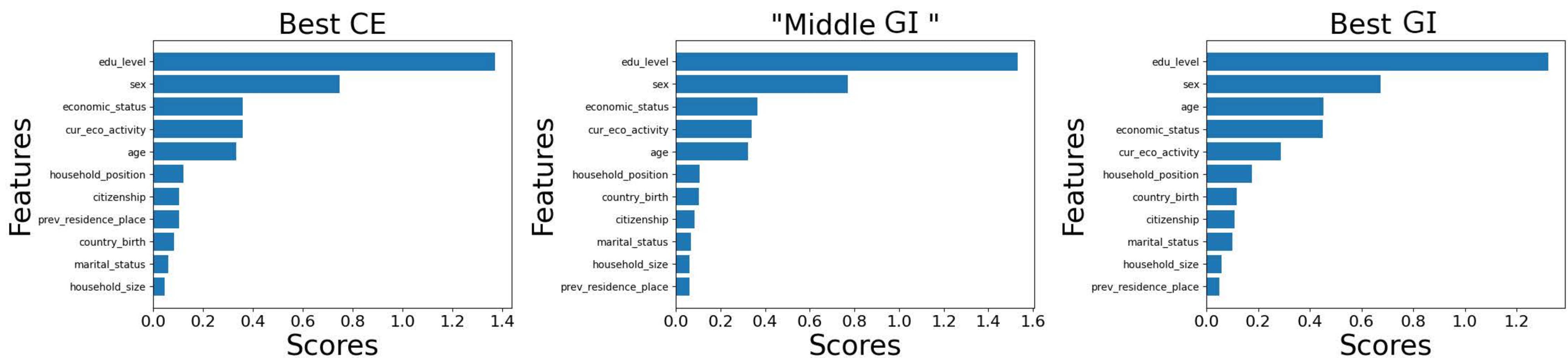}
    \label{fig:dutch_GU_merge_local}
    }
    \subfigure[Average local explanation of \textit{MONBM-CG} instantiation on \textit{LSAT} dataset]{
    \includegraphics[width=0.48\textwidth]{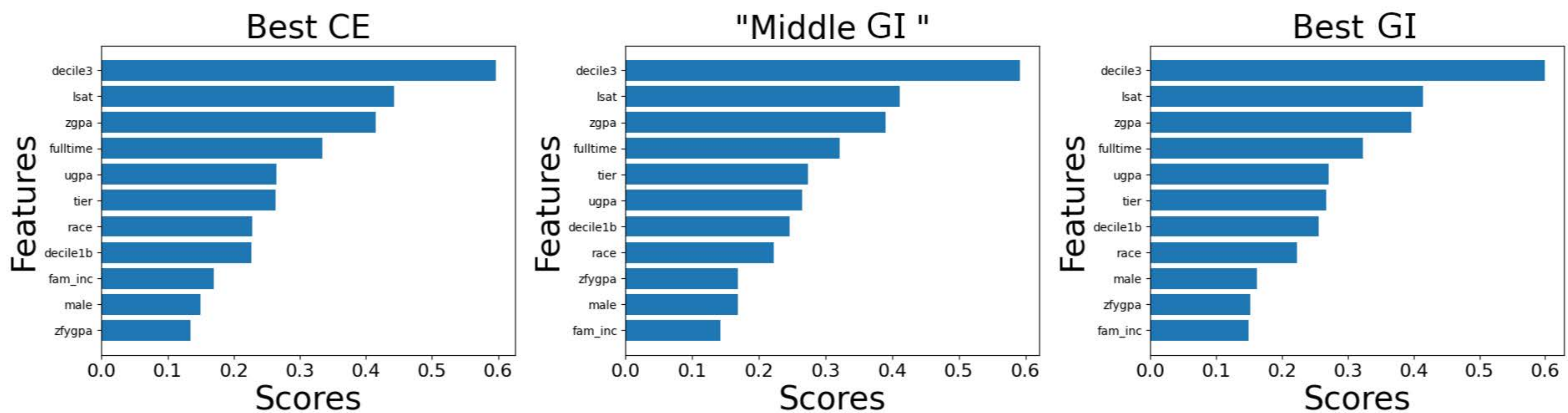}
    \label{fig:LSAT_GU_merge_local}
    }
    \caption{Comparison of the average local explanations to assess the impact of improved \textit{GI} on each feature. For each dataset, the three columns represent the average local explanation results of the model corresponding to the best \textit{CE}, ``middle \textit{GI},'' and best \textit{GI} points obtained by the \textit{MONBM-CG} instantiation in one arbitrary trial, respectively. For each local explanation, the y-axis represents the features and the x-axis represents the average of the absolute values of the shape function outputs corresponding to the features for all data points in the test set.}
    \label{fig:GU_local_explanation_sup}
\end{figure}

\begin{figure}[htbp]
    \centering
    \subfigure[Mean local explanation and standard deviation of\textit{MONBM-CG} instantiation over 15 independent trials on the \textit{Adult} dataset]{
    \includegraphics[width=0.48\textwidth]{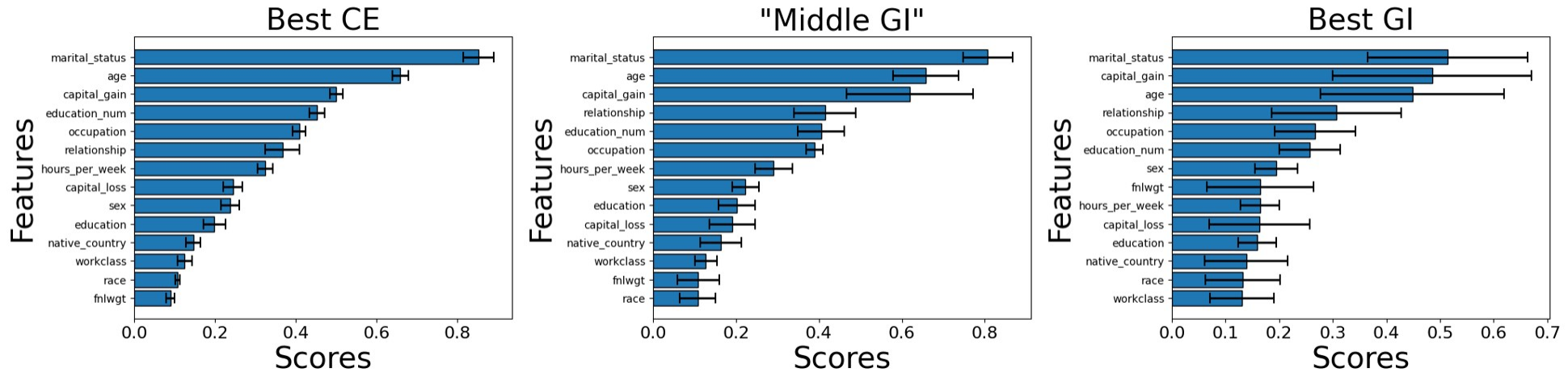}
    \label{fig:adult_GU_merge_local_avg}
    }
    \subfigure[Mean local explanation and standard deviation of\textit{MONBM-CG} instantiation over 15 independent trials on the \textit{Bank} dataset]{
    \includegraphics[width=0.48\textwidth]{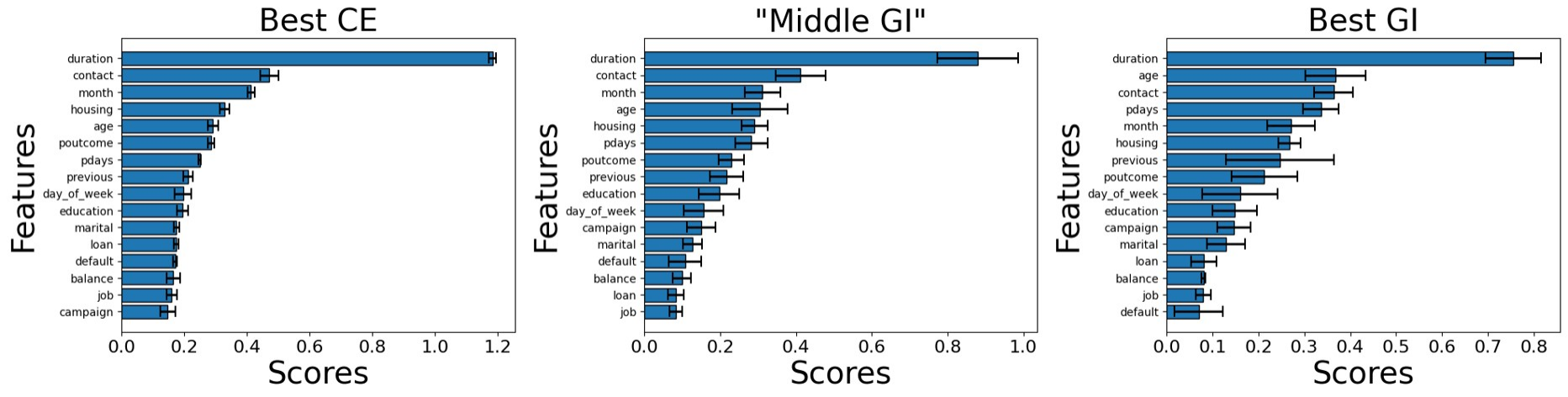}
    \label{fig:bank_GU_merge_local_avg}
    }
    \subfigure[Mean local explanation and standard deviation of\textit{MONBM-CG} instantiation over 15 independent trials on the \textit{COMPAS} dataset]{
    \includegraphics[width=0.48\textwidth]{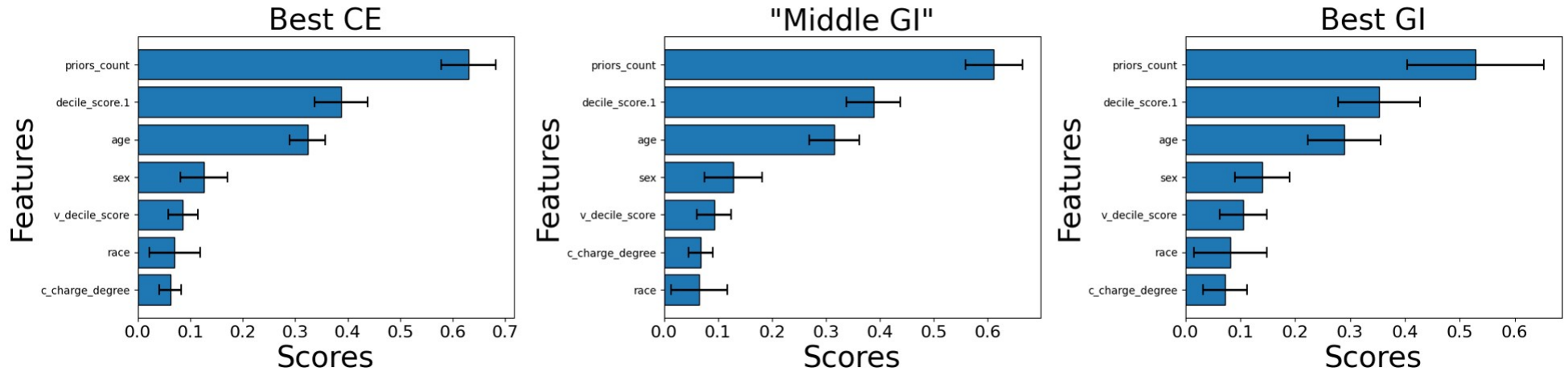}
    \label{fig:compas_GU_merge_local_avg}
    }
    \subfigure[Mean local explanation and standard deviation of\textit{MONBM-CG} instantiation over 15 independent trials on the \textit{Default} dataset]{
    \includegraphics[width=0.48\textwidth]{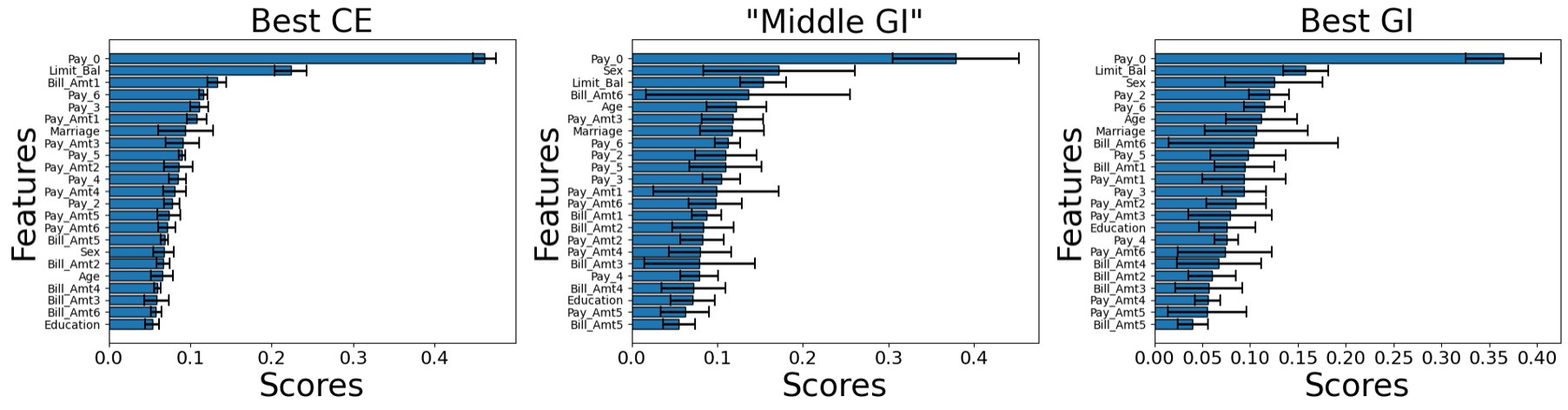}
    \label{fig:default_GU_merge_local_avg}
    }
    \subfigure[Mean local explanation and standard deviation of\textit{MONBM-CG} instantiation over 15 independent trials on the \textit{Dutch} dataset]{
    \includegraphics[width=0.48\textwidth]{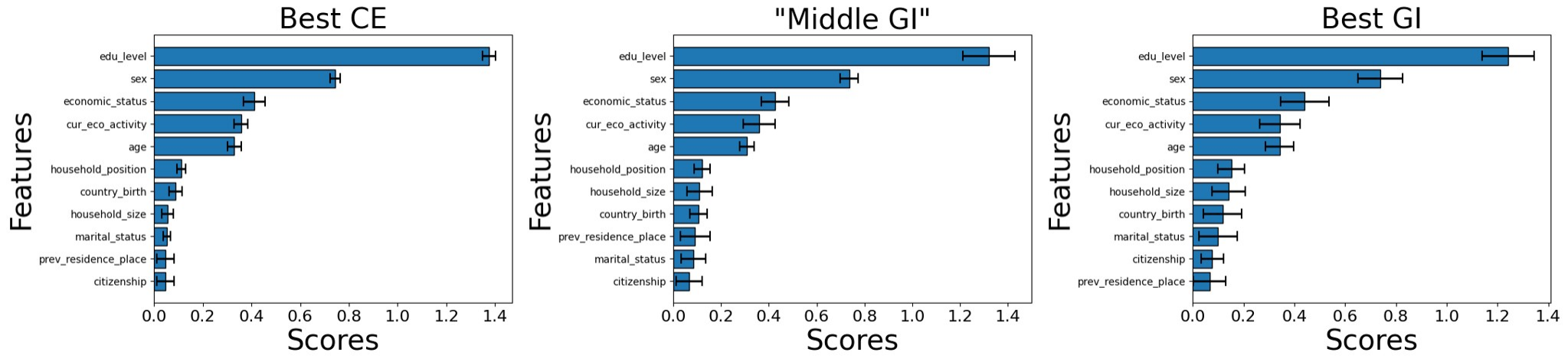}
    \label{fig:dutch_GU_merge_local_avg}
    }
    \subfigure[Mean local explanation and standard deviation of\textit{MONBM-CG} instantiation over 15 independent trials on the \textit{LSAT} dataset]{
    \includegraphics[width=0.48\textwidth]{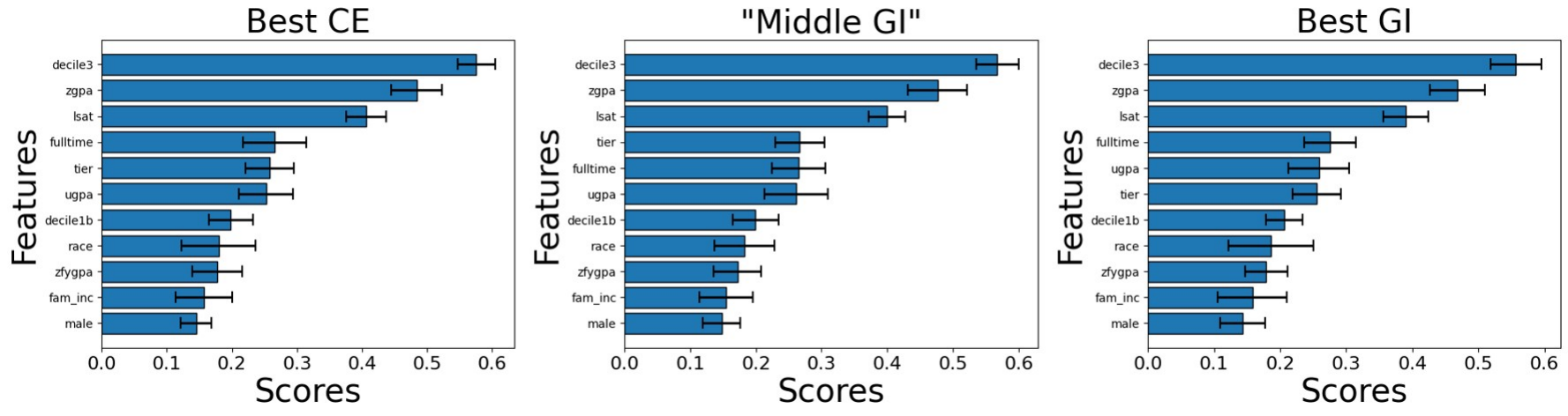}
    \label{fig:LSAT_GU_merge_local_avg}
    }
    \caption{Comparison of the average local explanations to assess the impact of improved \textit{GI} on each feature. For each dataset, the three columns represent the mean and standard deviation of the average local explanation results of the model corresponding to the best \textit{CE}, ``middle \textit{GI},'' and best \textit{GI} points obtained by the \textit{MONBM-CG} instantiation averaged over 15 independent trials, respectively. For each local explanation, the y-axis represents the features and the x-axis represents the average of the absolute values of the shape function outputs corresponding to the features for all data points in the test set, with error bars representing the standard deviation across the 15 trials.}
    \label{fig:GU_local_explanation_sup_avg}
\end{figure}

\begin{figure}[htbp]
    \centering
    \subfigure[Average local explanation of \textit{MONBM-CDLG} instantiation on the \textit{Bank} dataset]{
        \includegraphics[width=0.48\textwidth]{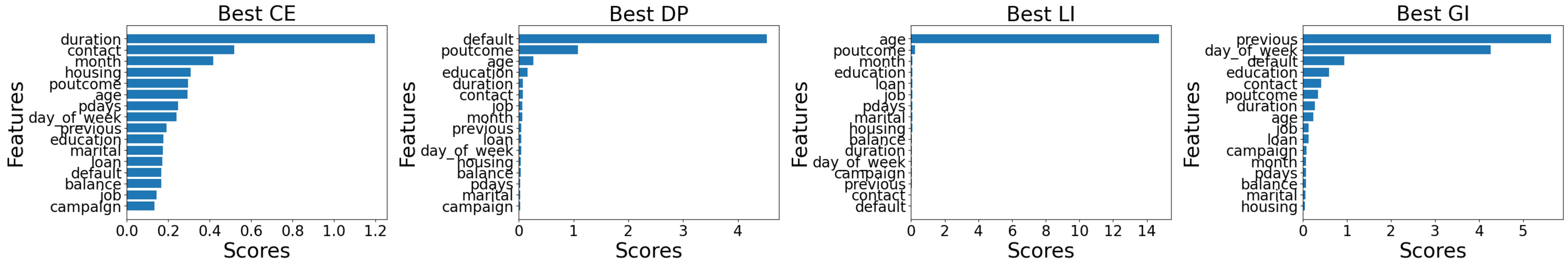}
        \label{fig:bank_4obj_merge_local}
    }
    \subfigure[Average local explanation of \textit{MONBM-CDLG} instantiation on the \textit{COMPAS} dataset]{
        \includegraphics[width=0.48\textwidth]{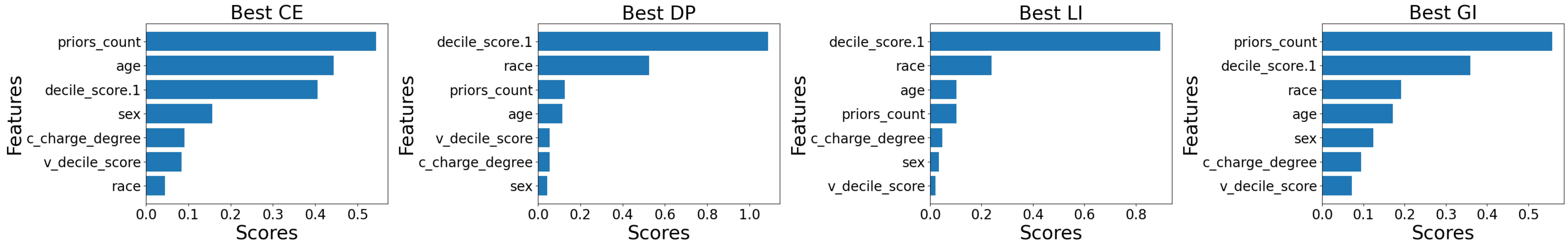}
        \label{fig:compas_4obj_merge_local}
    }
    \subfigure[Average local explanation of \textit{MONBM-CDLG} instantiation on the \textit{Default} dataset]{
        \includegraphics[width=0.48\textwidth]{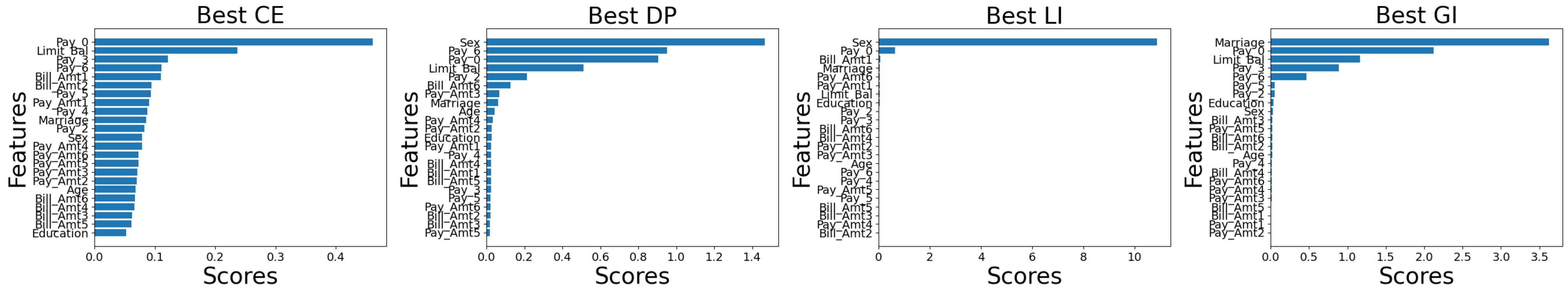}
        \label{fig:default_4obj_merge_local}
    }
    \subfigure[Average local explanation of \textit{MONBM-CDLG} instantiation on the \textit{LSAT} dataset]{
        \includegraphics[width=0.48\textwidth]{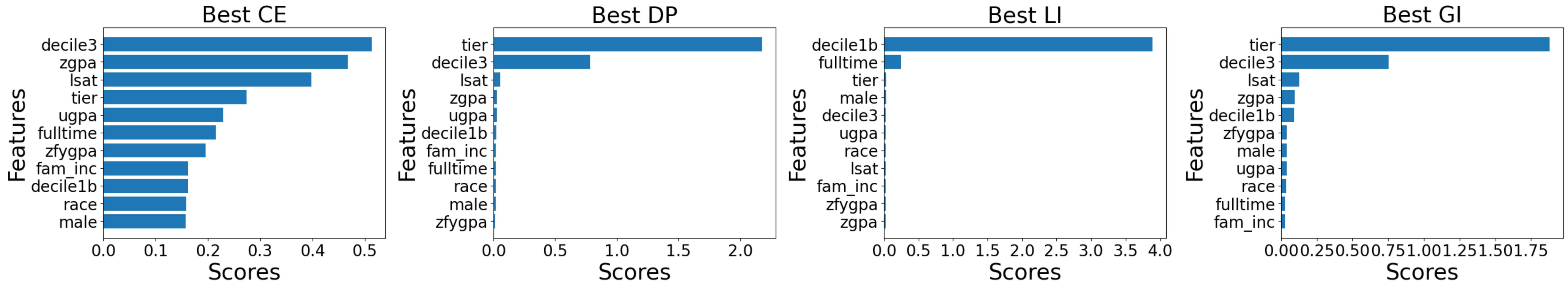}
        \label{fig:LSAT_4obj_merge_local}
    }
    \caption{Comparison of the average local explanations to explore the interrelationships among the four objectives. For each dataset, the four columns represent the average local explanation results of the model corresponding to the best \textit{CE}, best \textit{DP}, best \textit{LI}, and best \textit{GI} points obtained by the \textit{MONBM-CDLG} instantiation in one arbitrary trial, respectively. For each local explanation, the y-axis represents the features and the x-axis represents the average of the absolute values of the shape function outputs corresponding to the features for all data points in the test set.}
    \label{fig:4obj_local_explanation_sup} 
\end{figure}

\begin{figure}[htbp]
    \centering
    \subfigure[Mean local explanation and standard deviation of \textit{MONBM-CDLG} instantiation over 15 independent trials on the \textit{Adult} dataset]{
        \includegraphics[width=0.48\textwidth]{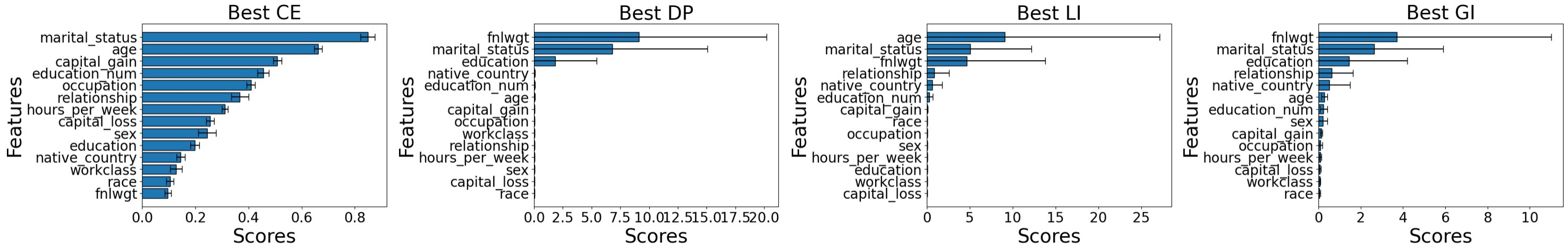}
        \label{fig:adult_4obj_merge_local_avg}
    }
    \subfigure[Mean local explanation and standard deviation of \textit{MONBM-CDLG} instantiation over 15 independent trials on the \textit{Bank} dataset]{
        \includegraphics[width=0.48\textwidth]{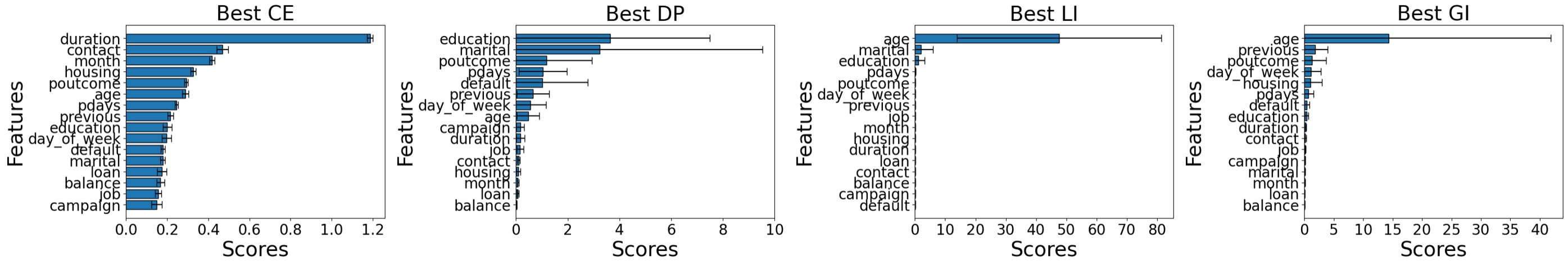}
        \label{fig:bank_4obj_merge_local_avg}
    }
    \subfigure[Mean local explanation and standard deviation of \textit{MONBM-CDLG} instantiation over 15 independent trials on the \textit{COMPAS} dataset]{
        \includegraphics[width=0.48\textwidth]{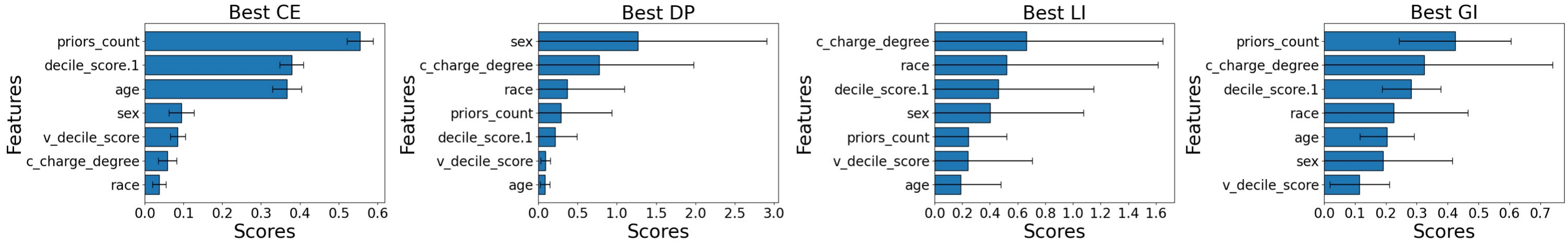}
        \label{fig:compas_4obj_merge_local_avg}
    }
    \subfigure[Mean local explanation and standard deviation of \textit{MONBM-CDLG} instantiation over 15 independent trials on the \textit{Default} dataset]{
        \includegraphics[width=0.48\textwidth]{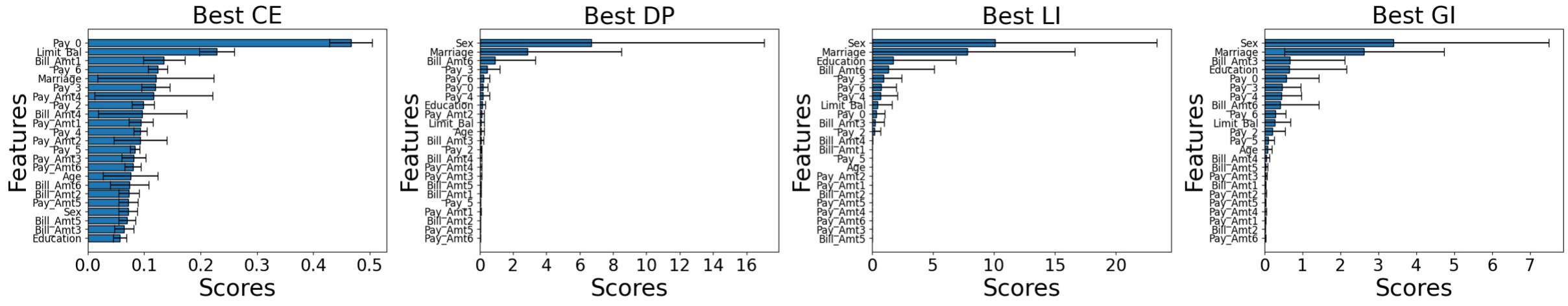}
        \label{fig:default_4obj_merge_local_avg}
    }
    \subfigure[Mean local explanation and standard deviation of \textit{MONBM-CDLG} instantiation over 15 independent trials on the \textit{Dutch} dataset]{
        \includegraphics[width=0.48\textwidth]{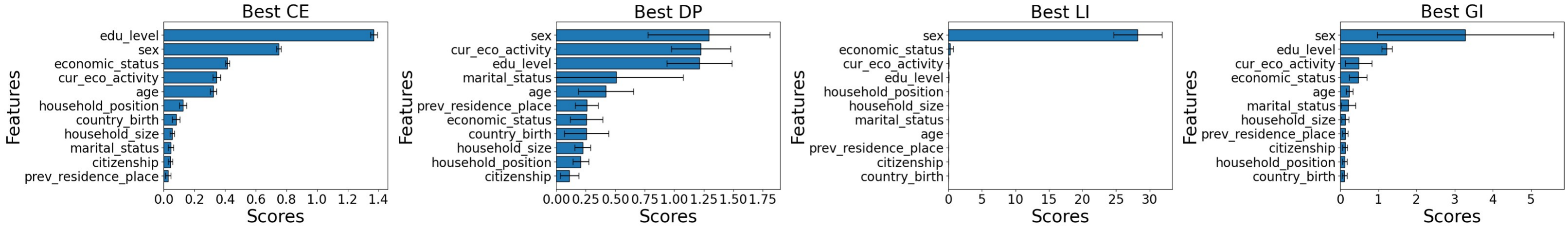}
        \label{fig:dutch_4obj_merge_local_avg}
    }
    \subfigure[Mean local explanation and standard deviation of \textit{MONBM-CDLG} instantiation over 15 independent trials on the \textit{LSAT} dataset]{
        \includegraphics[width=0.48\textwidth]{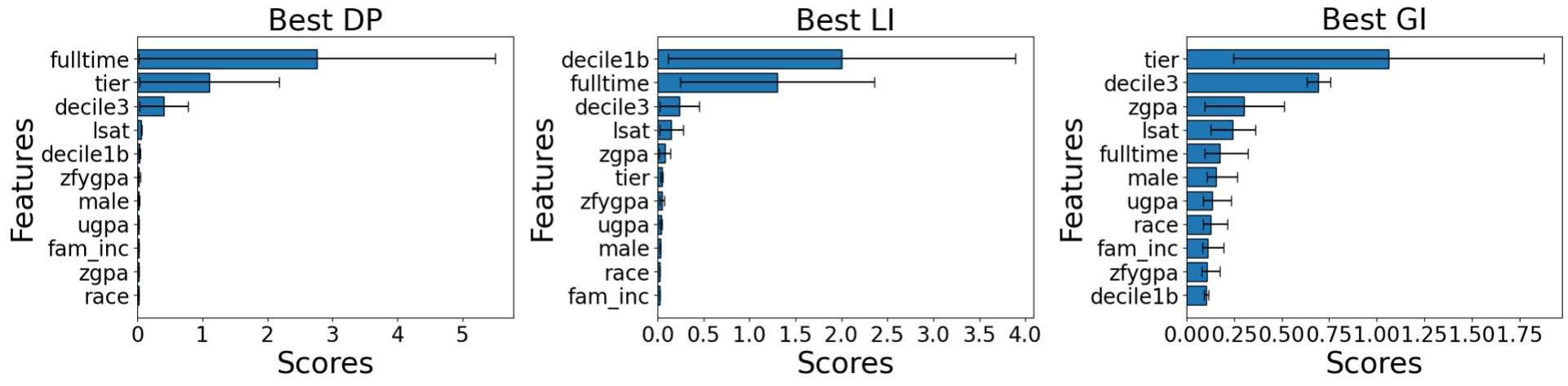}
        \label{fig:LSAT_4obj_merge_local_avg}
    }
    
    \caption{Comparison of the average local explanations to explore the interrelationships among the four objectives. For each dataset, the four columns represent the mean and standard deviation of the average local explanation results of the model corresponding to the best \textit{CE}, best \textit{DP}, best \textit{LI}, and best \textit{GI} points obtained by the \textit{MONBM-CDLG} instantiation averaged over 15 independent trials, respectively. For each local explanation, the y-axis represents the features and the x-axis represents the average of the absolute values of the shape function outputs corresponding to the features for all data points in the test set, with error bars representing the standard deviation across the 15 trials.}
    \label{fig:4obj_local_explanation_sup_avg}
\end{figure}

\begin{figure*}[h]
    \centering
    \subfigure[Distribution of \textit{MONBM-CDLG} solutions on the \textit{Bank} dataset]{
        \includegraphics[width=\textwidth]{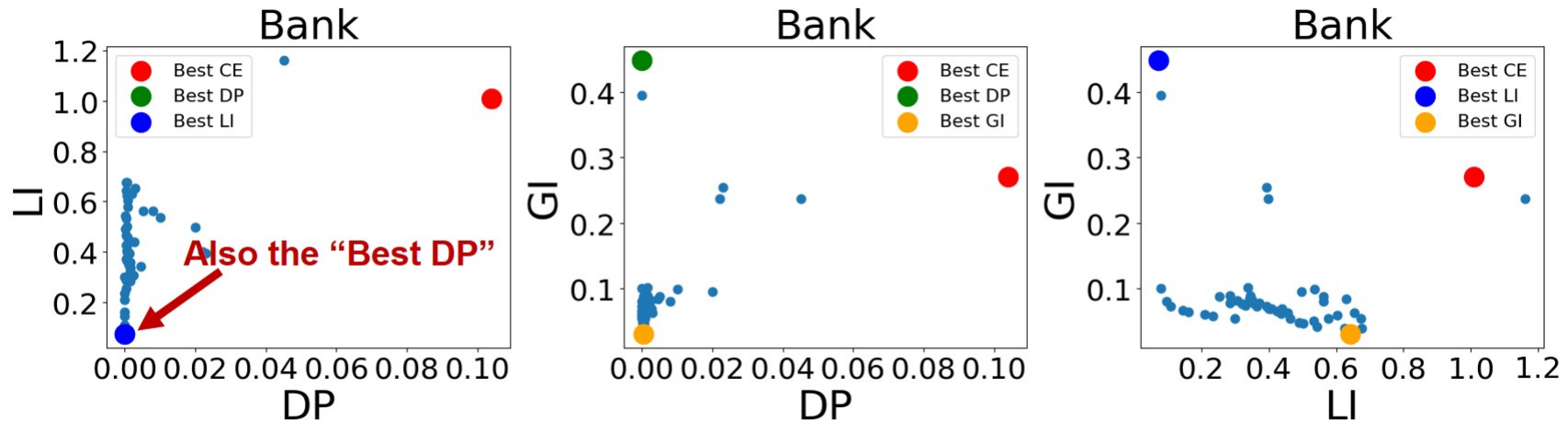}
        \label{fig:bank_CDLG_many_point}
    }
    \subfigure[Distribution of \textit{MONBM-CDLG} solutions on the \textit{COMPAS} dataset]{
        \includegraphics[width=\textwidth]{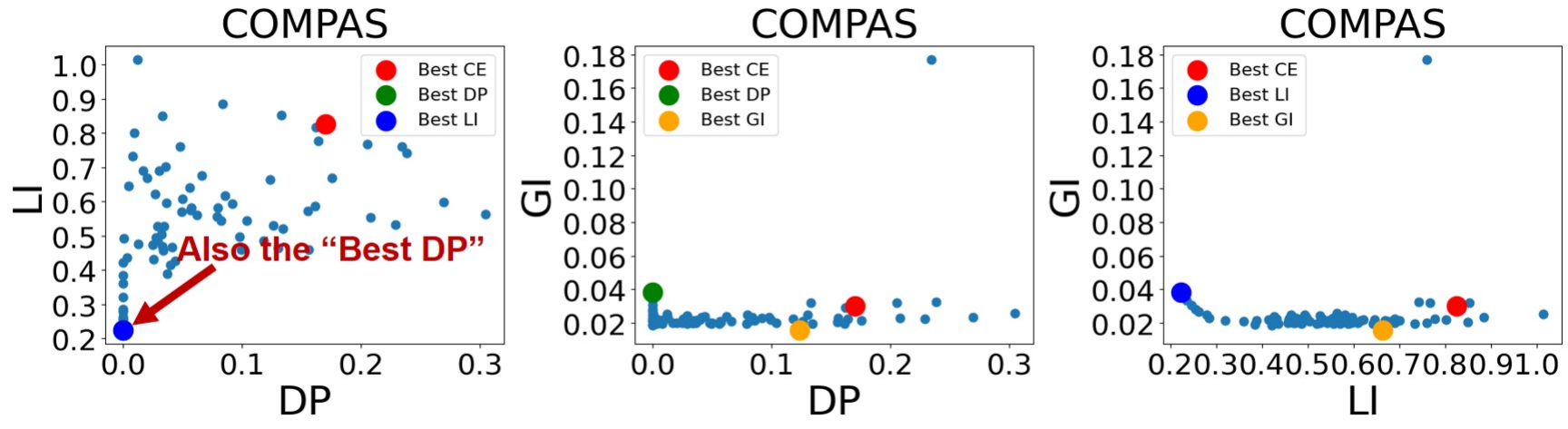}
        \label{fig:compas_CDLG_many_point}
    }
    \subfigure[Distribution of \textit{MONBM-CDLG} solutions on the \textit{Default} dataset]{
        \includegraphics[width=\textwidth]{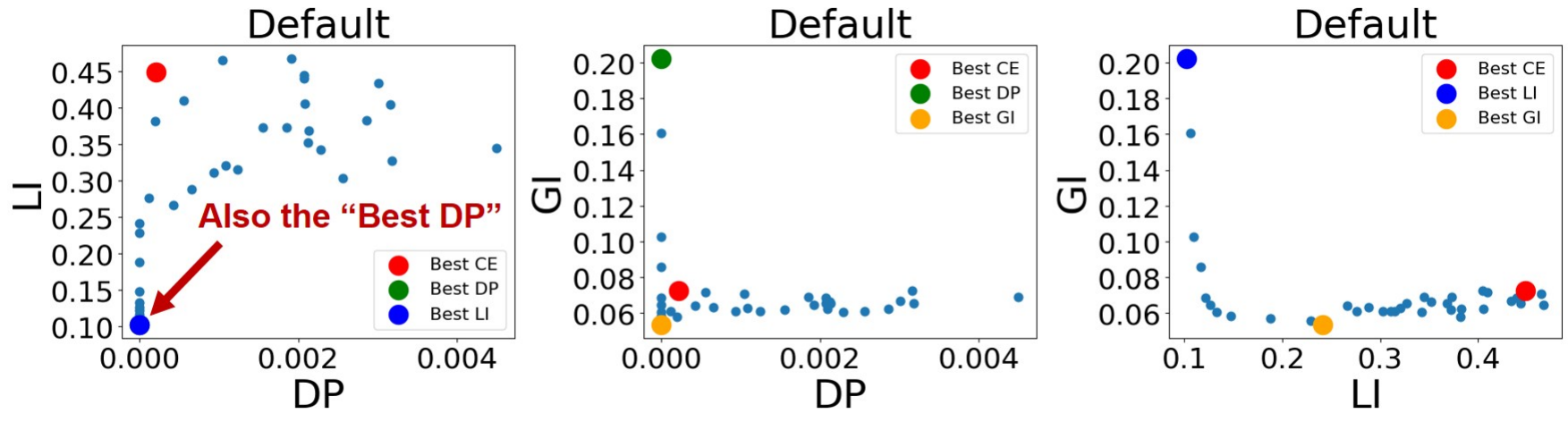}
        \label{fig:default_CDLG_many_point}
    }
    \subfigure[Distribution of \textit{MONBM-CDLG} solutions on the \textit{LSAT} dataset]{
        \includegraphics[width=\textwidth]{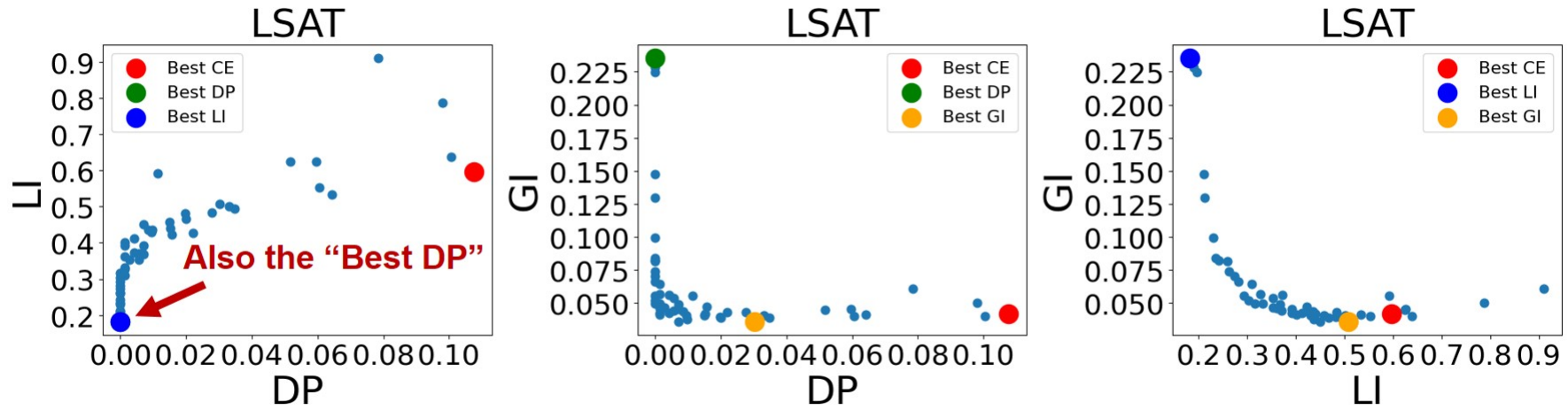}
        \label{fig:lsat_CDLG_many_point}
    }
        \caption{Distribution of all \textit{MONBM-CDLG} non-dominated solutions across pairs of trustworthy objectives to examine their interrelationships. Each subplot visualizes the solutions obtained from one arbitrary run of the \textit{MONBM-CDLG} instantiation on a given dataset, showing pairwise relationships between \textit{DP}, \textit{LI}, and \textit{GI}.}
    \label{fig:CDLG_many_point_sup}
\end{figure*}

\begin{figure*}[htbp]
    \centering
    \includegraphics[width=\textwidth]{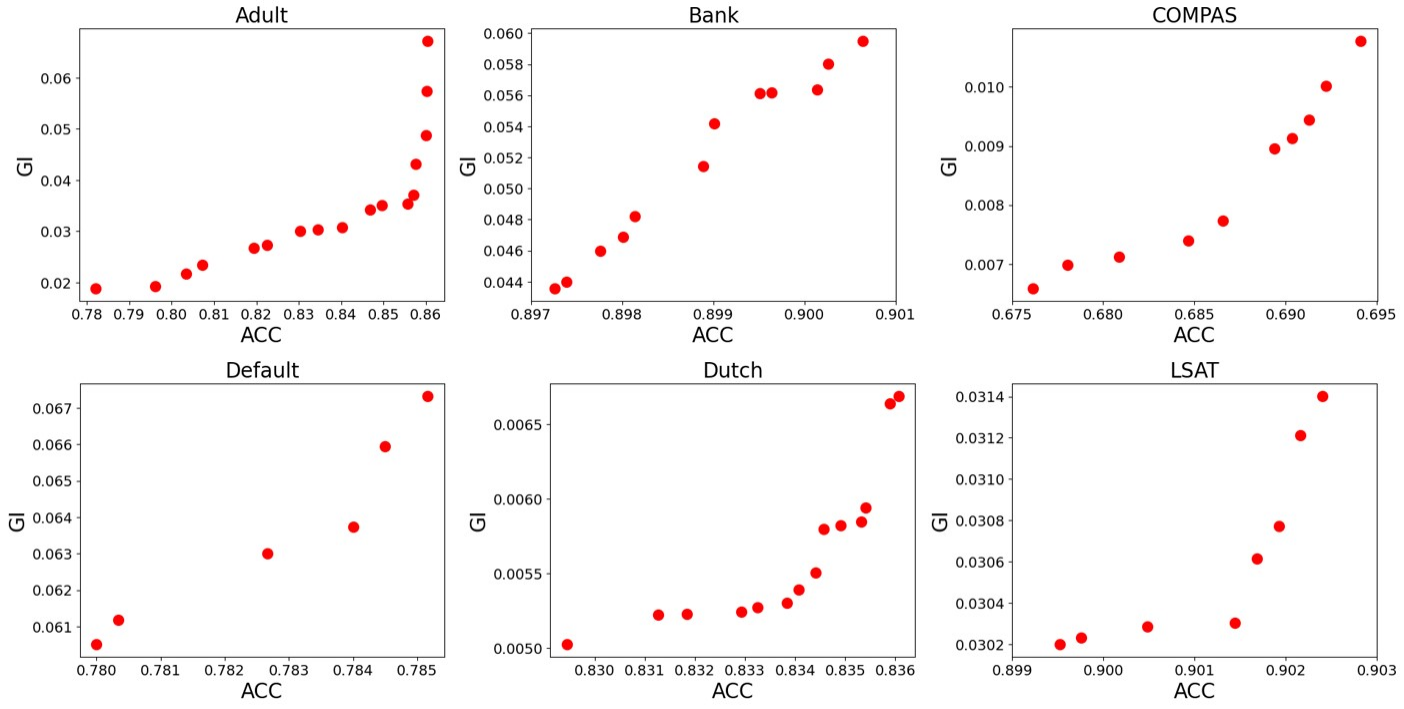}
    \caption{Performance of the non-dominant knee point solutions obtained by \textit{MONBM-CG} in terms of model accuracy and \textit{GI} metrics on one arbitrary trial.}
    \label{fig:BCE_GI_distribution_result}
\end{figure*}

\begin{figure*}[htbp]
    \centering
    \includegraphics[width=0.85\textwidth]{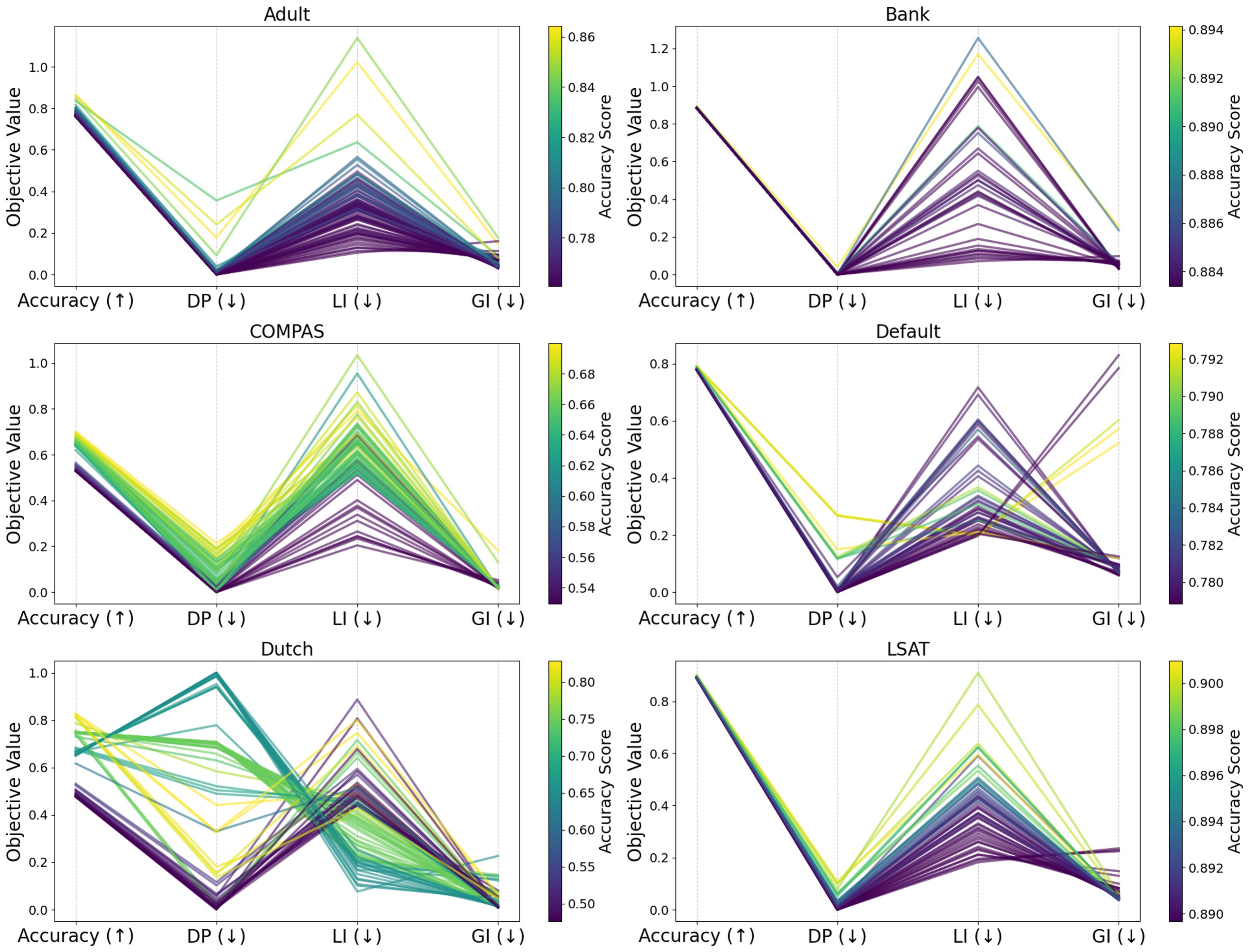}
    \caption{Performance of the non-dominated knee point solutions obtained by \textit{MONBM-CDLG} in terms of model accuracy, \textit{DP}, \textit{LI}, and \textit{GI} metrics on one arbitrary trial.}
    \label{fig:BCE_DP_LI_GI_distribution_result}
\end{figure*}

\clearpage

\section{Ablation Analysis of Global Interpretability: Smoothness versus Monotonicity}

To better understand how global interpretability interacts with other trustworthy dimensions, we conduct an ablation study to examine the individual effects of its two components: smoothness ($\mathcal{M}_{s}$) and monotonicity ($\mathcal{M}_{m}$). Specifically, we investigate whether the relationships observed in the original four-objective formulation are primarily driven by one of these two constraints. To this end, we decompose \textit{MONBM-CDLG} into two variants: Accuracy–Fairness–LI–Smoothness (\textit{MONBM-CDLS}) and Accuracy–Fairness–LI–Monotonicity (\textit{MONBM-CDLM}). Consistent with the analysis in the main text, experiments are conducted on the \textit{Adult} and \textit{Dutch} datasets, representing typical and atypical relationship patterns, respectively. The extreme solutions extracted from the final non-dominated sets are summarized in Table~\ref{tab:MONBM_CDL_smooth} and Table~\ref{tab:MONBM_CDL_mono} below.

\begin{table*}[h]
    \caption{Comparison of the mean values of the \textit{ACC}, \textit{DP}, \textit{LI}, and $\mathcal{M}_{s}$ on the test set of the models corresponding to the four extreme points chosen from the final solution set obtained by the \textit{MONBM-CDLS} instantiation on the \textit{Adult} and \textit{Dutch} datasets over 15 trials. \textit{ACC} values are expressed as percentages ($\%$).}
    \label{tab:MONBM_CDL_smooth}
    \centering
    \begin{tabular}{l|cccc|cccc}
        \toprule
            & \multicolumn{4}{c|}{Adult} & \multicolumn{4}{c}{Dutch} \\
            & ACC ($\%$) $\uparrow$ & DP $\downarrow$ & LI $\downarrow$ & $\mathcal{M}_{s}$ $\downarrow$ & ACC ($\%$) $\uparrow$ & DP $\downarrow$ & LI $\downarrow$ & $\mathcal{M}_{s}$ $\downarrow$ \\
            \midrule
            Best CE & 85.8 & 0.178 & 1.029 & 0.384 & 82.4 & 0.328 & 0.700 & 0.388 \\
            Best DP & 76.1 & 0.000 & 0.129 & 0.293 & 52.4 & 0.000 & 0.969 & 0.131 \\
            Best LI & 76.1 & 0.000 & 0.104 & 0.330 & 66.9 & 0.930 & 0.111 & 0.814 \\
            Best GI & 77.2 & 0.044 & 0.673 & 0.068 & 75.5 & 0.576 & 0.667 & 0.045 \\
        \bottomrule
    \end{tabular}
\end{table*}

\begin{table*}[h]
    \caption{Comparison of the mean values of the \textit{ACC}, \textit{DP}, \textit{LI}, and $\mathcal{M}_{m}$ on the test set of the models corresponding to the four extreme points chosen from the final solution set obtained by the \textit{MONBM-CDLM} instantiation on the \textit{Adult} and \textit{Dutch} datasets over 15 trials. \textit{ACC} values are expressed as percentages ($\%$).}
    \label{tab:MONBM_CDL_mono}
    \centering
    \begin{tabular}{l|cccc|cccc}
        \toprule
            & \multicolumn{4}{c|}{Adult} & \multicolumn{4}{c}{Dutch} \\
            & ACC ($\%$) $\uparrow$ & DP $\downarrow$ & LI $\downarrow$ & $\mathcal{M}_{m}$ $\downarrow$ & ACC ($\%$) $\uparrow$ & DP $\downarrow$ & LI $\downarrow$ & $\mathcal{M}_{m}$ $\downarrow$ \\
            \midrule
            Best CE & 85.9 & 0.182 & 1.024 & 0.067 & 82.7 & 0.336 & 0.719 & 0.009 \\
            Best DP & 76.1 & 0.000 & 0.229 & 0.049 & 52.4 & 0.000 & 0.861 & 0.006 \\
            Best LI & 76.1 & 0.000 & 0.115 & 0.076 & 65.8 & 0.972 & 0.099 & 0.048 \\
            Best GI & 76.1 & 0.022 & 0.498 & 0.003 & 71.7 & 0.054 & 0.781 & 0.000 \\
        \bottomrule
    \end{tabular}
\end{table*}

The results of both ablated formulations generally support the observations reported in the main text. First, compared with fairness and local interpretability, both smoothness and monotonicity impose relatively limited performance degradation on predictive accuracy. For example, models optimized toward extreme global constraints (Best \textit{GI}) maintain relatively high predictive performance across datasets, whereas pursuing extreme \textit{DP} often causes substantially larger accuracy reductions. This suggests that global interpretability constraints mainly reshape local geometric properties of the learned shape functions (making them smoother or more monotonic), rather than fundamentally altering the overall decision boundary.

Second, a persistent trade-off between local interpretability and both global interpretability components can still be observed. Improving \textit{LI} encourages sparse feature usage, whereas smoothness and monotonicity regulate the global structure of feature responses. These two forms of interpretability therefore operate through different mechanisms and cannot always be improved simultaneously.

More importantly, the ablation study provides additional insight into the atypical behavior observed on the \textit{Dutch} dataset. Comparing the two formulations reveals a clear difference between smoothness and monotonicity. Under monotonicity optimization (Best $\mathcal{M}_{m}$, Table~\ref{tab:MONBM_CDL_mono}), fairness remains relatively stable (\textit{DP} = 0.054), indicating that encouraging monotonic feature trends does not substantially interfere with fair decision making. In contrast, optimizing smoothness alone (Best $\mathcal{M}_{s}$, Table~\ref{tab:MONBM_CDL_smooth}) leads to a much larger demographic disparity (\textit{DP} = 0.576). This contrast suggests that the fairness–interpretability conflict observed in \textit{Dutch} is primarily associated with smoothness rather than monotonicity. The underlying reason is that the pursuit of smoothness suppresses local fluctuations in continuous shape functions and promotes a more uniform feature response. On the \textit{Dutch} dataset, this smoothing effect reduced the contribution of other features, causing the sensitive attribute ``sex'' to become the dominant feature and exacerbating decision-making bias.

Overall, these findings suggest that global interpretability should not be treated as a single homogeneous concept. Different interpretability constraints may interact with fairness in different ways, and such interactions depend strongly on the characteristics of the underlying data distribution.

\section{Preliminary Exploration on Adaptive Selection Pressure}
\label{sec:adaptive_pressure}

In many-objective optimization scenarios such as the four-objective \textit{MONBM-CDLG} instantiation, evolutionary search often becomes less effective due to increasingly severe conflicts among objectives and weakened selection pressure. As the number of objectives grows, distinguishing promising solutions becomes more difficult, which may reduce convergence performance and the overall quality of the final solution set. To explore whether this limitation can be mitigated, we conduct a preliminary study by introducing an adaptive selection pressure mechanism into the SRA adopted in our framework.

Specifically, the parameter $p_c$ in SRA controls the probability that stochastic ranking prioritizes convergence-oriented information during pairwise comparison and sorting. A larger $p_c$ increases the likelihood that solutions with better objective performance are ranked higher, thereby strengthening selection pressure and encouraging convergence. In contrast, a smaller $p_c$ allows more emphasis on maintaining diverse candidate solutions and preserving exploration capability. Therefore, $p_c$ naturally serves as a controllable mechanism for balancing exploration and convergence throughout the evolutionary process. Based on this observation, we introduce a dynamic parameter schedule to gradually adjust selection pressure during optimization:

\begin{equation}
p_c(g) = p_{start} + (p_{end} - p_{start}) \times \left(\frac{g}{G_{max}}\right)^2,
\end{equation}
where $g$ represents the current generation number, $G_{max}$ denotes the maximum number of generations, $p_{start}$ represents the initial probability, and $p_{end}$ denotes the target probability at the final generation. This quadratic schedule enables broader exploration in early generations by maintaining relatively weak selection pressure, while progressively increasing convergence pressure during later stages of optimization.

In our implementation, $p_{\textit{start}}$ is initialized as 0.45, consistent with the fixed configuration adopted in the original \textit{MONBM} framework, and $p_{\textit{end}}$ is set to 0.8. To evaluate the effect of this adaptive mechanism, we conduct experiments on the \textit{Adult} dataset using the four-objective \textit{MONBM-CDLG} instantiation. Table~\ref{tab:MONBM_compare_AS} presents the comparison between the original \textit{MONBM}with fixed $p_c$ (denoted as \textit{MONBM} w/o AS) and the adaptive variant (denoted as \textit{MONBM} with AS). Following the analysis approach in the main experiments, we report the four extreme solutions selected from the final solution set, where each solution corresponds to the model achieving the best value for one optimization objective.

\begin{table*}[h]
    \caption{Comparison of the mean values of the \textit{ACC}, \textit{DP}, \textit{LI}, and \textit{GI} on the test set of the models corresponding to the four extreme points chosen from the final solution set obtained by the \textit{MONBM-CDLG} instantiation with and without the adaptive strategy (\textit{MONBM} with AS vs. \textit{MONBM} w/o AS) on the \textit{Adult } dataset over 15 trials. \textit{ACC} values are expressed as percentages ($\%$). For each metric, the better average between \textit{MONBM} with AS and \textit{MONBM} w/o AS is highlighted in bold.}
    \label{tab:MONBM_compare_AS}
    \centering
    \begin{tabular}{l|cccc|cccc}
        \toprule
            & \multicolumn{4}{c|}{\textit{MONBM} w/o AS} & \multicolumn{4}{c}{\textit{MONBM} with AS} \\
            & ACC ($\%$) $\uparrow$ & DP $\downarrow$ & LI $\downarrow$ & GI $\downarrow$ & ACC ($\%$) $\uparrow$ & DP $\downarrow$ & LI $\downarrow$ & GI $\downarrow$ \\
            \midrule
            Best CE & 85.7 & \textbf{0.174} & \textbf{1.026} & 0.172 & \textbf{85.9} & \textbf{0.174} & 1.027 & \textbf{0.170}\\
            Best DP & 65.6 & \textbf{0.000} & 0.124 & 0.155 & \textbf{76.1} & \textbf{0.000} & \textbf{0.097} & \textbf{0.151} \\
            Best LI & 65.7 & \textbf{0.000} & 0.108 & 0.167 & \textbf{76.1} & \textbf{0.000} & \textbf{0.097} & \textbf{0.151} \\
            Best GI & 69.7 & 0.008 & 0.463 & 0.036 & \textbf{76.2} & \textbf{0.002} & \textbf{0.414} & \textbf{0.028} \\
        \bottomrule
    \end{tabular}
\end{table*}

The experimental results show that introducing adaptive selection pressure generally improves solution quality, especially for the fairness-, local interpretability-, and global interpretability-oriented solutions, while maintaining competitive predictive performance. Although the performance gap relative to lower-dimensional optimization settings is not completely eliminated—which is expected given the inherent trade-offs in many-objective learning—the results suggest that adaptive pressure adjustment is a promising direction for improving convergence under severe objective conflicts. Developing more advanced adaptive mechanisms or dedicated many-objective operators remains an important topic for future research.

\section{Comparing \textit{MONBM} Instantiations with Baseline Methods on All Four Objectives }

This section compares the mutual dominance relationships between the \textit{MONBM-CD}, \textit{MONBM-CL}, and \textit{MONBM-CDLG} instantiations and their corresponding baseline methods across all four objectives considered, namely \textit{ACC}, \textit{DP}, \textit{LI}, and \textit{GI}. The results are shown in Tables~\ref{tab:CD_dominate_compare_4obj},~\ref{tab:CL_dominate_compare_4obj}, and~\ref{tab:combined_dominance_all_method}. For the four-objective instantiation (\textit{MONBM-CDLG}), we compared it against all ten baseline methods considered in this paper. It is worth noting that \textit{FairEMOL}~\cite{zhang2022mitigating}, being a black-box method, lacks both local and global interpretability and is therefore excluded from this comparison.

From Tables~\ref{tab:CD_dominate_compare_4obj} and \ref{tab:CL_dominate_compare_4obj}, we can see that compared to the two-objective comparison (i.e., the results in Tables 7 and 8 in the main text), the probability of \textit{MONBM-CD} or \textit{MONBM-CL} dominating or being dominated by other methods on all four objectives has significantly decreased, even reaching 0.0. This is intuitive, as none of the methods considered additional objectives during optimization, and these objectives often conflict with each other. Therefore, \textit{MONBM} is also difficult to outperform baseline methods in terms of additional objectives and naturally difficult to be dominated by other methods. However, \textit{MONBM-CD} instantiation still has a non-negligible probability of finding solutions that dominate other methods, which is partially attributed to its prominent advantages in the two dimensions considered, namely accuracy and fairness.

\setlength{\tabcolsep}{1.5pt}
\begin{table*}[htbp]
    \centering
    \caption{Dominance relationship between \textit{MONBM-CD} and other methods in terms of \textit{ACC}, \textit{DP}, \textit{LI}, and \textit{GI} metrics. Results are averaged over 15 trials. ``Do.'', ``Non-Do.'', and ``Be-Do.'' mean the probability that \textit{MONBM-CD} dominates, does not dominate each other, and is dominated by the corresponding method. Dominance metrics for all baseline methods follow~\cite{zhang2022mitigating}.}
    \resizebox{\textwidth}{!}{
    \begin{tabular}{c|ccc|ccc|ccc|ccc|ccc|ccc|ccc}
    \toprule
    & \multicolumn{3}{c|}{Fair-NBM} & \multicolumn{3}{c|}{LFR-NBM} & \multicolumn{3}{c|}{Re-NBM} & \multicolumn{3}{c|}{ROC-NBM} & \multicolumn{3}{c}{LPP-NBM} & \multicolumn{3}{c}{Oracle-NBM} & \multicolumn{3}{c}{FRAPP{\'E}-NBM}\\
    & Do. & Non-Do. & Be-Do. & Do. & Non-Do. & Be-Do. & Do. & Non-Do. & Be-Do. & Do. & Non-Do. & Be-Do. & Do. & Non-Do. & Be-Do. & Do. & Non-Do. & Be-Do. & Do. & Non-Do. & Be-Do. \\
    \midrule 
    Adult & 0.13 & 0.99 & 0.00 & 0.00 & 1.00 & 0.00 & 0.27 & 0.99 & 0.00 & 0.73 & 0.85 & 0.00 & 0.13 & 0.99 & 0.00 & 0.00 & 1.00 & 0.00 & 0.13 & 0.99 & 0.00 \\
    Bank & 0.07 & 0.99 & 0.00 & 0.07 & 0.98 & 0.00 & 0.13 & 0.96 & 0.00 & 1.00 & 0.48 & 0.00 & 0.27 & 0.49 & 0.00 & 0.27 & 0.77 & 0.00 & 0.73 & 0.63 & 0.00 \\
    COMPAS & 0.00 & 1.00 & 0.00 & 0.33 & 0.87 & 0.00 & 0.20 & 0.98 & 0.01 & 0.00 & 1.00 & 0.00 & 0.00 & 1.00 & 0.00 & 0.00 & 1.00 & 0.00 & 0.07 & 0.99 & 0.00 \\
    Default & 0.47 & 0.95 & 0.00 & 0.53 & 0.89 & 0.00 & 0.13 & 0.97 & 0.01 & 0.47 & 0.97 & 0.00 & 0.53 & 0.92 & 0.00 & 0.47 & 0.94 & 0.00 & 0.43 & 0.82 & 0.00 \\
    Dutch & 0.00 & 0.99 & 0.01 & 0.73 & 0.87 & 0.00 & 0.00 & 1.00 & 0.00 & 0.00 & 1.00 & 0.00 & 0.00 & 1.00 & 0.00 & 0.00 & 1.00 & 0.00 & 0.00 & 1.00 & 0.00 \\
    LSAT & 0.07 & 0.99 & 0.00 & 0.07 & 0.99 & 0.00 & 0.00 & 0.96 & 0.04 & 0.00 & 1.00 & 0.00 & 0.40 & 0.95 & 0.00 & 0.33 & 0.97 & 0.00 & 0.33 & 0.96 & 0.00 \\
    \midrule
    \midrule
    Mean & 0.12 & 0.99 & 0.00 & 0.29 & 0.93 & 0.00 & 0.12 & 0.98 & 0.01 & 0.37 & 0.88 & 0.00 & 0.22 & 0.89 & 0.00 & 0.18 & 0.95 & 0.00 & 0.28 & 0.90 & 0.00 \\
    \bottomrule 
    \end{tabular}}
    \label{tab:CD_dominate_compare_4obj}
\end{table*}

\begin{table}[htbp]
    \centering
    \caption{Dominance relationship between \textit{MONBM-CL} and other methods in terms of \textit{ACC}, \textit{DP}, \textit{LI}, and \textit{GI} metrics. Results are averaged over 15 trials. ``Do.'', ``Non-Do.'', and ``Be-Do.'' mean the probability that \textit{MONBM-CL} dominates, does not dominate each other, and is dominated by the corresponding method. Dominance metrics for all baseline methods follow~\cite{zhang2022mitigating}.}
    \begin{tabular}{c|ccc|ccc|ccc}
    \toprule
    & \multicolumn{3}{c|}{SNAM} & \multicolumn{3}{c}{GAMI-Net} & \multicolumn{3}{c}{NPLAM} \\
    & Do. & Non-Do. & Be-Do. & Do. & Non-Do. & Be-Do. & Do. & Non-Do. & Be-Do.\\
    \midrule 
    Adult & 0.000 & 1.000 & 0.000 & 0.067 & 0.999 & 0.000 & 0.000 & 1.000 & 0.000 \\
    Bank & 0.000 & 1.000 & 0.000 & 0.067 & 0.999 & 0.003 & 0.000 & 1.000 & 0.000 \\
    COMPAS & 0.000 & 1.000 & 0.000 & 0.067 & 0.999 & 0.000 & 0.015 & 0.985 & 0.000 \\
    Default & 0.000 & 1.000 & 0.000 & 0.000 & 1.000 & 0.000 & 0.000 & 1.000 & 0.000 \\
    Dutch & 0.000 & 1.000 & 0.000 & 0.000 & 1.000 & 0.000 & 0.002 & 0.998 & 0.000 \\
    LSAT & 0.000 & 1.000 & 0.000 & 0.000 & 1.000 & 0.000 & 0.000 & 1.000 & 0.000 \\
    \midrule
    \midrule
    Mean & 0.000 & 1.000 & 0.000 & 0.033 & 1.000 & 0.001 & 0.003 & 0.097 & 0.000  \\
    \bottomrule 
    \end{tabular}
    \label{tab:CL_dominate_compare_4obj}
\end{table}

Regarding the comparison results of the \textit{MONBM-CDLG} instantiation in Table~\ref{tab:combined_dominance_all_method}, several key findings can be observed. First, the solutions obtained by \textit{MONBM-CDLG} are not dominated by any of the ten baseline methods across all datasets (the ``Be-Do.'' probability is consistently 0.000), confirming the framework's robustness in maintaining a Pareto-optimal frontier. Second, on baselines with relatively poor overall performance, such as \textit{LFR-NBM} and \textit{ROC-NBM}, \textit{MONBM-CDLG} exhibits a high probability of finding solutions that dominate them across all four objectives simultaneously (e.g., a mean ``Do.'' probability of 0.789 against \textit{LFR-NBM}).

However, for the majority of competitive baselines, \textit{MONBM-CDLG} either rarely or only with low probability achieves complete dominance across all four objectives. This is intuitive: while our specialized instantiations (\textit{MONBM-CD} and \textit{MONBM-CL}) consistently outperform baselines in their respective target dimensions, this localized advantage is often insufficient to guarantee simultaneous superiority across all four conflicting objectives once two additional dimensions are introduced. Nevertheless, the primary value of \textit{MONBM-CDLG} lies in its ability to provide a diverse set of high-quality compromise solutions that remain competitive and non-dominated across the entire trustworthiness spectrum, offering researchers more flexible trade-off options than specialized baselines.

\begin{table*}[htbp]
    \centering
    \caption{Dominance relationship between \textit{MONBM-CDLG} and 10 baseline methods in terms of \textit{ACC}, \textit{DP}, \textit{LI}, and \textit{GI} metrics. Results are averaged over 15 trials. ``Do.'', ``Non-Do.'', and ``Be-Do.'' mean the probability that our method dominates, does not dominate each other, and is dominated by the corresponding method.}
    \label{tab:combined_dominance_all_method}
    \setlength{\tabcolsep}{3pt}
    \resizebox{\textwidth}{!}{
    \begin{tabular}{c|ccc|ccc|ccc|ccc|ccc}
    \toprule
    & \multicolumn{3}{c|}{Fair-NBM} & \multicolumn{3}{c|}{LFR-NBM} & \multicolumn{3}{c|}{Re-NBM} & \multicolumn{3}{c|}{ROC-NBM} & \multicolumn{3}{c}{LPP-NBM} \\
    Dataset & Do. & Non-Do. & Be-Do. & Do. & Non-Do. & Be-Do. & Do. & Non-Do. & Be-Do. & Do. & Non-Do. & Be-Do. & Do. & Non-Do. & Be-Do. \\
    \midrule
    Adult   & 0.000 & 1.000 & 0.000 & 1.000 & 0.607 & 0.000 & 0.000 & 0.100 & 0.000 & 1.000 & 0.135 & 0.000 & 0.000 & 0.000 & 0.000 \\
    Bank    & 0.000 & 1.000 & 0.000 & 0.600 & 0.680 & 0.000 & 0.000 & 1.000 & 0.000 & 0.867 & 0.457 & 0.000 & 0.000 & 1.000 & 0.000 \\
    COMPAS  & 0.067 & 0.997 & 0.000 & 0.867 & 0.626 & 0.000 & 0.000 & 1.000 & 0.000 & 0.200 & 0.983 & 0.000 & 0.133 & 0.987 & 0.000 \\
    Default & 0.000 & 1.000 & 0.000 & 0.533 & 0.929 & 0.000 & 0.000 & 1.000 & 0.000 & 0.133 & 0.988 & 0.000 & 0.000 & 1.000 & 0.000 \\
    Dutch   & 0.000 & 1.000 & 0.000 & 0.800 & 0.821 & 0.000 & 0.000 & 1.000 & 0.000 & 0.067 & 0.999 & 0.000 & 0.000 & 1.000 & 0.000 \\
    LSAT    & 0.133 & 0.986 & 0.000 & 0.933 & 0.633 & 0.000 & 0.067 & 0.996 & 0.000 & 0.000 & 1.000 & 0.000 & 0.267 & 0.927 & 0.000 \\
    \midrule
    Mean    & 0.033 & 0.997 & 0.000 & 0.789 & 0.716 & 0.000 & 0.011 & 0.849 & 0.000 & 0.378 & 0.760 & 0.000 & 0.067 & 0.819 & 0.000 \\
    \midrule
    \midrule
    & \multicolumn{3}{c|}{Oracle-NBM} & \multicolumn{3}{c|}{FRAPPÉ-NBM} & \multicolumn{3}{c|}{SNAM} & \multicolumn{3}{c|}{GAMI-Net} & \multicolumn{3}{c}{NPLAM} \\
    Dataset & Do. & Non-Do. & Be-Do. & Do. & Non-Do. & Be-Do. & Do. & Non-Do. & Be-Do. & Do. & Non-Do. & Be-Do. & Do. & Non-Do. & Be-Do. \\
    \midrule
    Adult   & 0.000 & 1.000 & 0.000 & 0.000 & 1.000 & 0.000 & 0.000 & 1.000 & 0.000 & 0.000 & 1.000 & 0.000 & 0.000 & 1.000 & 0.000 \\
    Bank    & 0.000 & 1.000 & 0.000 & 0.000 & 1.000 & 0.000 & 0.067 & 0.997 & 0.000 & 0.000 & 1.000 & 0.000 & 0.000 & 1.000 & 0.000 \\
    COMPAS  & 0.200 & 0.971 & 0.000 & 0.400 & 0.980 & 0.000 & 0.067 & 0.998 & 0.000 & 0.000 & 1.000 & 0.000 & 0.000 & 1.000 & 0.000 \\
    Default & 0.000 & 1.000 & 0.000 & 0.133 & 0.991 & 0.000 & 0.000 & 1.000 & 0.000 & 0.000 & 1.000 & 0.000 & 0.000 & 1.000 & 0.000 \\
    Dutch   & 0.000 & 1.000 & 0.000 & 0.000 & 1.000 & 0.000 & 0.000 & 1.000 & 0.000 & 0.000 & 1.000 & 0.000 & 0.000 & 1.000 & 0.000 \\
    LSAT    & 0.067 & 0.994 & 0.000 & 0.333 & 0.840 & 0.000 & 0.200 & 0.974 & 0.000 & 0.000 & 1.000 & 0.000 & 0.000 & 1.000 & 0.000 \\
    \midrule
    Mean    & 0.045 & 0.994 & 0.000 & 0.144 & 0.968 & 0.000 & 0.056 & 0.995 & 0.000 & 0.000 & 1.000 & 0.000 & 0.000 & 1.000 & 0.000 \\
    \bottomrule
    \end{tabular}}
\end{table*}

\end{document}